\documentclass[11pt,vi,twoside]{mitthesis}

\usepackage{lgrind}
\usepackage{url}
\usepackage[pdftex]{graphicx}
\usepackage{booktabs}
\usepackage{float} 
\usepackage{cmap}
\usepackage[T1]{fontenc}
\usepackage{longtable}
\usepackage{tabularx}
\usepackage{tikz}
\usetikzlibrary{positioning}
\usepackage{caption}
\usepackage{subcaption}

\usepackage{pdflscape}
\usepackage{caption}

\usepackage{amsmath}
\usepackage{algorithm}
\usepackage[noend]{algpseudocode}

\makeatletter
\def\BState{\State\hskip-\ALG@thistlm}
\makeatother

\usepackage{hyperref}

\ifpdf\hypersetup{%
    pdftitle = {Breast Cancer Diagnosis Using Machine Learning Techniques},
    pdfsubject = {Master Thesis},
    pdfkeywords = {Machine Learning, Breast Cancer, Infrared Imaging, Python, DNN, CNN},
    pdfauthor = {Juan Pablo Zuluaga Gomez},
    pdfcreator = {\LaTeX},
    }
\fi

\hypersetup{pdfborder={0 0 0}} 

\begin{document}

\title{Breast Cancer Diagnosis Using Machine Learning Techniques}

\author{Juan Pablo Zuluaga Gomez}
\prevdegrees{Mechatronics Engineer, Universidad Autonoma del Caribe (2015)}
\department{Department of Electrical Engineering and Computer Science}

\degree{Master in Mechatronics Engineering}

\degreemonth{July}
\degreeyear{2019}
\thesisdate{July, 2019}

\copyrightnoticetext{Femto-ST Sciences \& Technologies\copyright, 2019.}

\supervisor{Noureddine Zerhouni}{Professor}
\supervisor{Zeina Al Masry}{Associate Professor}
\supervisor{Christophe Varnier}{Associate Professor}

\chairman{Pascal Vairac}{Chairman, Femto-ST Sciences \& Technologies\copyright}

\maketitle



\cleardoublepage
\begin{abstractpage}
%
%
%

Breast cancer is one of the most threatening diseases in women’s life; thus, the early and accurate diagnosis plays a key role in reducing the risk of death in a patient’s life. Mammography stands as the reference technique for breast cancer screening; nevertheless, many countries still lack access to mammograms due to economic, social, and cultural issues. Latest advances in computational tools, infrared cameras and devices for bio-impedance quantification, have given a chance to emerge other reference techniques like thermography, infrared thermography, electrical impedance tomography and biomarkers found in blood tests, therefore being faster, reliable and cheaper than other methods. In the last two decades, the techniques mentioned above have been considered as parallel and extended approaches for breast cancer diagnosis, as well many authors concluded that false positives and false negatives rates are significantly reduced. Moreover, when a screening method works together with a computational technique, it generate a "computer-aided diagnosis" system. The present work aims to review the last breakthroughs about the three techniques mentioned earlier, suggested machine learning techniques for breast cancer diagnosis, thus, describing the benefits of some methods in relation with other ones, such as, logistic regression, decision trees, random forest, deep and convolutional neural networks. With this, we studied several hyper-parameters optimization approaches with parzen tree optimizers to improve the performance of baseline models. An exploratory data analysis for each database and a benchmark of convolutional neural networks for the database of thermal images are presented. The benchmark process, reviews image classification techniques with convolutional neural networks, like, Resnet50, NasNetmobile, InceptionResnet and Xception.
\end{abstractpage}




\pagestyle{plain}
\tableofcontents
\newpage
\listoffigures
\newpage
\listoftables

\chapter{Introduction}
\label{ch:introduction}

Cancer is a significant public health disease that affects many people across the world. The early detection of cancer is mandatory to save the patient’s life \cite{cancer_1, cancer_4, art_1}. The cancer is a name given to a variety of diseases caused by the division without stopping and spreading of body's cells \cite{cancer_1, cancer_4}. The normal cycle of cells includes growing, division and finally death, where they would die for both, have got old or become damaged; then, when a cell dies a new one will take their spot. The United Kingdom Cancer Research Institute mentioned that the cancer is produced when abnormal cells divide in an uncontrolled way caused by gene changes, after the disease produce a \textit{\textbf{"Primary Tumor"}}, sometimes cancer could spread to other parts of the body, called \textit{\textbf{"Secondary Tumor"}} \cite{cancer_2}. 

The Globocan 2018 is a fact sheet from the International Agency for Research on Cancer - World Health Organization (WHO), which shows the number of new cases and deaths in 2018 from cancer, just in 2018 the male number of cancer's new cases reach 9.456.418 victims and 8.622.539 new females were registered as well. On the other hand, the mortality number reach 5.385.640 for males and 4.169.387 for females \cite{cancer_3}, further information about the cases is in appendix \ref{appendix:fig}, Figure \ref{cancer_actual}.

Freddie Bray et al. \cite{bray2018global} predicted that there have been 18.1 million new cases of cancer and 9.6 million of deaths in 2018, as well is exposed that lung cancer is the most commonly diagnosed cancer (11.6\% of total cases), standing as the leading cause of cancer death (18.4\% in total) followed by breast cancer (11.6\%). In contrast, there is more than 200 types of cancer \footnote{Article based on the Global Cancer Statistics of 2018 - GLOBOCAN \cite{cancer_3}.}. Indeed, the proportion of breast cancer deceases may vary depending on each world’s region and the risks mentioned above. Precisely, studies have uncovered that the breast cancer mortality-to-incidence ratio in developed countries is 0.20, wherein less developed countries is almost twice, thus 0.37 \cite{bray2018global, bray2012}. Simultaneously, Emerging economies are prone to a higher risk of cancer, so the socioeconomic factor \cite{gersten2002, omran49} together with the aging and growth of the population could lead to higher chances of developing cancer. Equally important, a recurring observation of some types of cancers, shows a significant relation of infection-related and poverty-related diseases with the so-called \textit{"Westernization of lifestyle"} \cite{gersten2002, bray2012, maule2012}, besides, the Human Development Index (HDI) is highly correlated with the presence of cancer. Cancer, as mentioned before, could be presented in more than 200 types, specifically, the second most prevalent cancer disease is \textbf{breast cancer}. The Globocan institution determined that countries such as Colombia, France and Switzerland, during 2018 have been diagnosed 13.380, 56.162 and 7.029 new breast cancer cases in females, respectively. Accordingly, that represents a 24.8\%, 28.6\% and 26.8\% of the whole portion of "new instances" \cite{cancer_3} (respectively). The early detection of this pathology could help to reach a survival rate greater than 90\% of the patients. Then, it is needed to develop an accurate algorithm capable of detect BC in early phase, whose will be cheap and easy to use.

Finally, the International Agency for Research on Cancer, also predicts a rising of 46.5\% in the new cases of breast cancer by 2040 (globally), compared to 2018. Under the above circumstances, the present report aims to apply machine learning techniques to different databases of Computer-Aided diagnosis/detection (CAD) systems to develop an algorithm to detect the probability of having breast cancer with high accuracy. Nonetheless, the whole report will explain the main techniques for detect breast cancer, like imagining, electrical impedance, ultrasound, magnetic resonance, among others.	

This thesis makes part of the SBRA project or "\textit{\textbf{Smart BRA}}", where the main goal is to develop and implement new technology (device) capable of detect breast cancer; first, in a non-invasive and non-intrusive way. Secondly, the tool should be customizable, comfortable, accurate and non-risk for the health. In order to achieve these goals, the first part involves the development of intelligent techniques to accurately detect the disease being the main aim of the present thesis. The team's principal motivation is that globally, most of the underdevelopment countries have not available personal and devices to screen all the population, in fact, the majority of market-available equipment is expensive or/and require trained people, making it many cases non-accessible for underdevelopment countries. The SBRA team has proposed a device which is made from 24 points for measuring the temperature and 24 to apply electrical current and then measure the electrical conductivity of the skin. The technique for detection of breast cancer through skin's heat is called \textbf{\textit{Thermography}}; second, the \textbf{\textit{Electrical Impedance Tomography - EIT}} could measure tissue's conductivity or "impedance". Those techniques feeding a machine learning (ML) model could improve an algorithm's performance, thus providing higher accuracy than the majority of standalone techniques. The SBRA team is composed by Hospital Nord Franche-Comté as medical institute helping with the screening of patients who have the disease, \textit{ZTC Technology} and \textit{CSEM} from Switzerland as companies in charge of the device's technical details and construction, École nationale supérieure de mécanique et des microtechniques and Universite de technologie de Belfort-Montbeliard as educational institutes, finally Femto-ST as French research institute in charge of the state-of-the-art and thermography. To mention, alongside the two "SBRA" techniques (thermography and EIT), it is used a database comprised of features from a blood test. The next sections are focused on explaining the statement of the problem, scope and justification of the present research and the databases. Afterward, the results section is presented the chosen methods for artificial intelligence.

\section{Statement of the problem}
\label{ch:problem}
Breast cancer is a disease that threatens many women's life, thus the early-diagnosis plays a crucial role in saving a patient's life \cite{art_1}. Several studies have found that early-diagnosis of breast cancer could save more than 90\% of all cases with for the five next years. On the contrary, nowadays many countries keep multiple barriers to developing an effective breast cancer screening system, like organizational, psychological, structural, sociocultural and religious. For example, in 2006, more than 25 million women in the United States had no access to health care, make almost impossible obtain an early and accurate diagnostic \cite{art_2}, currently the Kaiser Family Foundation in late 2018 have reported that 11\% from the total amount of women in the USA have not any type of social insurance, which represents more than 10 million women \cite{kaiser}. Differently, a few countries have religious rules where the woman cannot expose the breast; therefore, the common and available methods on the medical field are non-viable for accurate and prior detection of breast cancer. In contrast, devices and techniques that would not need physicians’ direct contact like thermograms or bio-impedance images will make a considerable impact. Presently, several techniques are available in the medical field for breast cancer screening and diagnosis, despite the variety, the main differences lie on cost, method, specificity, sensitivity and patient’s discomfort during the test, among others. Table \ref{x_comparison_between_technique} shows a comparison of the main techniques for breast cancer diagnosis and screening described by Kandlikar et al. \cite{review_thermography_2}. The Mammography is an x-ray technique used as a breast cancer screening and diagnosis method, when an abnormality is in early-stage the mortality index is reduced between 15 to 25\% \cite{art_4, art_5}.  In spite of the mammograms' benefits, the over-diagnosis (false positives), painful procedure, high number of false negatives (usually when the person who evaluate the results, make erroneous assumptions, or in dense breast) and use of x-rays have been making it a method which need to be renovated \cite{art_3} or even replaced by new techniques like thermography and EIT. Under those circumstances, no matter the individual risk of breast cancer, either, genetically (family) or unhealthy lifestyle the current guidelines suggest breast checks every 1 or 2 years starting at the age of 40 or 50 year \cite{art_6}. In general, more information about, guidelines, health benefits, recommended gap time between tests, type of breast cancer and so on, are \cite{art_4, art_6, art_7}. Truthfully, the European Commission has published a document regarding the breast cancer screening and diagnosis guidelines, summarizing that an accurate system is made of screening, diagnosis, communication to the patient, training, interventions to reduce inequalities, monitoring and evaluation of screening and diagnosis. Given these points regarding breast cancer, the development of a entire system capable of minimize the overdiagnosis, composed by different types of screening methods, also reduce the false positives and false negatives cases, additionally, a system where the patients or users can evaluate in a non-invasive and non-intrusive way her/his breast with a high accuracy (precision and recall), comfortable and accessible, is required globally in order to reduce the mortality rate among women having breast cancer. Nevertheless, develop such a system requires many people even teams capable of mix together each benefit from different types of screening methods and make a platform (even Apps) for users and doctors. Therefore, what are the main limitations of the current systems for the detection of breast cancer? How could be developed a system made of different breast cancer screening methods? Which is the best machine learning technique to obtain an accurate result in the screening of breast cancer? It is possible to create an accurate system keeping it cheap, accessible, non-intrusive and non-invasive? 

\begin{landscape}
\addtolength{\voffset}{40pt}
\thispagestyle{empty}
\begin{table}[]
\centering
\caption[Comparison of breast cancer screening and diagnosis techniques]{Comparison of breast cancer screening and diagnosis techniques, structured from \cite{review_thermography_2}}
\label{x_comparison_between_technique}
\begin{tabular}{p{3cm}p{2.5cm}p{1.4cm}p{1.4cm}p{1.4cm}p{3.5cm}p{1.4cm}p{3.5cm}p{3.5cm}}
\toprule

Technique & Mechanism of operation & Sensitivity & Specificity & Cost & Method & Wearable & Cause of discomfort & Recommend for \\ 
\midrule
Mammography & Low energy X-rays & 90\% & \textgreater{}94\% & Moderate & Compressed the breast & No & Pain in the breast & Screening and diagnostic \\

Magnetic Resonance Imaging (MRI) & Magnetic field and pulsating radio waves & 90\% & 50\% & High & Contrast substance injected and dynamic images obtained & No & Claustrophobia, reaction to contrast agent, renal insufficiency patients & Screening in women at high risk for breast caner \\

Positron Emission Tomography (PET) & Gamma rays emitted by tracer substance & 90\% & 86\% & High & Small amount of radioactive tracer injected in the body & No & No & Determine if cancer has spread to other part of the body \\

Ultrasound & High frequency sound waves & 82\% & 84\% & Low & Hand-held or automated ultrasound device & No & No & Screening in dense breast \\

Tomosynthesis (3D Mammography) & Low energy X-rays & 84\% & 92\% & Low & Compressed the breast & No & Pain in the breast & Screening and diagnostic \\

Electronic Palpation Imaging (EPI) & Pressure changes & 84\% & 82\% & Low & Hand-held electronic, tactile sensor & Possible & Pain in the breast & Follow-up after abnormal findings \\

Thermography & Surface Temperature measurement & \textgreater{}90\% & \textgreater{}90\% & Low & Temperature sensors attached to the skin’s surface & Yes & No & Screening \\

Electrical Impedance Tomography (EIT) & Electrical Impedance in the tissue & 87\% & 82\% & Low & Electrodes attached to the skin’s surface & Yes & Tickling for current variation & Screening \\

Biomarkers from Blood Sample Test & Blood samples biomarker & 82\% - 88\% & 85\% - 90\% & Low & Blood results and interpretation & No, test in situ & No & Screening \\ 

\bottomrule

\end{tabular}
\end{table}
\end{landscape}

\section{Scope and justification of the study}

This study will focus on developing a Python-based algorithm for early-detection of breast cancer (BC), to achieve the main goal, it is employed several machine learning techniques (MLT) for score the probability of having -or not- cancer on three different BC databases. Similarly, this study makes part of the SBRA project as explained previously, where it will be used a device composed of 24 temperature sensors for thermography, and 24 points to inject/measure electrical impedance of the body (EIT). Studies from previous years explain the common methods and precautions in applying electrical current to the body \cite{art_8, art_9}, this, cause the device will be placed in the breast and connected to an App for transmitting and interpreting the data. Nonetheless, for the present thesis, the three databases of screening methods (thermography, EIT, blood test) are already provided. The derivable for the SBRA project is an algorithm which makes a prediction based on each of the three types of test. Machine learning techniques, such as linear, logistic regression, decision trees, random forest and artificial neural networks are used to demonstrate the performance.

\section{Limitation of the study}

Thermography, EIT and blood test databases are used to develop a ML model. The SBRA project aims to make a full system (end-to-end) for early breast cancer detection, despite the fact that the physical device is not currently available, the present thesis is only based on the above-mentioned databases. The thermography database is composed of 56 patients where 37 carried anomalies and 19 were healthy women, the population is from Brazil and the following references depict the performance of the Marques, R., \cite{marques2012} and Silva, D., \cite{silva2015} algorithms, also from \cite{silva2014new}. Secondly, the EIT database was created by J., Jossinet \cite{eit_1} in 1996, additionally in 2000 was presented a method for classification of breast tissue by EI spectroscopy, the statistical classification was obtained from a data-set of 106 cases representing six classes of breast tissue. It shows an overall accuracy of 92\% \cite{eit_2}. In chapter \ref{ch:background} and \ref{ch:methodology} are conveyed the database and the main features. Finally, the third database is based in Miguel Patricio., et al \cite{patricio_1} team. They develop an algorithm capable of classifying the presence of cancer on 64 patients with breast cancer and 52 healthy controls, using only blood tests and body mass index (BMI). Then, they apply Monte Carlo Cross-Validation and support vector machines (\textit{SVM}). 

\newpage

\chapter{Related work and main concepts}
\label{ch:background}

Nowadays, many countries have access to several modalities for diagnostic of breast cancer like, X-ray (mammography), computed tomography, MRI, nuclear medicine, ultrasound scans, thermography, EIT, and so on, whereas the majority of these techniques are not easily available in many countries to women having breast cancer; additionally, for more information regarding the global situation of breast cancer, see chapter \ref{ch:introduction} and \ref{ch:problem}. The next sections will define the background of each technique\footnote{Regarding: thermography, EIT and blood test + BMI}.

\section{Thermography}

Digital Infrared Thermal Imaging (DITI) or Thermography is the measurement of the temperature based on the infrared radiation, in contrast to other modalities; it is a non-invasive, non-intrusive, passive and radiation-free technique. In medicine, the skin's surface temperature expose many features because, the radiance from human skin generally is an exponential function of the surface temperature, in other words, is influenced by the level of blood perfusion in the skin \cite{krawczyk2013}, in fact Krawczykm B., et al. summarize "Thermal imaging is hence well suited to pick up changes in blood perfusion which might occur due to inflammation, angiogenesis or other causes" based on \cite{art_10}. As mentioned before, the early detection of breast cancer provides significantly higher chances of survival \cite{art_1, art_1_1}. Thermography, truly has advantages over other techniques, in particular when the tumor is in a early-stage or in dense tissue\footnote{Dense tissue: high index of fibrous or glandular tissue and low of fat} \cite{mammography_2}, indeed, many authors\footnote{AACR, American Association for Cancer Research} had explain before, the high risk for breast cancer when mammographic density is strong \cite{mammography_1}, also in \cite{mammography_3} demonstrated the correlation between body weight, parity, number of births and menopausal status, with regard to breast cancer. The above authors have point out the highly rate of mammograms' false positive cases and also the fact that mammography can detect tumors only once they exceed certain size; in brief, thermography could be a solution to these problems.   

In the medical field, diagnostic of breast cancer using thermography keeps having two different points of view, while one side said, thermography images as an essential tool in decision-making produce a high number of false positives, in conclusion the thermal images were not enough for the initial evaluation of symptomatic patients in Kontos research \cite{kontos2011}, another authors mentioned low precision and recall \cite{low_recall_1, low_recall_2} after the initial evaluation. The other side, explain thermography as an imaging technique capable of overcoming the limitations of mammography. 

\subsection{Initial years of thermography}

\begin{figure}[H]
\centering
\includegraphics[width=3.5in]{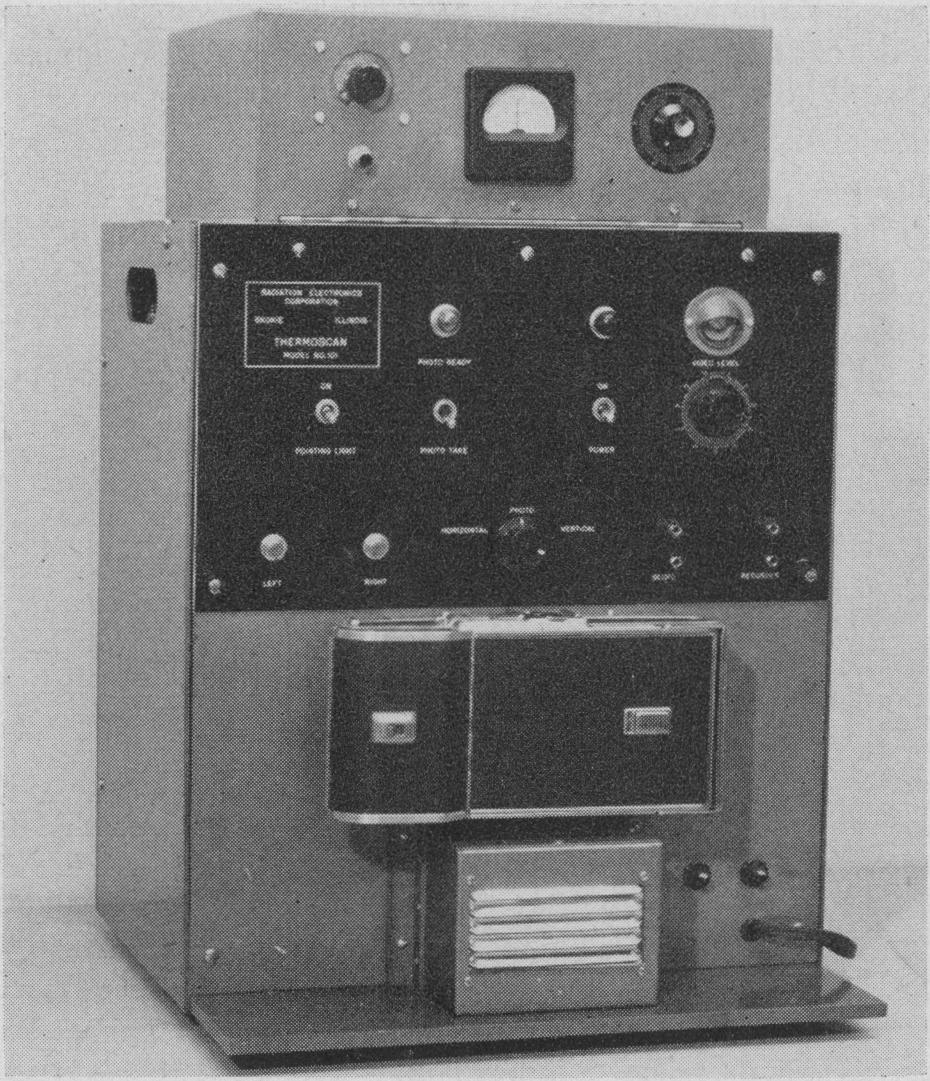}
\caption[Device for skin's infrared imaging]{Device for skin's infrared imaging \cite{thermography_3}.}
\label{first_device}
\end{figure}

The first time ever that was used a thermal/infrared imaging to aid the breast cancer diagnosis was in Montreal in 1956 when the M.D., Lawson, R., recorded the skin's heat energy using a "thermocouple", known as a device made of two dissimilar metals that allows to calculate the electromotive force created by the juncture of these two metals \cite{thermography_1}, also he mentioned that Massopoust, L., and Gardner, W., had used some kind of a system called "Infrared phlebogram\footnote{(1) A graph indicating the pulsing of the blood within the vein. (2) An X-ray image of a vein that has been injected with a dye that is visible on the image taken, Collins Dictionary}" to aid the diagnosis of breast complaints \cite{thermography_2} in 1200 cases. Nevertheless, not was before 1958 when Lawson, R., presented one of the first devices capable of create a infrared imaging, he described the process as follows "At any instant during the scan, the infrared energy radiated from the point on the body at which the scanning mirrors are "looking", is reflected on to a parabolic mirror, thereby focusing the energy from a point on the object on the infrared detecting cell" \cite{thermography_3}, as can be see in the figure \ref{first_device}, the infrared imaging device was called "Thermoscan", in 1965 Lawson's Team obtain a patent where explain the thermography as a diagnostics tool, more information can be found in \cite{patent_1}. Afterwards, a team from Texas used a device called \textit{Pyroscan} for measure the skin temperature, they considered the equipment was expensive but technically was simple, however the false positives were similar compared with mammography \cite{thermography_4}, Williams et al. also presented studies with many common features in 1960 \cite{thermography_5} and in 1964 was granted with a patent \cite{patent_2}, on the other hand and 1964 Mansfield et al. participated in a research testing different heat-sensing devices in Cancer therapy, Swearingen in 1965 concluded two main things, first, the true positives rates was greatly increased when mammography and thermography were applied together, second, the thermography was seen as a new technique for diagnostic procedure in mass screening of the breast \cite{thermography_7}. 

Equally important, during '60s Bowling, B., presented two patents\footnote{More information in Google Patents} regarding thermographic scanners and recorder, he described it as a infrared radiometer mounted on a carriage which can be moved back and forth along a predetermined guided path \cite{patent_3} (with Engborg, N., in 1970), he also patented the process of diagnosis a disease by infrared thermography \cite{patent_4} in 1966. The thermography was became remarkably accepted among many research teams, then in 1971 Isard, H, et al. cooperate in a ten-thousand-cases study, during the four-year research they determined that 61\% of cases were correctly diagnosed with thermography, 83\% with mammography and 89\% applying both techniques \cite{thermography_8}.

Despite, the improvement in infrared imaging technologies, the personal computers' creation, and the efforts shown in the last references regarding the thermography as a "promising" procedure to help physicians in breast cancer diagnostic, the new emerging technologies like MRI, computerized tomography, ultrasound and mammography, stopped and weakened the rising of the infrared imaging studies. Until the '90s when many authors tried to shift from phenomenological thermography to pathophysiologically\footnote{The disordered physiological processes associated with disease or injury, Oxford Dictionary} based thermal imaging, establishing the abnormalities in the skin temperature as a sign of disease. Anbar, M., in 1998 explain the skin's abnormal thermal behavior can be manifested in two different ways; first, changes in "normal" dynamic behavior, i.e., cooling, warming or periodic modulation of temperature; second pathological changes in the spatial distribution of temperature over the skin surface \cite{thermography_9}.

\subsection{Protocols for thermography}

The thermography test, may be considerably affected when guidelines are not followed. In the past, many studies had lack standards and protocols when record thermograms; those could be one of the primary reasons for the poor results. Kandlikar \cite{review_thermography_2} and Ng \cite{ng2009review} mention the following of several standards, in order to obtain high quality and unbiased results. Firstly, it is recommend that patients should avoid tea or coffee before the test, large meals, alcohol and smoking may affect the physicist’s or CAD’s judgement. Secondly, the camera needs to run at least 15 min prior the evaluation, keep a resolution of 100mK at $30^{\circ}$C  and at the same time the camera should have a 120x120 points temperature matrix. Third, is recommend a room’s temperature between 18 and $25^{\circ}$C, humidity between 40\% and 75\%, carpeted floor and avoid any source of heat. Also important, the post-processing phase should be able of identifying the type of breast cancer, either, by physicians or a CAD system. Similarly, Ng et al. in a ninety patients study propose a temperature-controlled room between 20$^{\circ}$C and 22$^{\circ}$C with and humidity of 60\% $\pm$5\%, the patient rested for 15 minutes\cite{tem_new_1}. On the other hand, in order to ensure that patients are within the recommended period, they needed to be in the 5th to 12th and 21st day after the onset of menstrual cycle, since at this time the vascularization is at basal level, with least engorgement of blood vessels \cite{tem_new_2}. 

\subsection{Temperature-based technologies for breast cancer diagnosis}

The term "thermography" is not limited to measure the skin’s temperature, but also rearrange these values in one "image", like an illustration, creating a heat map of the breast’s region of interest (ROI), where each "pixel" express an equivalent temperature value. Ng et al. mention that the presence of localized or focal areas of approximately 1.0$^{\circ}$C or more, including the areola region and significant vascular asymmetry forming "clusters" are features that need to be considered as abnormal \cite{tem_new_1}, they obtained an global accuracy of 59\%, and true positive accuracy of 74\% using Bayes Net. Arena et al. \cite{new_thermography_1} in 2003 have mentioned the benefits of the digital infrared imaging also called "DII", they tested a weighted algorithm in 109 tissue proven cases of breast cancer alongside generating positive or negative evaluation result based on six features (threshold, nipple, areola, global, asymmetry and hot spot), they employed a infrared camera with a 320x240 pixels, and sensitivity of 0.05 degrees. Comparatively, some researchers not only have been focused on the classification of breast cancer, but also on the localization itself of the tumors, Partridge and Wrobel modeled in 2007 a method using dual reciprocity coupled with genetic algorithms to localize tumors, likewise, the smaller tumors or deeply located produce only a limited perturbation making impossible the detection, was concluded \cite{new_thermography_2}, also estimation of tumor characteristics can be found in \cite{new_thermography_2_1}. The research by Kennedy, D., et al. discussed the thermography as breast cancer screening technique alongside the commonest ones, like mammograms and ultrasound, consequently is mentioned the mammography's limitation and drawbacks. On the other hand they concluded that thermograms are early indicators of functional abnormalities that could lead to breast cancer \cite{new_thermography_3}. The infrared cameras used for thermography provide the result in both, a temperature matrix or a heat map image,  Rajendra, U., et al. \cite{new_thermography_4} built an algorithm using support vector machines - SVM\footnote{SVM is one of the most popular machine learning algorithms nowadays.} classifier for automatic classification of normal and malignant breast cancer, the selected database is the same one created by the Brazilian team from \cite{marques2012, silva2015}. Nevertheless, some authors have created a non-public databases that are used for private purposes only. Ng et al. \cite{tem_new_1} presented a computerized detection system with bayes net rules on a ninety patients group, the algorithm yield a 59\% accuracy, but, they also in 2002 proposed a new system using artificial intelligence. Ng's team \cite{tem_new_3} employs an artificial neural network (ANN) coupled with a bayes net ruler, obtaining an accuracy of 61.54\%, but not was before 2008 when his team create a two-steps algorithm, where a linear regression decided whether to choose a ANN with radial basis function or a back-propagated ANN. This study using the same ninety-person database from Singapore (ML) \cite{tem_new_4} achieved a greater accuracy of 81\%. Later, in 2009 Schaefer, G., et al. performed a fuzzy logic classification algorithm where found an accuracy of nearly 80\%, with a population of 150 cases, they explain that statistical feature analysis is a key source of information in order to achieve a high accuracy, i. e., symmetry (mean) between left and right breast, standard temperature deviation, then use the absolute difference as a feature, also cross-correlation with right and left breast histograms, and so on \cite{new_thermography_5}. Araujo, M. presented a symbolic data analysis on 50 patients' thermograms (data: temperature matrices), obtaining 4 variables, minimum and maximum temperature values from the morphological and thermal matrices \cite{new_thermography_6}, also leave one out cross validation framework was implemented. 

\begin{figure}[H]
\centering
\includegraphics[width=5.5in]{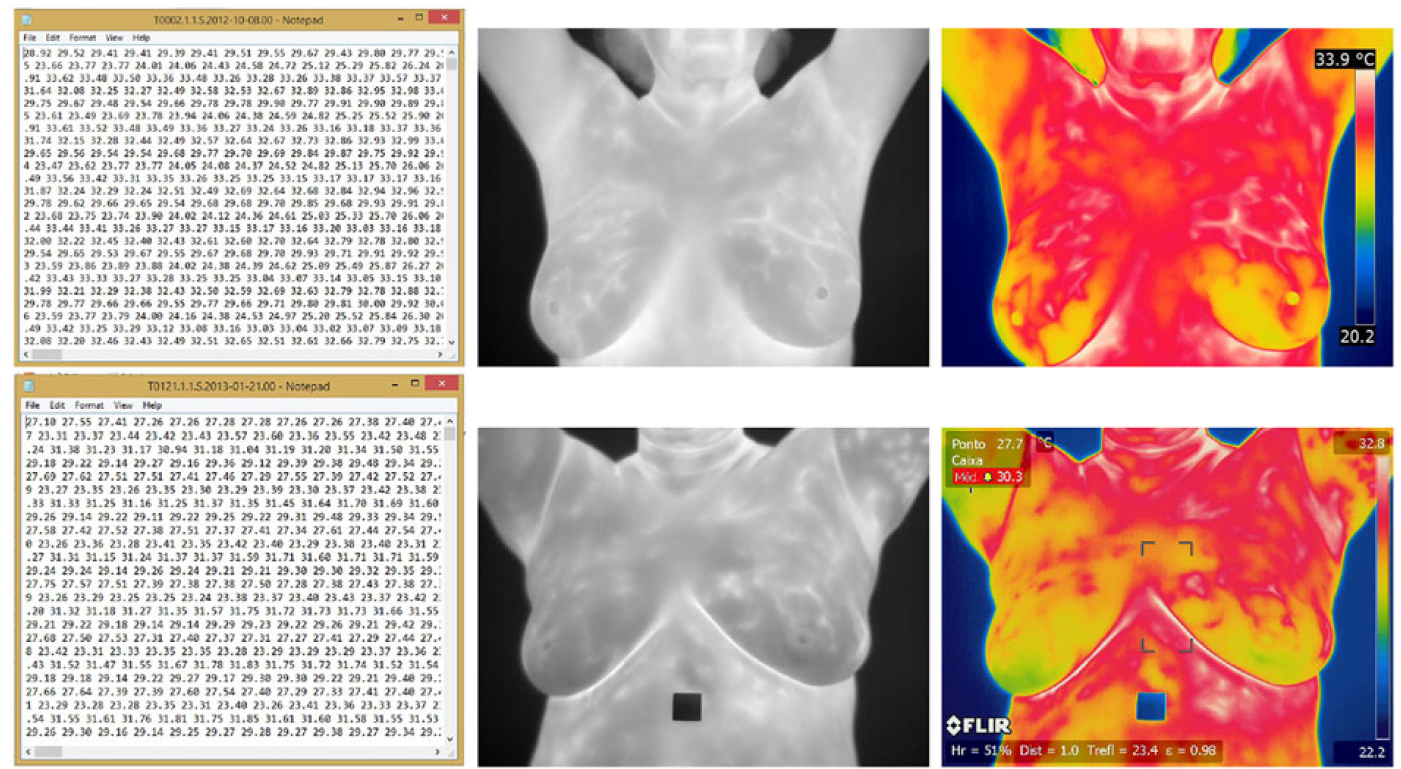}
\caption[Representation of breast thermograms in temperature scale and matrices]{Representation of breast thermograms (a) Temperature matrix (b) Grayscale image (c) Pseudo-color image from \cite{new_thermography_8}.}
\label{new_thermography_8_pic}
\end{figure} 

The number of instances or population size is a key feature for achieving a successful machine learning algorithm, however, in some cases the quantity is not the problem, rather the balance of these ones, consequently Krawczyk, B., et al. in 2013 proposed an ensemble algorithm\footnote{Meta-algorithms that combines several machine learning techniques into just one predictive model decreasing variance, bias and accuracy, the resulted model is better than the other ones separately} for the clustering and classification in breast cancer thermal images, additionally, a 5x2 cross-validation \footnote{A cross-validation method, where is randomly selected a fraction of the data as test and the remaining as training, the procedure is repeat n times, similarly as k-folds} F test was made \cite{krawczyk2013}. Mambou et al. \cite{new_thermography_10} article describes a method to use Deep Neural Networks and support vector machines using the mentioned before database. Initially, they pre-process each thermal image for fitting them in a Deep Neural Network (DNN), then, they extract and normalize the features for feeding into a machine learning algorithm. The database is composed of 56 patients where, 37 carried anomalies and 19 were healthy women, the population are from Brazil (same database).

During the last 6 year, several reviews from infrared technologies have emerged and created a well delimited guide of the current status, main protocols and mew directions of breast cancer diagnosis with thermography \cite{review_thermography_3, review_thermography_1, review_thermography_2}. The segmentation of the produced images from thermal cameras, is another issue to manage in order to boost the global performance of the algorithm, in \cite{new_thermography_7} is mentioned a optimized method of breast thermography images using Extended hidden Markov models (EHMM) in a 140 instances database from the IUT OPTIC non-public database from Iran. Furthermore, Sathish, D., et al. have explained that the thermal camera's information can be interpreted in 3 ways, as a temperature matrix, secondly, gray scale image, or pseudo-color image (or heat map), where the temperature matrix possess more information than the other two, and the normalization of these images could improve the general algorithm \cite{new_thermography_8}, furthermore, the figure \ref{new_thermography_8_pic} help to understand the above assumptions, thermograms taken from two patients. 

In conclusion, certainly the improvement of the computer, the price reduction of the microcontrollers and the increase of breast cancer among women, have brought more and more research teams interested in non-conventional techniques for detect the indicated disease, such as temperature time series with dynamic thermography \cite{new_thermography_9} or \cite{new_thermography_9_1}, deep neuronal networks and SVM \cite{cancer_4} which have a interesting ensemble machine learning method for increase the model's performance, also some authors present a new intelligent textile to measure the skin temperature \cite{new_thermography_11} and dynamical infrared thermal imaging or "DITI" accuracy \cite{new_thermography_13, tem_new_5}. The Table \ref{summarize_table} in the appendix \ref{appendix:tables} summarizes the main comments and performance regarding algorithms used in thermography through the last decades. The first column, comment the scope of the project and the main methodology implemented. The second column indicates which machine learning technique is used in order to predict whether the breast is healthy. The last column exhibits the main achieved results. On the other hand, the last decades improvement in microcontrollers and personal computers have created not only many software and programming languages focused in machine learning techniques, such Python\textregistered, Matlab\textregistered, Orange3\textregistered~(based on Python) and WEKA, but also a global community interested in improve the available libraries. On the other hand, many MLT have been used onto XXI century research in Thermography, improving the final decision of many physicians. Indeed, algorithms like Genetic Algortihms (GA), linear discriminant analysis (LDA), AdaBoost (AB), K-nearest neighborhood (KNN), Support Vector Machines (SVM) with kernels (like, Radial Base Function - RBF, or Gaussian), Naive Bayesian Networks (NBN), Decision Trees (DT), Random Forest (RF), Artificial Neuronal Networks (ANN) and Deep Neuronal Networks (DNN), are examples of the last advances.

\begin{figure}[H]
\centering
\includegraphics[width=3in]{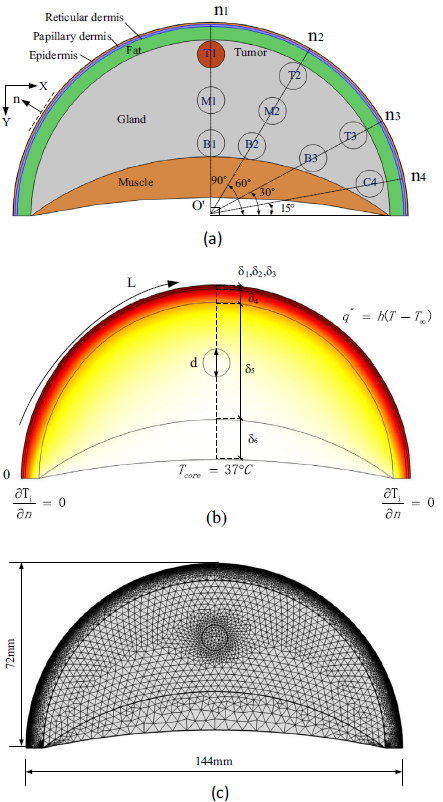}
\caption[Schematic of the breast tissue layers and tumor location on a computational domain]{(a) Schematic of the breast tissue layers and the tumor locations in the computational domain; (b) schematic of the breast tissue layers’ dimension with boundary conditions for steady state; (c) the computational mesh and breast tissue dimensions from \cite{new_thermography_12}.}
\label{new_thermography_12_pic}
\end{figure}

\subsection{Breast: 3D simulation and thermal properties}

The temperature emanated from a human breast may vary depending on a range of features, both, static and dynamical. The first are tumor size, depth and location; also, volume of the breast and quadrant of the suspected tumor. On the other hand, the pathophysiological characteristics surely are different from patient to patient, therefore, some authors have implemented DITI, where the breast undergo a thermostimulation reducing her temperature, then letting it reach a steady state temperature, it is measure the response. The review from Zhou and Herman \cite{new_thermography_12} present 3D models of the heat distribution in healthy and non-healthy breasts, the Figure \ref{new_thermography_12_pic} depicts a breast 3D model in COMSOL\textregistered ~for computing the heat distribution when a tumor is present, \cite{new_thermography_12_1} present similar results. An analysis of thermal patches in the breast could improve many algorithms' accuracy \cite{new_thermography_13}, also Gogoi, U et al. propose a method to locate suspicious regions in thermograms matching them with tumor locations in mammograms \cite{new_thermography_14}, thus, knowing the ground true, they were able to evaluate the efficiency in 3D model and real thermal images. 
  
\begin{table}[H]
\centering
\caption[Properties of breast tissue layers]{The properties of breast tissue layers (from \cite{new_thermography_12})}
\label{temp_properties}
\begin{tabular}{p{2.5cm}p{1.3cm}p{2cm}p{2cm}p{1.5cm}p{2cm}p{1.5cm}}
\toprule
Breast tissue layers & \small Thickness $\delta ~(mm)$ & \small Specific heat C $(J/Kg~K)$ & \small \raggedright Thermal conductivity k(W/m K) & \small  Density $\rho$(kg/$m^3$) & \small Perfusion rate $w_b~(1/s)$ & \small Metabolic HG Q(W/$m^3$) \\ \midrule
Epidermis & 0.1 & 3589 & 0.235 & 1200 & 0 & 0 \\
\raggedright Papillary dermis & 0.7 & 3300 & 0.445 & 1200 & 0.00018 & 368.1 \\
\raggedright Reticular dermis & 0.8 & 3300 & 0.445 & 1200 & 0.00126 & 368.1 \\
Fat & 5 & 2674 & 0.21 & 930 & 0.00008 & 400 \\
Gland & 43.4 & 3770 & 0.48 & 1050 & 0.00054 & 700 \\
Muscle & 15 & 3800 & 0.48 & 1100 & 0.0027 & 700 \\
Tumor & d=10 & 3852 & 0.48 & 1050 & 0.0063 & 5000 \\
\bottomrule
\end{tabular}
\end{table}

Pennes in 1948 \cite{pennes1948analysis} found an equation \ref{Pennes_equ} that model the heat transfer in the human tissue:

\begin{equation}
\label{Pennes_equ}
\rho _i c_i \dfrac{\partial T_i}{\partial t} = k_i \nabla ^2 T_i + \rho _b c_b w_{b,i} (T_b - T_i) + Q_i
\end{equation}

In the equation (\ref{Pennes_equ}), i represents the breast tissue layers of epidermis, papillary dermis, reticular dermis, fat, gland and muscle respectively. $\rho _i$, $c_i$, $k_i$, $T_i$, $Q_i$ and $w_{b,i}$; correspond to tissue layer density, specific heat, thermal conductivity, temperature, metabolic heat generation (HG) rate and blood perfusion rate, respectively. Then, $\rho _b$, $c_b$ and $T_b$; stand for blood density, blood specific heat and arterial blood temperature, respectively. also called, a transient heat conduction Bioequation (\ref{Pennes_equ}), helped \cite{new_thermography_12} research to develop 3D models with the properties of the Table \ref{temp_properties}.

A last key point to realize is the comparison between steady state and dynamical thermography. While steady state thermography measure the uninfluenced breast temperature, the dynamical one, first reduce the breast temperature with cooling in a desired time (usually between 2 and 6 minutes) on top of the breast and afterwards is measure the surface temperature. Nevertheless, parameters like cooling time, cooling temperature, general protocols and patient's age still revision and validation, besides, most of the studies remain in simulation phases \cite{review_thermography_2}. Kandlikar et al. review the main considerations regarding breast tumors simulation, like geometrical parameters, depth, size, and location of malignant or benign tumors \cite{review_thermography_2}. Finally, Lin et al. introduce a new methodology to simulate the early breast tumours using finite element thermal analysis considering parameters like temperature variance, breast contours, deepness of the tumour, and so forth \cite{tem_new_6}. The next section reviews the main techniques and devices for perform EIT and the CAD available systems.

\section{Electrical impedance tomography}

Electrical Impedance Tomography (EIT) or Electrical Impedance Spectroscopy (EIS) is a technique used for evaluate conductivity (also, permittivity) distribution inside the desired object by measuring the voltages between electrodes located in a specific surface. The procedure consists in applying a high-frequency and low current signal through electrodes in the skin, identically, some electrodes are used to record the voltage response in the skin, obtaining a "permittivity" factor. The electric conduction in a tissue can vary depending the type of tissue, the separation between electrodes, and significantly in the presence of cancerous tissue or a tumor, as an illustration, Kubicek, W., et al. have used a four-band electrode (tetra-polar) configuration and the EIT techniques to measure the cardiac output \cite{eit_3}. Equally important, the features of EIT techniques must be explained, the impedance of a living tissue is a complex number, expressed by both, magnitude and phase, in fact, from this information certain sub-features set may be obtained, since in order to reduce noise and make convenient for machine learning techniques. Over the last decades, many research teams suggest basic protocols for reduce the noise and standardization, for example, frequency, max current and limiting circuit, room temperature, time of analysis, quantity of recorded signals (i. e., tetra-polar), impedance, input stray capacitance, and so on. Brown \cite{eit_new_new_1} gives a wider explanation on EIT for health care.

\begin{figure}[H]
\centering
\includegraphics[width=5in]{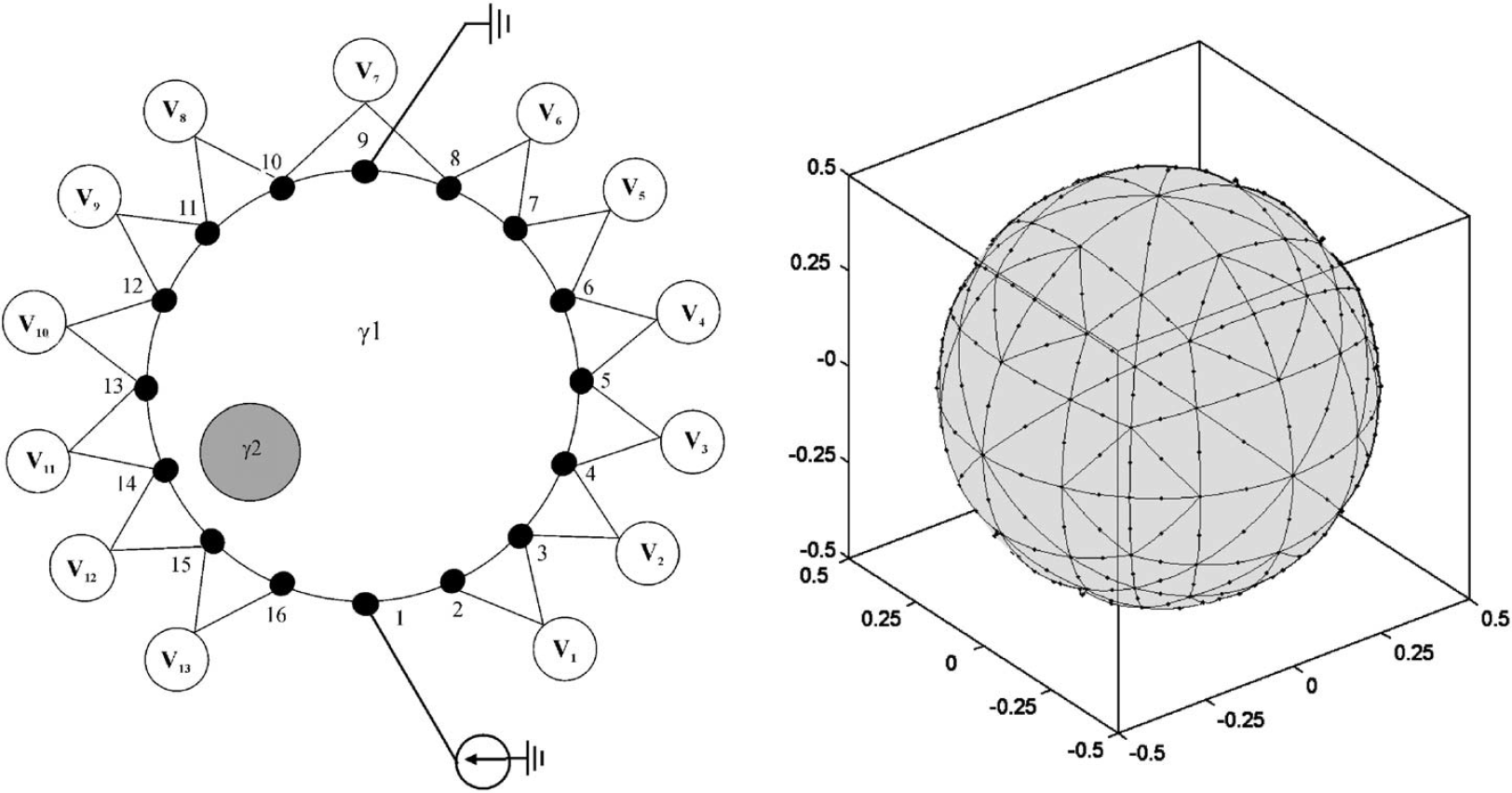}
\caption[3D electrical-simulation of a breast tumor]{(left) electrode-to-electrode configuration; (right) discretization of the internal perturbation with isoparametric element in normalized dimensions from \cite{eit_6}.}
\label{eit_6_pic}
\end{figure}

\subsection{Initial years of electrical impedance tomography}
\label{back:EIT:database}
The EIT systems have been used as a tool to help physicians understand the electro-physical changes in the human body when a tumor or cancerous tissue is present. As mentioned before, in \cite{eit_3} not only used a tetra-polar configuration to measure the cardiac output, but also referenced the initials research of electrical impedance tomography. In the chapter \ref{ch:problem} was mentioned a EIT database, which contain the features of 105 samples of breast tissue and in essence will be the information for the machine learning techniques, also feature engineering\footnote{Process of transforming raw and noise data into features that improve the predictive models, like accuracy, sensitivity and specificity} will be applied. Jossinet, J., et al. in \cite{eit_1} and \cite{eit_2} have explained the main protocols for measure the body's electrical impedance, like frequencies between 0.488kHz and 1MHz using 12 points in the sample, on the other hand the features gathered from the sample were: impedivity ($\Omega$) at zero frequency (I0), phase angle at $500~kHz$ (PA500), high-frequency slope of phase's angle (HFS), impedance distance between spectral ends (DA), area under the spectrum (AREA), area normalized by DA (A/DA), maximum value of the spectrum (MAX IP), distance between I0 and the real part of the maximum's frequency point (DR) and finally, length of the spectral curve (P), more information can be found in \cite{eit_1} and \cite{eit_2}, in summary, the final analysis was in the software called STATISTICA\textregistered , helping to create a set of rules based on features, thus obtaining an overall classification efficiency of 92\%.

In 2003 Zou, Y., and Guo, Z., have reviewed some techniques regarding EIT for breast cancer detection, the main comments were based in the correct separation between malignant and benign tumors, because some evidence has been found that malignant breast tumors have lower electrical impedance than the surrounding normal tissue \cite{eit_5}, in particular, Zou, Y., cited a research article from 1926\footnote{The journal of cancer research, AACR. Department of Biophysics, Cleveland Clinic Foundation, Ohio}, representing the first recorded ever of the electric capacity of breast tumors (see \cite{eit_first}) explaining "A suspension of biological cells or a biological tissue when placed in a conductivity cell, behaves as though it were a pure resistance in parallel with a pure capacity ... In short, it was found that certain types of malignant tumors have a rather high capacity in comparison with benign tumors or with inactive tissues of the same or similar character.", they concluded \cite{eit_first}. 

On the other hand Cheney, M., et al. have proposed a Noser Algortihm\footnote{The inverse conductivity problem is the mathematical problem that must be solved in order for electrical impedance tomography systems to be able to make images \cite{eit_4_1}} approach to solve the EIT reconstruction's problem \cite{eit_4} in brief, the recommend methodology helped other authors. Principal component analysis (PCA) is a statistical procedure to transform from n-dimensional space into a smaller space, taking in consideration the possibly of correlation between variables or features. The main advantage is the reduction in the quantity of features, reducing the overall computational cost, but decreasing the accuracy, usually implemented for machine learning algorithms. As an illustration, in 2007 Stasiak, M., et al. presented a method of PCA analysis together with neuronal networks, for the localization of breast irregularities with EIT \cite{eit_6}, the figure \ref{eit_6_pic} illustrate the electrodes arrangement on the breast, and also the detected voltage, on the right side could be seen the simulated irregularity employing to boundary element method (BEM). 

\begin{figure}[H]
\centering
\includegraphics[width=3.6in]{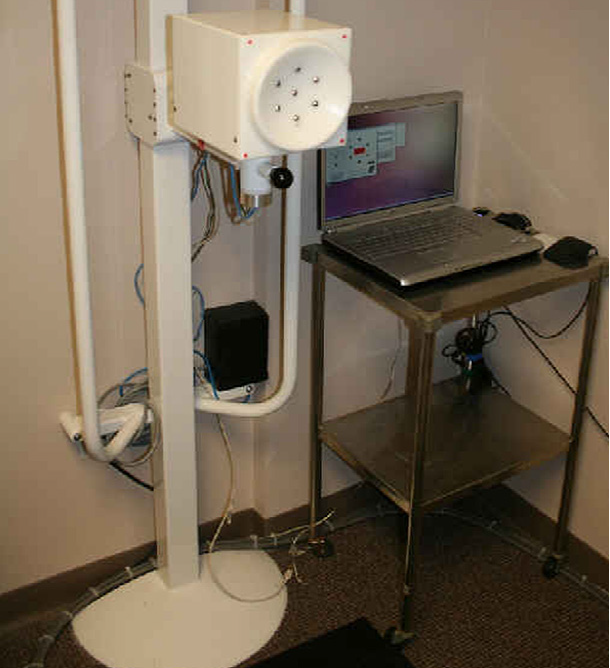}
\caption[Multiprobe resonance-frequency electrical impedance spectroscopy system installed in a clinical breast imaging facility]{The multiprobe resonance-frequency electrical impedance spectroscopy (REIS) system installed in a clinical breast imaging facility from \cite{eit_7}.}
\label{eit_7_pic}
\end{figure}

The artificial neural networks have made a huge impact in the pattern recognition in the last years, thus, Zheng, B., et al. have made a study focused in resonance-frequency electrical impedance spectroscopy (REIS), with a initial set of 140 patients, including 56 who had biopsies; the performance of the overall system was evaluated with ANN and a case-based leave-one-out method \cite{eit_7}, easily can be seen in the figure \ref{eit_7_pic} the 7 electrode-probe used on the patients. In addition to ANN for EIT prediction, in 2012 is presented a multi-layer perceptron\footnote{A class of feedforward artificial neural network, at least is composed by 3 layers} (MLP) model who achieved a 96\% accuracy \cite{eit_8}. Logistic regression, KNN and Naive Bayesian networks were used by Calle-Alonso, F., et al. to classify the EIT data set from \cite{eit_1, eit_2}, furthermore, the key point in obtaining a global accuracy of 97.5\% was to transform the possible six-classes breast tissue: (1) connective tissue, (2) adipose tissue, (3) glandular tissue, (4) carcinoma, (5) fibroadenoma, and (6) mastopathy, into two classes, (1) Carcinoma and (2) Fib+Mas+Gla \cite{eit_9}, to explain, the table \ref{summarize_table_eit} in the appendix \ref{appendix:tables}, the "Acc-1"  refers the two classes approach, "Acc-2" three classes, finally "Acc-3" six classes. 

Advances in EIT have allowed the construction of different devices able to map and create a Electrical Impedance Map (EIM), in 2015 one team have use the T-Scan 2000ED\footnote{T-Scan 2000ED, from Mirabel Medical Systems, Austin, TX} in a 1.103 women, and identifying 29 cancers, also a multiple logistic regression analysis was used for associate clinical variables and EIS results \cite{eit_10}. Subsequently, Haeri, Z., et al.\footnote{Study from: Fraser Health Authority and Jim Pattison Outpatient Care and Surgery Centre (JPOCSC) with study number FHREB2014-065 and 2015s0156, respectively} presented a clinical study using a two different EIT devices, the first setup, is composed by a Covidien electrodes, spectroscope HF2IS and trans-impedance amplifier HF2TA\footnote{HF2IS and HF2TA from Zurich Instruments}. The second setup, is EIS-Probe similar to the first one, but its electrodes and their location of installation are different, the algorithms least absolute deviation (LAD) and least square method (LSM) were implemented for data's analysis \cite{eit_11}, equally important Zarafshani, A., et al. propose a 85 electrodes board to create a Electrical Impedance Mammogram, the main device is described as follows "wide bandwidth EIM system using novel second generation current conveyor operational amplifiers based on a gyrator (OCCII-GIC)", moreover the input current range from  10kHz to 3MHz \cite{eit_12}. 

The Table \ref{summarize_table_eit} (appendix \ref{appendix:tables}) describes references regarding electrical impedance technologies. The background of the electrical impedance tomography as a early breast cancer diagnosis system, have been considered above, nonetheless, the main EIT devices are presented in the next section.

\subsection{Electrical impedance tomography: devices}

The EIT devices available on the market and research area are presented in table \ref{table_eit_devices} (for further details see appendix \ref{appendix:tables}). The main remarks towards this type of equipment are the number of electrodes, where range between 64 and 256, also the method of measurement, between laying on the bed, a probe managed from an expert or a wearable bra. The EIT devices available on the market and research area are presented in table 6.  The main remarks towards this type of equipment, physical, is the number of electrodes, where range between 64 and 256, also the method of measurement like just lying  on the bed, a probe which is human-expert managed or a wearable bra. Electronics, the frequency and magnitude of the low-current signal, the electronic components, and the minimum detectable size. In general, these devices made part of a CAD system where an expert or algorithm give details of the resultant images or signals. Likewise, each year more authors explain the advantages of combining CAD systems with human experts, changing the one-step into two-steps diagnosis systems.  

Given these points, Feza, H., et al. and Lima, G., et al. present two models of electro-thermal system for medical diagnosis. They have concluded that the improvement on the accuracy is greatly augmented when the techniques are employed the same time, rather than performing the diagnostic separately \cite{device_10, device_11}. The next section explains a bit more of both techniques. 

\begin{figure}[H]
\centering
\includegraphics[width=4.8in]{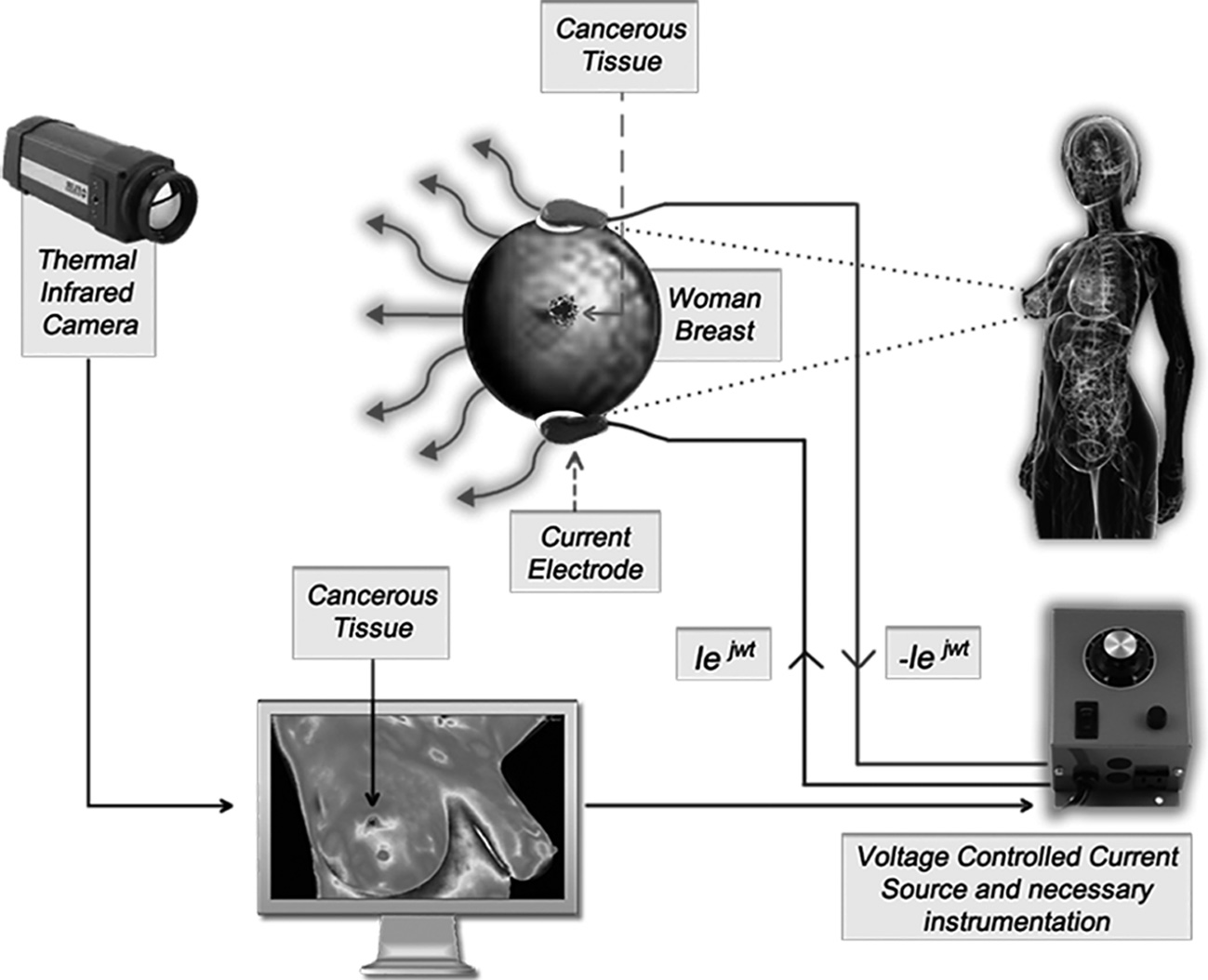}
\caption[Electro-Thermal Imaging System]{Electro-Thermal Imaging System, CAD system coupled with a EIT system and IR camera \cite{device_10}}
\label{eit_thermal_1}
\end{figure}

\section{Electrical impedance tomography and thermography combined systems}

In the last decade, several authors present different Electro-Thermal architectures for breast cancer diagnosis. Feza et al. suggest that a hybrid system is need in order to improve the performance of breast carcinoma diagnosis, because each technique has weaknesses that are highly reduce in electro-thermal systems. Indeed, the method provides a better contrast resolution, in other words, tumors between 3mm and 9mm can be seen with this CAD technique. In general, the method still being theoretical, nonetheless, it works as follows, first a low-current is injected on the breast, afterwards an IR camera take a snapshot of the breast. Under those circumstances, what will be the difference? In detail, the cancerous tissue has almost five to ten-times larger electrical conductivity factor than normal tissue, for that reason, the breast heat map will change and show other insights on the final image, in addition, the frequency of the applied current to the body may change the outcome, for that reason in \cite{device_10}, different parameters are tested. The Feza et al. system works as presented in Figure \ref{eit_thermal_1}, firstly, an electrical current goes through surface of the breast, controlling both, voltage and current. Secondly, an IR camera capture the breast surface temperature, afterward, a CAD system could provide a result using the mapped information. A more recent study from 2019 (Menegaz and Guimaraes) explain that some natural or unnatural (others diseases) processes in the human body, can lead to notably temperature gradients, as a result the thermography evaluation could give erroneous outcomes. On the other hand, they validate the method with silicone phantom samples using hyperplastic materials with simple geometry. A damage metrics or "cancerous tissues" with different thickness were used to measure the global performance \cite{device_11}. The next chapter explain further about the last technique for diagnosis breast cancer, blood test biomarkers.

\section{Blood test: biomarkers for breast cancer diagnosis}

The last decades DNA (deoxyribonucleic acid) revolution have boosted many teams in looking up new techniques for cancer diagnosis like breast, bladder \cite{blood_1}, leukemia \cite{blood_2}, human colorectal carcinoma \cite{blood_3}, prostate \cite{blood_4}, breast \cite{blood_5}, and so forth. Even though, these techniques remain highly expensive for the middle and low-income patients. Nevertheless, at the same time, models based on low-cost data, which could be gather in routine consultations, as blood test and patient's information (age, weight, nutritional habits and so forth), currently stand as a side screening method for detecting breast cancer. 

\subsection{Biomarkers in DNA}

A biomarker is a quantifiable biological indicator, that can help physicians determine whether a person have a specific disease or not. Actually, breast cancer has some established biomarkers that could give insights about a woman's health. First, the estrogen receptor (ER) is one of the most important biomarkers for breast cancer diagnosis. The ER expression indicate which type of treatment to use \cite{blood_6}. On the other hand, the progesterone receptor (PR) is strongly dependent from the estrogen one. The presence of PR but not of ER, tell the patient should undergo a retesting \cite{blood_7}. Fu et al. explain the impact of the hormone receptor (HR) on later periods of breast cancer depending on the age \cite{blood_8}. The specific prognosis and treatment of the high or low aggressive breast tumors, is usually performed with Oncotype DX (ODX) gene expression assay. This stands as another biomarkers technique for select the type of breast cancer treatment. Zemouri \cite{blood_9} study a DNN for breast cancer classification using ODX, however, the high cost in comparison with other techniques establish the ODX assay as not suitable for most of the population. The BRCA1 and BRCA2 genes, could lead to specific types of cancer, depending on their mutations, specifically, breast and ovarian cancers in females. Statistical analysis comment that 12\% of woman in the general population will develop breast cancer sometime during their lives \cite{blood_10}, to put it differently, 72\% and 69\% of women who inherit a harmful BRCA1 or BRCA2 mutation, respectively, will develop a breast cancer by the age of 80 \cite{blood_11}. Another key study shows that BRCA1 mutation carriers would increase the lifetime risk of cancer up to 80\%, nevertheless many others factors may change this number \cite{blood_12}. After all, in Weigel and Dowsett \cite{blood_13} study, they remarks other types of emerging biomarkers that possibly led to breast cancer, greatly reviewing then. Finally, Brennan mention several DNA centered breast cancers \cite{blood_14}.

\subsection{Biomarkers from nipple aspirate fluids and proteomics}

The Nipple Aspirate Fluid (NAF) is a non-invasive method to identify biomarkers for breast cancer early-detection. Wrensch, et al. present a study where NAF is assessed in woman with and without breast cancer \cite{blood_15}, other authors report a significantly increase in NAF basic fibroblast growth factor (bFGF) among breast-cancer confirmed patients \cite{blood_16}. Paweletz’ team \cite{blood_17} use a laser desorption and ionization time of flight mass spectrometry to identify patterns of proteins that might define a proteomic (proteomic: study of proteins) signature for breast cancer, leading to high-risk of developing this one. Proteomics and bioinformatics have helped the discovery of new biomarkers, increasing the sensitivity and specificity in detecting early breast cancer; \cite{blood_18} propose a new technique like the one explained before, with a 103-population, distributed in four possible breast cancer stages and already pathologically confirmed, also is achieve a high Sp and Sn with three biomarkers. An extensive study on proteomics are in \cite{blood_19}. 

\subsection{Low-cost biomarkers and CAD systems}

The main difference between the previous-mentioned biomarkers, and those present in this section is the cost for carrying out a whole diagnosis, from taking the blood test until giving a patient's result. Thought the last two decades advances in MLT and new type of breast cancer biomarkers, many authors explain the benefits and advantages for predicting that disease. Patricio, M., et al. \cite{patricio_1} develop a SVM, RF and LR models to predict breast cancer from Glucose, Insulin, HOMA, Leptin, Adiponectin, Resistin and MCP-1 as biomarkers, achieving a Sn of 82 and 88\% and Sp between 85 to 90\%. On the other hand, Dalamaga et al. \cite{blood_20} assessed serum Resistin as a predictor in a population of 103 post-menopausal woman with pathologically confirmed breast cancer, achieving ROC AUC of 0.72. The Serum levels of leptin, adiponectin and carbohydrate antigen 15‐3, as well as anthropometric and biochemical parameters are analyzed in 88 female patients in order to see the correlation with breast cancer; the models reach a Sn of 83.3\%, Sp of 80\%, positive predictive value of 83.3\% and negative predictive value of 80\% \cite{blood_21}. Hsiao-Lin, H., et al. \cite{blood_22} explored carcinoembryonic antigen, breast cancer‐specific antigen 15.3, tissue polypeptide specific antigen, interleukin-2 receptor and insulin-like growth factor binding protein-3, as biomarkers to detect the present of breast cancer.
 
Another key point may come with Vandenberghe, et al. research, where they assure that deep learning models like DNN and CNN outperform classical MLT in cell image classification of HER2 (human epidermal growth factor receptor-2) gene classification for breast cancer diagnosis. The dataset consists of 74 whole-slide images of breast tumor resection samples, also the algorithm could distinguish between stroma cells, immune cells, 0 tumor cells, 1+ tumor cells, 2+ tumor cells, 3+ tumor cells. In order to compare the results, three types of model are implemented, CNN, RF and SVM, with 10-fold Cross-validation and hyper-parameter tuning \cite{blood_23}. RF, SVM, Multiple instance learning, CNN, autoencoders, RNN, and adversarial networks are some types of MLT reviewed in \cite{blood_24}, also is an extensive review of CAD systems in imaging processing for diagnostic of breast pathologies. Additionally, Saha and Chakraborty, create a specific architecture for HER2, called Her2Net. Her2Net, is made of a convolutional and deconvolutional part, consists mainly of multiple convolution layers, max-pooling layers, spatial pyramid pooling layers, deconvolution layers, up-sampling layers, and trapezoidal long short-term memory (TLSTM). The achieved scores are Sn 96.6\%, Sp 96.7\%, F1 score 93\% and accuracy of 98.3\% \cite{blood_25}. Finally yet importantly, Mukundan explain the importance of feature selection and characteristics for having a good MLT algorithm, for example, reduce information redundancy, maximizing inter-class separability, and improving classification accuracy in the combined feature set, in conclusion they perform an extensive analysis of classification algorithms as LR, DT, RF, SVM, ANN and CNN \cite{blood_26}.

\section{Machine learning techniques}
\label{ch:ML_techniques}
The science covering the learning from data is well-known as \textit{\textbf{Machine Learning}}. In other words, machine learning (ML) is an application of artificial intelligence, where a program, system or algorithm have the ability to learn and improve its "experience" automatically, without being explicitly coded for it. A general definitions of machine learning:

\begin{quote}
\textit{"Machine Learning is the field of study that gives computers the ability to learn without being explicitly programmed"}\\
- Arthur Samuel, 1959
\end{quote}

A more engineering-oriented definition is:

\begin{quote}
\textit{"A computer program is said to learn from experience E with respect to some task T and some performance measure P, if its performance on T, as measured by P, improves with experience E"}\\
- Tom Mitchell, 1997
\end{quote}

The main keys for obtain a good machine learning models, begins with a optimal observations or "data", e.g. examples or instances, then the system will take as input the data (or database) for training purposes. In most of the cases, there is a train, validation and testing set, that compose the "learning and testing phase". Secondly, the aim of machine learning is to allow computers learn automatically without -or minimal- human intervention, depending on the degree of intervention during the training process, the algorithm can be categorized. Supervised, unsupervised, semi-supervised and reinforcement learning, are the type of ML methods, the next subsections review each group and provide a definition based on  Aurelien Geron's book called  \textit{\textbf{"Hands-On Machine Learning with Scikit-Learn and Tensorflow"}} \cite{geron2017hands}.

\subsection{Supervised machine learning algorithms} 

The supervised learning approach applies when the data used for train includes the desired solutions, in many cases called "targets" or "labels". The two principal task in this methodology are "classification" or "predict a numeric target", an example for this cases are, predict whether a woman have cancer and predict the chances of having cancer, respectively. Additionally, the learning process is done under a defined cycle of try, error, update and learn. Finally, after the training process, the algorithm should be capable of create an inferring function to make predictions.  

\subsubsection{Linear Regression}

The linear regression is one of the most simply techniques for prediction. A linear function is created computing a weighted sum of the input features, plus a constant called the \textbf{\textit{bias term}}, the equation \ref{ML_linear_re} is a model of a linear regression.  

\begin{equation}
\dot{y} = \theta _0 + \theta _1x_1 + \theta _2x_2 + \cdots + \theta _nx_n
\label{ML_linear_re}
\end{equation}

Where:
\begin{itemize}
\item $\dot{y}$ is the predicted valued,
\item n is the number of features,
\item $x_i$ is the $i^{th}$ feature value,
\item $\theta_j$ is the $j^{th}$ model parameter (including the bias term $\theta_0$) \cite{geron2017hands}.
\end{itemize}

\subsubsection{Logistic Regression}

The logistic regression is commonly used to estimate the probability that an instance belongs to a particular class, in other words, multi-class classification. The equation \ref{ML_log_re} is a logistic regression in a vectorized model, and the logistic also called the \textit{logit}, noted $\sigma (\cdot)$ -is a sigmoid function (i.e., S-shaped) that outputs a number between 0 and 1 \cite{geron2017hands}. From equation \ref{ML_log_re}, $\sigma$ represents the "sigmoid" function (S-shaped) that outputs a number between 0 and 1, $\theta$ is the model's parameter vector, containing the bias terms. Finally, $x$ is the instance's feature vector. 

\begin{equation}
\dot{p} = h_{\theta} (x) = \sigma(\theta ^ T \cdot x)
\label{ML_log_re}
\end{equation}

\subsubsection{k-Nearest Neighbors}

K-Nearest Neighbors or "KNN" is a non-parametric pattern recognition algorithm capable of solving regressions and classification problems. This instance-based ML technique, works with multi-dimensional input vectors, which are labeled. In the training process are assigned "K" number of "neighbors", afterwards the algorithm should recognized which neighbor is closer to each training sample. After obtaining a rigorous model, each new test sample will be recognized and weighed depending on their closer neighbor. 

\subsubsection{Support Vector Machines}

Support Vector Machine (SVM) is a very powerful and versatile Machine Learning model, capable of performing linear or nonlinear classification, regression, and even outlier detection. SVMs are particularly well suited for classification of complex but small- or medium-sized datasets. The benefits of SVMs over linear or logistic regression lies in the decision boundaries created for the model or in other words \textbf{\textit{"Support Vectors"}}. Again, the SVM decision boundaries are known as large margin classification. More information regarding the algorithms could be found in the chapter \ref{ch:methodology}. 

\subsubsection{Decision Trees}

Decision Trees (DT) are versatile Machine Learning algorithms that can perform both classification and regression tasks, and even multi-output tasks. A DT can be inferred as a decision support tool that uses a tree-like algorithm, that have many parameters that could change the outcome. DT are also the fundamental components of Random Forests. The DT could be linearized into decision rules, that depending on the output, may change the output in the leaf node. DT commonly automatically create orthogonal decision boundaries (all splits are perpendicular to an axis) during training, which makes them sensitive to training set rotation  \cite{geron2017hands}.

\subsubsection{Random Forest}

A random forest (RF) algorithm is an ensemble learning technique\footnote{A group of predictors working towards a common goal is called an Ensemble Learning Predictor} full of decision trees, generally trained via the bagging method\footnote{The sampling is performed with replacement} (or sometimes pasting\footnote{The sampling is performed without replacement}). Reducing the over-fitting is one of the main advantages between RF over DT. Similarly to DT, the RF has a equal (or more) quantity of hyper-parameters. The RF is recognized as a top architecture for structured or stacked data, as it is the blood biomarkers + BMI and EIT database. The next chapters explain the main hyper-parameters tuned on this model.

\subsubsection{Artificial Neural Networks}

Artificial Neural Networks\footnote{Some neural network architectures can be unsupervised, such as autoencoders and restricted Boltzmann
machines. They can also be semisupervised, such as in deep belief networks and unsupervised pretraining \cite{geron2017hands}} (ANN) are computational architectures inspired by the brain. The ANN are based on a group of connected "units" or "neurons" which in a some way model the neurons in an biological brain. Identically to the variety of neurons in a brain, currently exist many types of ANN, like, Deep Neural Network (DNN), Convolutional Neural Network (CNN), Recurrent Neural Network (RNN) and Long short-term memory (LSTM). A basic ANN structure is made of layers, input, "x" number of hidden layers and an output layer; depending on the number of class to label, the output layer could have 1 or more units (when is required a two-label classification, one unit could be enough).

\def\layersep{2.5cm}

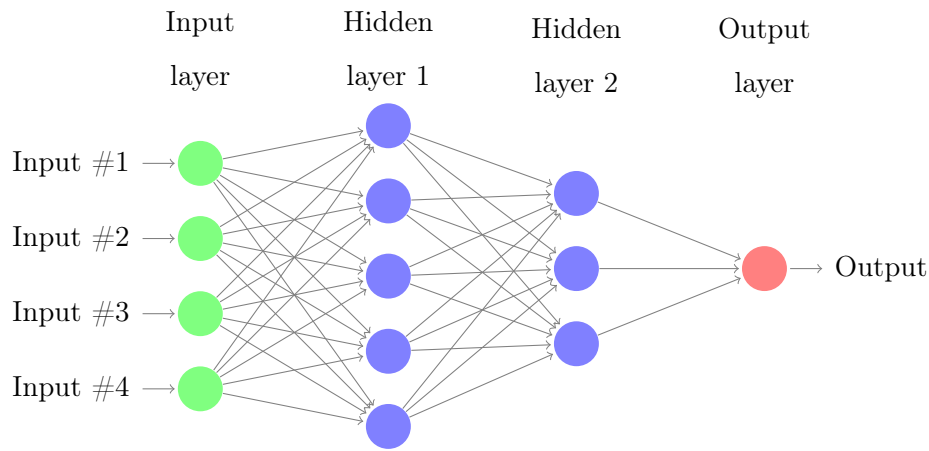
\begin{figure}[H]
\begin{center}
\begin{tikzpicture}[shorten >=1pt,->,draw=black!50, node distance=\layersep]
    \tikzstyle{every pin edge}=[<-,shorten <=1pt]
    \tikzstyle{neuron}=[circle,fill=black!25,minimum size=17pt,inner sep=0pt]
    \tikzstyle{input neuron}=[neuron, fill=green!50];
    \tikzstyle{output neuron}=[neuron, fill=red!50];
    \tikzstyle{hidden neuron 1}=[neuron, fill=blue!50];
	\tikzstyle{hidden neuron 2}=[neuron, fill=blue!50];
    \tikzstyle{annot} = [text width=4em, text centered]

    \foreach \name / \y in {1,...,4}
        \node[input neuron, pin=left:Input \#\y] (I-\name) at (0,-\y) {};

    \foreach \name / \y in {1,...,5}
        \path[yshift=0.5cm]
            node[hidden neuron 1] (H-\name) at (\layersep,-\y cm) {};

    \foreach \name / \y in {1,...,3}
        \path[yshift=-0.4cm]
            node[hidden neuron 2] (HH-\name) at (\layersep + \layersep, -\y cm) {};

    \node[output neuron,pin={[pin edge={->}]right:Output}, right of=HH-2] (O) {};

    \foreach \source in {1,...,4}
        \foreach \dest in {1,...,5}
            \path (I-\source) edge (H-\dest);

    \foreach \source in {1,...,5}
        \foreach \dest in {1,...,3}
            \path (H-\source) edge (HH-\dest);

    \foreach \source in {1,...,3}
        \path (HH-\source) edge (O);

    \node[annot,above of=H-1, node distance=1cm] (hl) {Hidden layer 1};        	\node[annot,above of=HH-1, node distance=1.8cm] (h2) {Hidden layer 2};
    \node[annot,left of=hl] {Input layer};
    \node[annot,right of=h2] {Output layer};
\end{tikzpicture} 
\end{center}
\caption{Artificial Neural Network layout}
\label{ANN}
\end{figure}

A basic ANN structure is shown in Figure \ref{ANN}, which have three layers, an input, one hidden and one output layer. The number of input and output units, could tell the type of database, for example, being four units as input, it is supposed that there are four features and the target is a regression problem.  

\subsection{Unsupervised machine learning algorithms}   

Contrary to supervised learning, the unsupervised approach try to identify hidden patterns and structures into the database creating an inferring function. The information or databases used are neither classified nor labeled. In some cases, the main goals of this type of learning are clustering, visualization and dimensionality reduction or association rule learning. It is important to recall that the unsupervised techniques allow to find hidden patterns from unlabeled data, that other techniques cannot. In fact, some of this techniques are used as starter point for more complex algorithms.

\subsubsection{Principal Component Analysis}
Principal Component Analysis (PCA) is the most popular dimensionality reduction algorithm. The algorithm first try to identify the best and closest hyperplane that lies closest to the input data, afterwards it projects the data onto it. A basic step before the transformation, is the selection of a desired hyperplane, for example, onto 2D or 3D. A standard PCA algorithm should be capable of find the best projection, preserving the maximum amount of variance while minimizing the losing of information when compared with other projections.

\subsubsection{Locally-Linear Embedding}

Locally-Linear Embedding (LLE) is a manifold learning technique that does not depend on projections like PCA. The LLE method works by first measuring
how each training instance linearly relates to its closest neighbors, and then looking for a low-dimensional representation of the training set where these local relationships are best preserved.

\subsubsection{t-distributed Stochastic Neighbor Embedding}

t-distributed Stochastic Neighbor Embedding or "t-SNE" reduces dimensionality while trying to keep similar instances close and dissimilar instances apart. It is mostly used to visualize clusters of instances in high-dimensional space.

\subsubsection{Linear Discriminant Analysis}

Linear Discriminant Analysis (LDA) is a classification algorithm, which during the training it learns the most discriminative axes between the classes, and these axes can then be used to define a hyperplane onto which to project the data. LDA will keep classes as far apart as possible, so LDA is a good technique to reduce dimensionality before running another classification algorithm such as an SVM or RF classifier.

\subsection{Semi-supervised machine learning algorithms}  

The semi-supervised approach, as it can be guessed, fall between supervised and unsupervised. Generally, a clustering is made initially during the training phase, where most of the data is unlabeled and just a few instances are classified. An example of semi-supervised algorithm is the deep belief networks (DBNs), based on unsupervised components called restricted Boltzmann machines (RBMs) stacked on top of one another. RBMs are trained sequentially in an unsupervised manner, and then the whole system is fine-tuned using supervised learning techniques \cite{geron2017hands}. 
 
\subsection{Reinforcement machine learning algorithms}   
 
\begin{figure}[H]
\centering
\includegraphics[width=3.2in]{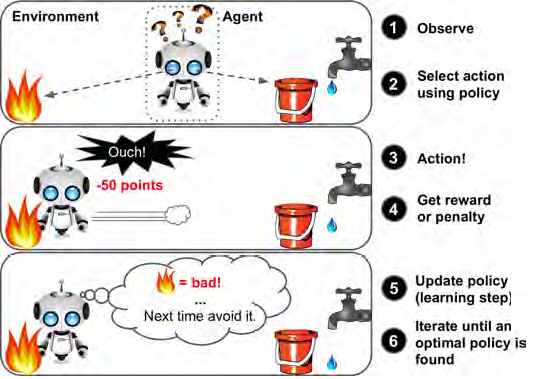}
\caption[Reinforcement learning work-flow]{Reinforcement learning work-flow \cite{geron2017hands}}
\label{ML_reinformcement_learning}
\end{figure}

The reinforcement approach is totally different from the above-mentioned, cause the system (also the "agent") try to interact with its environment, where select and perform an action,  getting a reward or penalty based on some rules. In general, the trial and error search strategy, allow the agent to learn. The learn strategy, is also called "policy", getting the most reward over time. The figure \ref{ML_reinformcement_learning} show a schematic about the trial and error algorithm. 
                  
\section{Hyper-Parameters Optimization}

Nowadays, the \textit{\textbf{"Hyper-parameter"}} term has emerged as one of the leading way to prevent overfitting and boost all machine learning techniques. First of all, a hyperparameter is a parameter whose value is fix  before the training and learning processes begins\footnote{Towards Data Science webpage, "Hyperparameter Tuning", 2019}. Each machine learning technique has specific set of hyperparameters, they may vary in quantity, repercussion and names, but at the end the common goal is to reduce the error (boost up the accuracy) and/or decrease the processing times in training or testing. For example, in a ANN hyper-parameter optimization, is required to find the best number of hidden layers, for both, reducing the error and also avoiding overfitting having more layers than the necessary (at the same time reducing the computational cost). Currently, there are many types of hyperparameters optimization, like grid and random search, genetic algorithms and Bayesian optimization. The present master delves into  Bayesian Optimization, specifically, Parzen Tree Hyper-parameters Optimization.

\begin{figure}[H]
\centering
\includegraphics[width=5in]{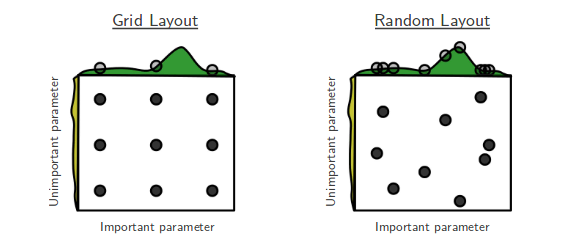}
\caption{Grid and Random 2D search}
\label{ML_grid_rand_search}
\end{figure}

\subsection{Grid and Random Search}

Grid search is a traditional way to perform hyperparameter optimization. It works by searching exhaustively through a specified space of hyperparameters. This type of search assure you will find the best model always, because all possibilities are evaluated. Nevertheless, the time is always an important feature in most of the machine learning pipelines and projects, therefore, grid search in many cases (depending in the quantity of the hyper-parameters space) may take several days, months, or even years. 

Random search differs from grid search mainly in that it searches the specified space of hyperparameters randomly instead of meticulously. The major benefit being decreased processing time. Important to realize that the processing time is decreased, but surely we will not find the optimal combination of hyper-parameters. The figure \ref{ML_grid_rand_search} show a brief explanation of how works grid and random optimization, the 'y' axis represent the back parameters, and the 'x' axis represent the position arrangement in a experiment. 

\subsection{Bayesian Optimization}

There are many types of hyper-parameters optimization, some try to find the best set among all the possible combinations, others just use randomness, but some techniques are based on the phrase "The optimization is just a minimization problem, that's mean searching for the set that yields the lowest error". Bayesian optimization, stands as one of the main techniques for find the optimal set of hyper-parameters for a ML model. 

Bayesian optimization is a probabilistic model based approach for finding the minimum of any function that returns a real-value metric. This function may be as simple as $f(x)=x^2$, or it can be as complex as the validation error of a deep neural network with respect to hundreds of model architecture and hyperparameter choices\footnote{Towards Data Science webpage, "An Introductory Example of Bayesian Optimization in Python with Hyperopt", 2018}.

Bayesian Optimization is also called a sequential model-based optimization, cause it implements the idea of building a probability model of the objective function based on past results. Each time the model receive a new evidence, it updates the probability model (also called "surrogate") creating a new one with the last example/evaluation. The longer the algorithm runs, the closer the surrogate function comes to resembling the actual objective function. The surrogate function can be build differently, then, exist random fores regression, gaussian processs and finally the Tree Parzen Estimator (TPE).

The TPE is the type of surrogate function used in this Master Thesis. The HyperOpt library for Python\textregistered ~allows us to create a hyper-parameters optimization pipeline, using a TPE as surrogate function. Nevertheless, there are four phases in order to build the pipeline. Firstly, the objective function (ML model), then a domain space that will change depending the model, after, the optimization algorithm (TPE), finally, the results and scoring section in order to update the model. 

\pagebreak
\chapter{Methodology}
\label{ch:methodology}

Breast Cancer is the major type of cancer among women, in fact many countries still lack access to diagnosis systems capable of determine whether a woman is having a cancer. Under those circumstances, this Master Thesis pretend to analyze the main advances in three different techniques for breast cancer diagnosis. Then, it is possible to create efficient intelligent algorithms for determining whether women are having cancer? Is it enough the amount of available data on the web regarding blood biomarkers, thermography and EIT for correctly tell a woman her breast state? On the other hand, hereby is developed part of the "SBRA" project main tasks about an extensive bibliography review and first models for breast cancer diagnosis.

\section{Databases: characterization and collection}

The methodologies for collect each databases may vary notably, due the differences between research teams, the availability on computational resources at the moment of gathering the data (i. e. the year) and the phenomenon itself. The table \ref{table_databases} describe each of the target databases for the present Master Thesis. 

First, the thermal database, the author explain in \cite{silva2014new} that the "Database for Mamma Research with Infrared Image" or \textit{\textbf{DMR-IR}}. It possess IR images (digitalized mammograms and clinical data) obtained from patients of the Antonio Pedro University Hospital. These patients are from both, screening and gynecologic department. The DMR-IR has data from healthy and confirmed breast cancer patients. The IR images are captured by a FLIR thermal camera, model SC620, which has sensitivity of less than 0.04 $^{\circ}$C and capture temperatures between -40 $^{\circ}$ C to 500 $^{\circ}$C. The IR images present a dimension of 640x480 pixels (or temperature points) \cite{silva2014new}. The URL for accessing the database is http://visual.ic.uff.br/dmi. 

\begin{table}[t]
	\begin{center}
	\caption{\label{table_databases}Data exploration over the three databases}
	\begin{tabular}{p{2cm}  p{4cm} p{4cm}  p{3cm}}
\hline
Database & Thermography & Electrical Impedance Tomnography & Blood Biomarkers \\ 
\hline
Population 	& 56 		& 106 & 116 \\
Age range 	& 21 - 80 	& 18 - 72 & 24 - 89 \\
Healthy 	& 19 & 70 	& 53 \\
Cancerous 	& 37 		& carcinoma(21), \linebreak fibro-adenoma(15) & 63 \\
Validation Method 		& Mammography, \linebreak Ultrasound & Biopsy & Mammography \\
File(s) 	& Thermal Images & Excel File & Excel File \\
Number of Features & 640x480 (307.200) & 9 & 9 \\

Year 		& 2014 		& 2000 & 2018 \\
Country 	& Brazil 	& France and Portugal & Portugal \\
Comments 	& The most used databased cause is the only available on internet (for most information about MLT and thermal images, see chapter \ref{ch:background})& Reference Database for EIT since 2000, possibly most of the features are high correlated due to the information has came from one source & Recent database, that is a state-of-the-art technique for breast cancer diagnosis \\
Reference 	& \cite{marques2012, silva2015, silva2014new} & \cite{eit_1, eit_2} & \cite{patricio_1} \\
\hline  	   	   
	\end{tabular}
	\end{center}
\end{table}

Secondly, the EIT database consisted of 120 spectra-recorded in samples of breast tissue from 64 patients undergoing breast surgery. Each spectrum consisted of twelve impedance measurements taken at different frequencies ranging from 488 Hz to 1 MHz, 14 spectra were discarded since they exhibited manifestly abnormal features (erroneous current or phase-angle) due to poor tissue collection care and/or data measurement. The classification features were extracted from an electrical impedance spectroscopy (EIS) plots on the Argand plane\footnote{Argand Plane, also called complex plane or z-plane is a geometric representation of the complex numbers established by the real axis and the perpendicular imaginary axis} the co-ordinates of which were the real impedance part ($R_s$) and the negative of the imaginary impedance part ($-X_s$), normally, these plots should have a circular shape, they concluded \cite{eit_1}. 

Lastly, the blood biomarkers database is presented in \cite{patricio_1}, they explain that the database is made of volunteer women newly diagnosed with breast cancer, recruited from the Gynecology Department of the University Hospital Centre of Coimbra (CHUC) between 2009 and 2013. For each patient, the diagnosis came from a positive mammography, also histologically confirmed, therefore the samples were naïve, i.e. collected before surgery and treatment. On the other hand, some patients were excluded due prior cancer treatment, acute diseases; in particular 38 participants were now excluded from the initial database (from 2013) due to having BMI above 40 kg/m2 or due to the absence of at least one of the quantitative variables \cite{patricio_1}.

\section{Databases: analysis, processing and optimization}

After presenting a brief introduction about the databases, most compelling evidence show that the EIT and blood biomarkers are similar type of databases, where the data is stacked in row and columns and is also called \textit{\textbf{"stacked data"}} among the ML communities. In fact, these two databases have the same number of features, nine, nevertheless in EIT there is 6 possible labels, contrary to blood biomarkers whose have "two labels". Differently, the thermal data, is divided in 56 patients, where is "healthy" or carry any type of breast cancer. Each patient has a set of twenty thermal images of the chest, the database has two, a ".txt" file with a 640x480 temperature points, and the thermal image in grey scale. Therefore, the methodology has two approaches the EIT - blood biomarkers and the thermography one. This section has both approaches; each one presents, the data cleaning a pre-processing process, afterwards the data engineering and design of experiments methodologies, finally a hyper-parameter optimization will give the best model, being the most suitable for an evaluation process.

\subsection{Electrical impedance tomography and blood biomarkers}

The nine-feature databases provide the information already cleaned i.e. without noise, then, we can start analyzing the data directly. The first phase for each methodology starts with a highly-recognized technique among the machine learning communities, called \textit{\textbf{"Exploratory Data Analysis"}} or "EDA". It is important to know and have insights regarding the target database, i. e. correlation, mean and distribution plots. 

\subsubsection{Exploratory data analysis}
\label{sub:EDA}
\begin{figure}[H]
\centering
\includegraphics[width=5in]{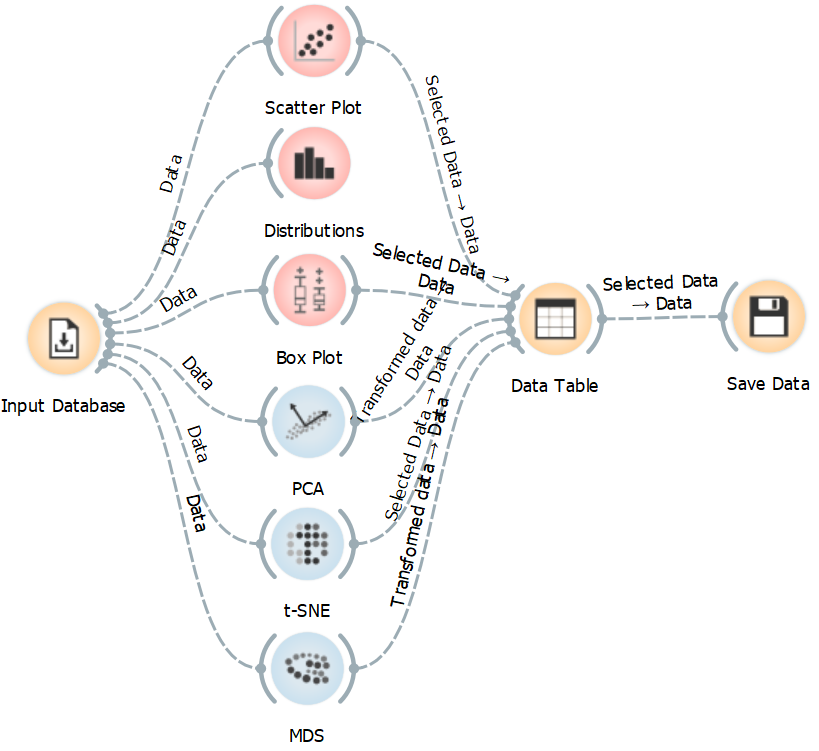}
\caption{Exploratory Data Analysis}
\label{Met_EDA}
\end{figure}

The EDA is a powerful methodology to obtain insights and take actions regarding the database, sometimes means delete features or samples that can harm the model. The figure \ref{Met_EDA} show the main techniques (plots and dimension) used for understanding the data, like atypical values, mean and standard deviation. Firstly, the data is upload in Python\textregistered ~version 3.7 with the Pandas\footnote{Pandas is an open source, BSD-licensed library providing high-performance, easy-to-use data structures and data analysis tools for the Python programming language} library, then is used several packages like Numpy\footnote{NumPy is the fundamental package for scientific computing with Python} and Matplotlib\footnote{Matplotlib is a Python 2D plotting library which produces publication quality figures in a variety of hardcopy formats and interactive environments across platforms} for reading, interpreting and plotting. 

Later, a correlation matrix helped to identify the highly-correlated features. Nonetheless, in the first approach it is used the full database to both, create a base lines models, and comparing them with those ones found on the bibliography. The EIT database possess six possible classification labels, as already mentioned before, hence there will be three set of experiments, converting the six into three-labels and two-labels database. On the other hand, the blood biomarkers database does not need any transformation.

\begin{algorithm}
\caption{Data Engineering Process}\label{MT_alg_1}
\begin{algorithmic}[1]
\Procedure{Data Engineering}{}
\State $\textit{Database} \gets \text{Select which database to use}$
\State $\textit{InputData} \gets \text{Read Excel: }\textit{Database}$
\State $labels \gets 2$
\BState \emph{top}:
\If {$\textit{Database} = EIT$}:
\State $labels \gets \text{Provide \# of labels (2, 3 or 6)}$
\State $\textit{Database} \gets \text{Process a new }\textit{Database}\textit{(labels)}$
\EndIf

\BState \emph{main}:
\State $\texttt{(Please, select one or more enhancement techniques)}$
\If {$\text{Scale Database}\to \textit{True}$}
\State $\textit{Database} \gets \text{Scale (}\textit{Database})$
\EndIf

\If {$\text{Augment Database}\to \textit{True}$}
\State $degree \gets \text{Provide a degree of augmentation (default=2)}$
\State $\textit{Database} \gets \text{Increment \textit{degree} times the }\textit{Database}$
\EndIf

\If {$\text{Expand Database}\to \textit{True}$}
\State $\textit{Database} \gets \text{Add min, max, mean, std as features (}\textit{Database})$
\EndIf

\If {$\text{Add polynomial Database}\to \textit{True}$}
\State $degree \gets \text{Provide a degree of polynomial features (default=2)}$
\State $\textit{Database} \gets \text{Create \textit{degree} poynomial features (}\textit{Database})$
\EndIf
\State \textbf{goto} \emph{Training Algorithm}.
\State \textbf{END}
\EndProcedure
\end{algorithmic}
\end{algorithm}

\subsubsection{Data engineering}

The scarcity of public available databases reduce the ML models to a small number of instances (population), in some cases biasing it. Therefore, it is important to enhance the accessible sets in the finest possible way. Data engineering is a process where a ML specialist can transform an input database into a new one with better and lower/higher number of features, without noise, dealing with unbalance and atypical instances problems. The EIT and blood biomarkers databases maintain similarly the same structure, in consequence the algorithm \ref{MT_alg_1} describe the taken approach for enhance the two datasets. 

The algorithm \ref{MT_alg_1} has three main section, initially, it select which database to process using the Pandas functionality for reading excel files. Then, it is checked whether EIT is the selection, if yes, the process asks to select the number of labels that we are going to use. It is important to recall, that the EIT database has six classes, nevertheless, the classes could be restricted to "cancerous and non cancerous tissue" (it has been seen  better performance with two-classes than six-classes), then, it is necessary to add on the algorithm that "if" statement. Finally, the main section is also divided in four data engineering tactics. 

\begin{figure}[H]
\centering
\includegraphics[width=6in]{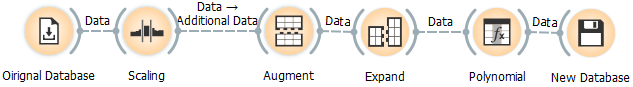} 
\caption{Data Engineering Flow Diagram}
\label{Met_data_eng}
\end{figure}

Firstly, the \textbf{"scale"} function, scale all the data (each feature alone, i.e. each column) with mean 0 and std 1. Secondly, it is possible to create new instances using the initial dataset, hence, the \textbf{"augmentation"} function creates new sets, depending on the augmentation index. To further explain, the algorithm first divide the groups, healthy and sick; then, from one group is created a new instance or "patient", the patient inherits randomly-chosen features from the initial group, reassuring that will be a "unique" instance. Thirdly, the \textbf{"expand"} function aggregates seven new features to each instance, the features are min, max, mean, median, standard deviation, skewness and kurtosis index. 

Lastly, the \textbf{"polynomial"} produce a new set of features from polynomial multiplication between the features, i. e. if a $"X"$ degree is selected in the function, the algorithm will obtain all the possible combinations for multiplications between all the features, but at the same time, satisfying the maximum $"X"$ degree set before. In the EIT and blood biomarkers database, there is the same quantity of features (nine). Therefore, nine features mean 45 combination, plus each feature squared we obtain a total of 54 features per instance. Given these points, in order to obtain and compare the results obtained on the data engineering process, we implement a design of experiments in a grid search (review each possible combination of the 4 methods) for create 16 baseline models, specifically, it is created a set of machine learning models for each possible database e. g. original database or scaled, expanded and augmented database with polynomial features, therefore the figure \ref{Met_data_eng} represent one of the sixteen possible approaches. Quantifiable results are explained in the following chapter, nevertheless hereby is the global approach for data enhancement.

\begin{figure}[H]
\centering
\includegraphics[width=6in]{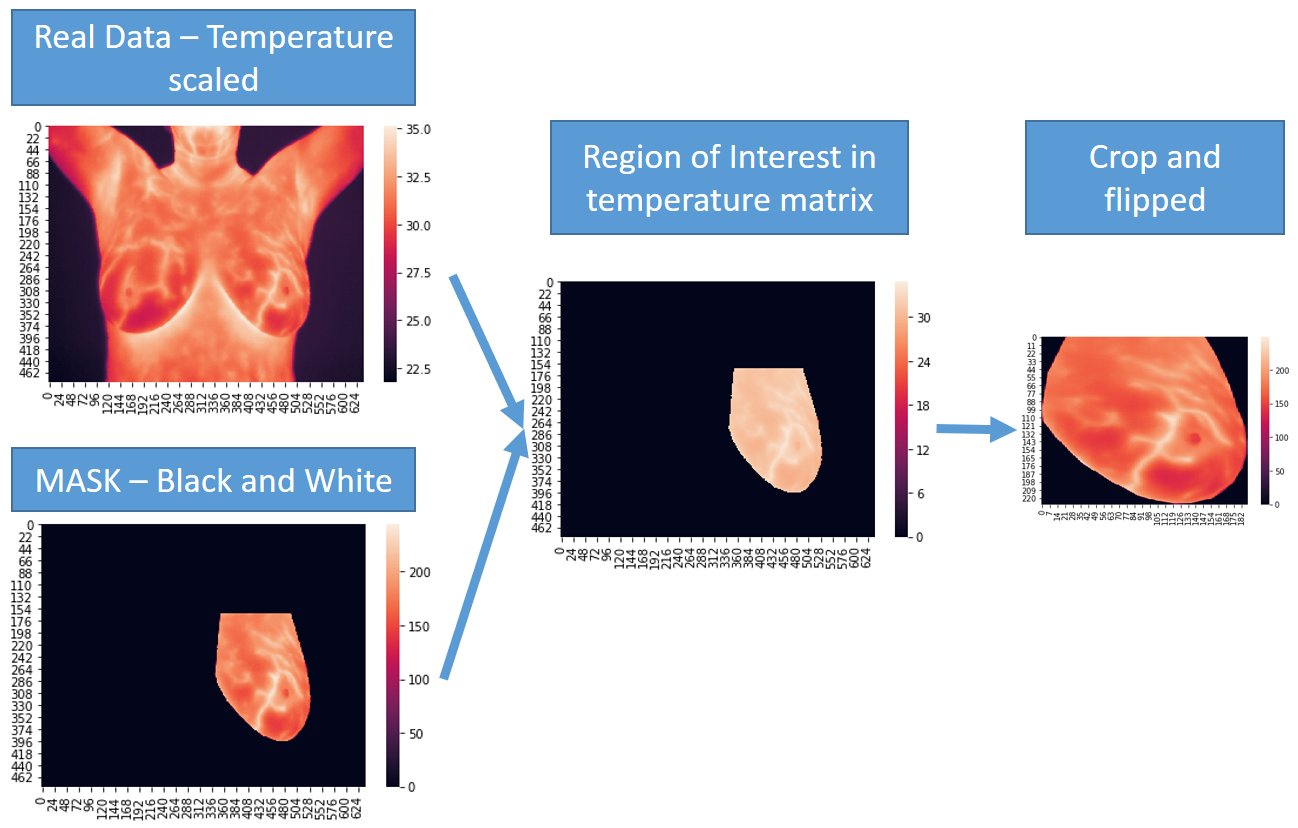} 
\caption{Data engineering flow diagram for the thermal database}
\label{Met_data_eng_thermal}
\end{figure}

\subsection{Thermography}

The thermography database has 56 set of images, 37 suffering breast cancer and 19 healthy cases. Each set has twenty 480x640 images, therefore it is 1140 training samples; each image posses a mask of the breast, which will help to obtain the region of interest. The first step starts with an EDA.    

\subsubsection{Exploratory data analysis}

Similarly to the section's \ref{sub:EDA} EDA methodology (Figure \ref{Met_EDA} and \ref{Met_data_eng}), first, the data is upload in Python\textregistered ~version 3.7 with the Pandas library, then is used Numpy and Matplotlib for reading, interpreting and plotting the main insights from Table \ref{table_databases}. The thermography database, posses two labels "breast cancer" and "healthy" patients, then is not needed a transformation before entering the data engineering phase.

\subsubsection{Data engineering}

Similarly, to the data engineering process presented in section \ref{sub:EDA}, the thermal database has a data correction and engineering phase, where techniques, like data generation and normalization are implemented. The algorithm compile \ref{MT_alg_2} the main idea for the thermal database, this approach \ref{MT_alg_2} has two key section, initially, it read each thermal file (".txt" file), creating a list of arrays, where each element correspond to one subject, and subsequently, each subject has a matrix where each row correspond to one image, and each column correspond to the rolled parameters (converting the 480x640 image into one vector of 307200 temperature points), to put in context there is 1140 samples or temperature points matrices. Afterwards, the main section is also divided in three data engineering techniques, nevertheless the database is scaled before with mean 0 and std 1. 

\begin{algorithm}
\caption{Data Engineering Process for the Thermal database}\label{MT_alg_2}
\begin{algorithmic}[1]
\Procedure{Data Engineering}{}
\State $\textit{InputData} \gets \text{Read ".txt" files: }\textit{Database}$
\BState \emph{main}:
\State $\texttt{(Please, select one or more enhancement techniques)}$
\State $\textit{Database} \gets \text{Scale the input \textit{Database}}$
\If {$\text{Apply mask Database }\to \textit{True}$}
\State $\textit{Mask} \gets \text{Read ".PNG" mask files}$
\State $\textit{ROI} \gets \text{Apply mask files into: }\textit{Database}$
\State $\textit{Database} \gets \text{Allocate \textit{ROI} into variable \textit{Database}}$
\EndIf

\If {$\text{Augment database }\to \textit{True}$}
\State $degree \gets \text{Provide a degree of augmentation (default=2)}$
\State $type \gets \text{Give the type of augmentation (default='flip', 'crop' and 'noise')}$
\State $\textit{Database} \gets \text{Create \textit{degree} times of }\textit{Database} \text{ based on \textit{type}}$
\EndIf

\If {$\text{Normalize database }\to \textit{True}$}
\State $\textit{type} \gets \text{Select a normalization type (e, g \cite{new_thermography_8})}$
\State $\textit{Database} \gets \text{apply \textit{type} into each image in \textit{Database}}$
\EndIf

\State \textbf{goto} \emph{Training Algorithm}.
\State \textbf{END}
\EndProcedure
\end{algorithmic}
\end{algorithm}

The first part of the main section in the algorithm \ref{MT_alg_2}, works as follows: the thermal database has a 480x640 image where the background is set as "black" but the sections including, neck, arm and part of the chest and belly are not. Those areas are completely no necessary for telling whether a person does have cancer, for that reason the implementation of a mask is important to obtain the region of interest, therefore reducing the processing time and increasing the performance of our algorithm. The resulting database is a 1140 instances of 307200 points each one, all in a scaled matrix. Second, it is possible to create new instances using the initial dataset, hence, the augmentation technique output a new whole set. The main known techniques are white noise addition, crop, flip and rotation. 

The leading intention of those types of data enhancement is to create a new database just from the \textbf{training set}, also trying to balance the database. Thirdly, the "normalization" conditional establish a new database depending on a fixed min and max temperature value, the normalization process may vary depending on the technique, for example, Sathish et al. \cite{new_thermography_8} propose a normalization method based on a lookup table. Given these points, in order to obtain and compare the results obtained on the data engineering process, we implement a design of experiments in a grid search (reviewing each possible combination of the 3 methods) for create 9 baseline models, specifically, it is created a set of DNN and CNN models for each possible new dataset e. g. the simplest one, original database or the full enhanced: augmented and normalized database with mask. The Figure \ref{Met_data_eng_thermal} exemplify a fully-enhancement process of a normal image, where the full image is reduced with the mask and then it obtain the ROI for cropping and flipping process. 

\begin{figure}
\centering
\includegraphics[width=5in]{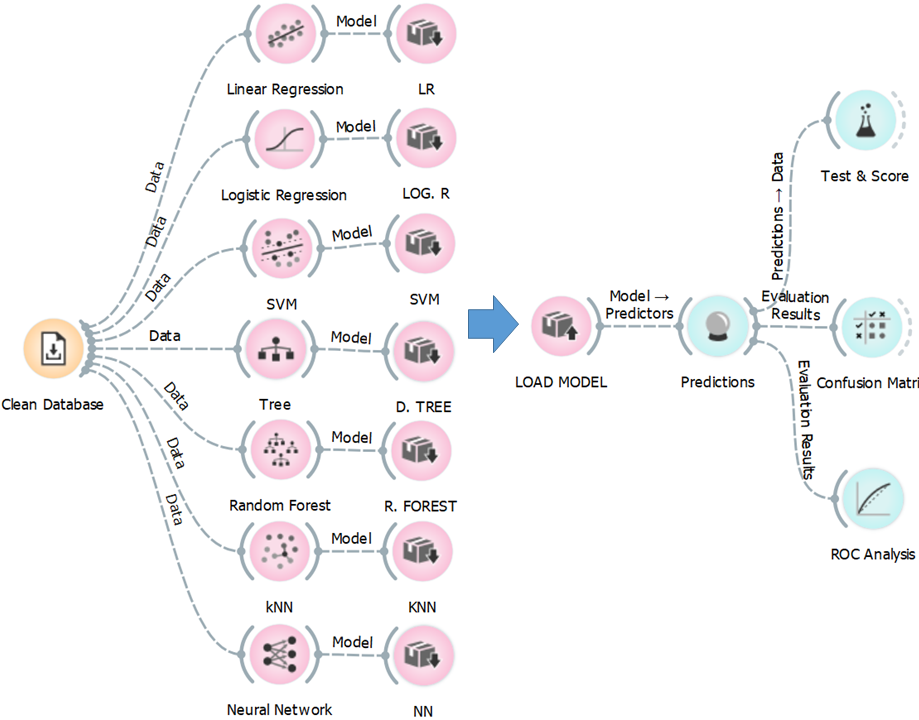} 
\caption{Machine Learning and Evaluation Approach}
\label{Met_train_eval}
\end{figure}

\section{Machine learning techniques}

The section \ref{ch:ML_techniques} went deep concerning types of machine learning techniques, optimization algorithms, dimensionality reduction and so forth. The EIT and blood biomarkers databases due to being categorized as  "stacked databases" should perform better on algorithms like support vector machines or random forest, nonetheless, it has been selected several techniques to obtain baseline models and generate quality data. The training process and evaluation go under the following premises:

\begin{quote}
\begin{itemize}
\item The data has been cleaned and is ready for the ML model,
\item The data is split in train and test set,
\item To avoid overfitting (thus obtaining unbiased models), early stopping is applied. Also, the training set is split in train and validation set,
\item The validation set is 20\% of the train set,
\item The validation set feed the k-fold cross-validation algorithm,
\item After training, each model goes under evaluation with the test set. 
\end{itemize}
\end{quote}

Similarly, the thermal database, posses the same training and evaluation process, nevertheless, the pre-processing and data engineering phases are different, cause the type of data are mainly images and temperature matrices. Generally, ScikitLearn\footnote{Simple and efficient tools for data mining and data analysis. Built on NumPy, SciPy, and matplotlib} \cite{pedregosa2011scikit} is the main library for build the machine learning algorithms in this project, but alongside, TensorFlow\footnote{TensorFlow is a machine learning system that operates at large scale and in heterogeneous environments} \cite{abadi2016tensorflow} package, provide the tools to develop the artificial neural networks, like DNN, CNN and RNN (mainly used for the thermal dataset).

The figure \ref{Met_train_eval} convey the ideas for training and evaluation of the baseline models, but as well of the optimization process. On the other hand, after having the baseline models, several optimized models are created using the HyperOpt package for Python\textregistered ~changing several variables and hyper-parameters, there is further information in the results chapter. Finally, the evaluation processes have several evaluation metrics like: accuracy (acc), precision, sensitivity, specificity, F1 score and the receiver operating characteristic - area under curve (roc-auc), that will allow us to evaluate the models, in an unbiased way, further information is presented in the next section.    

\section{Evaluation metrics}

The performance of an algorithm could be made during or after (prediction) the training steps. Evaluating the model is an essential part from the whole pipeline, cause allow us to determine whether the model needs an improvement or adjustment. Nevertheless, in advance ML algorithms are known that depending on the database some metrics are not useful, why the structure of a database say the type of useful metrics? The following characteristics explain a bit further this assumption:

\begin{itemize}
\item The balance of the database; has it the same number of positive results than negative results?
\item Misclassification in the minor class samples are very high,
\item Clinical databases frequently can not allow certain percentage of TP or TN, therefore the "acuraccy" metric is not tolerable.
\end{itemize}

\begin{figure}[H]
\begin{center}
\caption[Confusion matrix layout]{Representation of a confusion matrix, both explain the distribution in right and wrong classification.}
\label{met_conf_matrix}
\begin{subfigure}{.5\textwidth}
\begin{tikzpicture}[
box/.style={draw,rectangle,minimum size=2cm,text width=1.5cm,align=left}]
\matrix (conmat) [row sep=.1cm,column sep=.1cm] {
\node (tpos) [box,
    label=left:\( \mathbf{p'} \),
    label=above:\( \mathbf{p} \),
    ] {True \\ positive};
&
\node (fneg) [box,
    label=above:\textbf{n},
    label=above right:\textbf{total},
    label=right:\( \mathrm{P}' \)] {False \\ negative};
\\
\node (fpos) [box,
    label=left:\( \mathbf{n'} \),
    label=below left:\textbf{total},
    label=below:P] {False \\ positive};
&
\node (tneg) [box,
    label=right:\( \mathrm{N}' \),
    label=below:N] {True \\ negative};
\\
};
\node [left=.05cm of conmat,text width=1.5cm,align=right] {\textbf{Real value}};
\node [above=.05cm of conmat] {\textbf{Prediction outcome}};
\end{tikzpicture}
  \caption{Confusion Matrix.}
  \label{met_conf_matrix_1}
\end{subfigure}%
\begin{subfigure}{.5\textwidth}
  \centering
  \includegraphics[width=1.9in]{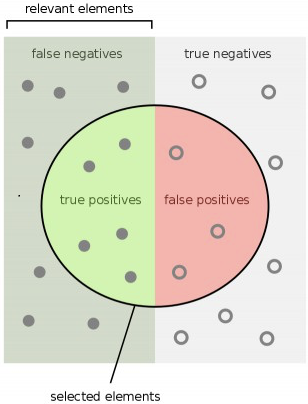}
  \caption{Graphical representation of a confusion matrix.}
  \label{met_conf_matrix_2}
\end{subfigure}
\end{center}
\end{figure}
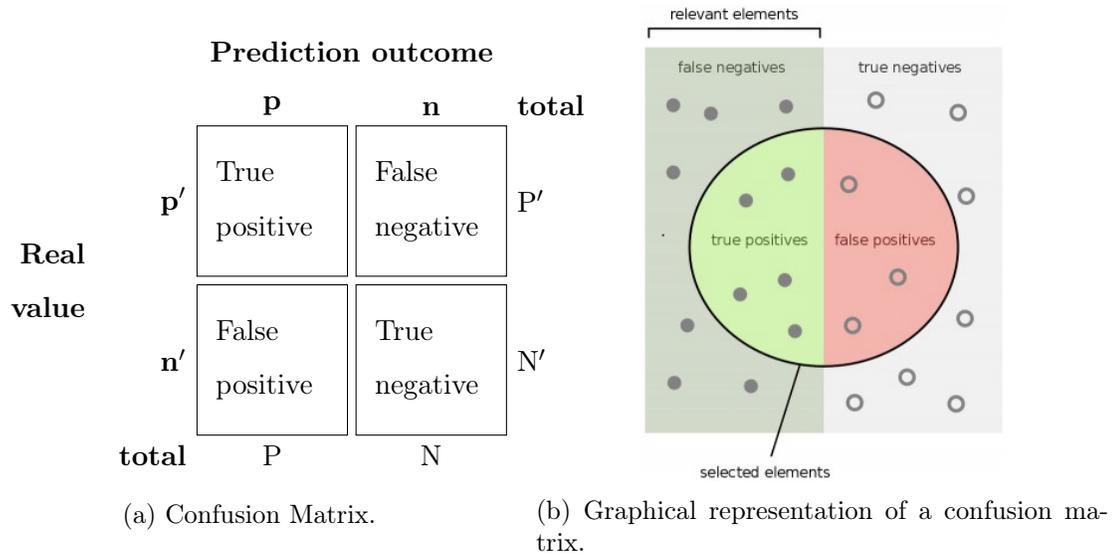

Therefore, the following equations are employed during this study for measure the performance of each algorithm, 

\begin{equation}
Accuracy~=~\dfrac{\#~of~correct~predictions}{Total~\#~of~predictions}
\label{met_accuracy}
\end{equation}

The equation (\ref{met_accuracy}) usually gives a significant result when the number of samples per class is equal over the whole database. On the other hand, the confusion matrix is a matrix where is described the whole performance of the model. The confusion matrix from Figure \ref{met_conf_matrix_1} is made of 4 cells, where each cell represents a specific metric. The following list, will define each cell within the "clinical" databases used on this study: 

\begin{itemize}
\item \textbf{True positives}: it predicted "breast cancer" and the patient is having "breast cancer", 
\item \textbf{True negatives}: it predicted "healthy" and the patient is "healthy",
\item \textbf{False positives}: it predicted "breast cancer" and the patient is "healthy",
\item \textbf{False negatives}: it predicted "healthy" and the patient has "breast cancer".
\end{itemize}

Accordingly, from the above terms and the Figure \ref{met_conf_matrix} it is possible to define metrics that could reduce the bias in "metrics" when we have unbalance databases.

\begin{equation}
True~Positive~Rate~=~\dfrac{True~Positive}{False~Negative~+~True~Positive}
\label{met_TPR}
\end{equation}

The equation (\ref{met_TPR}) or \textbf{Sensitivity} corresponds to the fraction of positive samples that are correctly predicted as positive, with respect to the whole positive data samples. 

\begin{equation}
True~Negative~Rate=~\dfrac{True~Negative}{False~Positive~+~True~Negative}
\label{met_FPR}
\end{equation}

The equation (\ref{met_FPR}) or \textbf{Specificity} corresponds to the fraction of negative samples that are correctly predicted as negative, with respect to the whole negative data samples. 

\begin{equation}
Positive~Predictive~Value=~\dfrac{True~Positive}{True~Positive~+~False~Positive}
\label{met_PPV}
\end{equation}

\begin{equation}
Negative~Predictive~Value=~\dfrac{True~Negative}{True~Negative~+~False~Negative}
\label{met_NPV}
\end{equation}

The equation (\ref{met_PPV}) and (\ref{met_NPV}) stand as positive and negative predictive value, respectively. They are the proportions of positive and negative values that are true positive and true negative results, respectively. Likewise, the F1 score equation (\ref{met_f1_score}) stands as the harmonic mean of precision (PPV) and recall, hence, it will be used for scoring the global performance. 

\begin{equation}
F1~score=~\dfrac{2 \cdot True~Positive}{2 \cdot True~Positive~+~False~Positive~+~False~Negative}
\label{met_f1_score}
\end{equation}

\begin{figure}[H]
\centering
\includegraphics[width=6in]{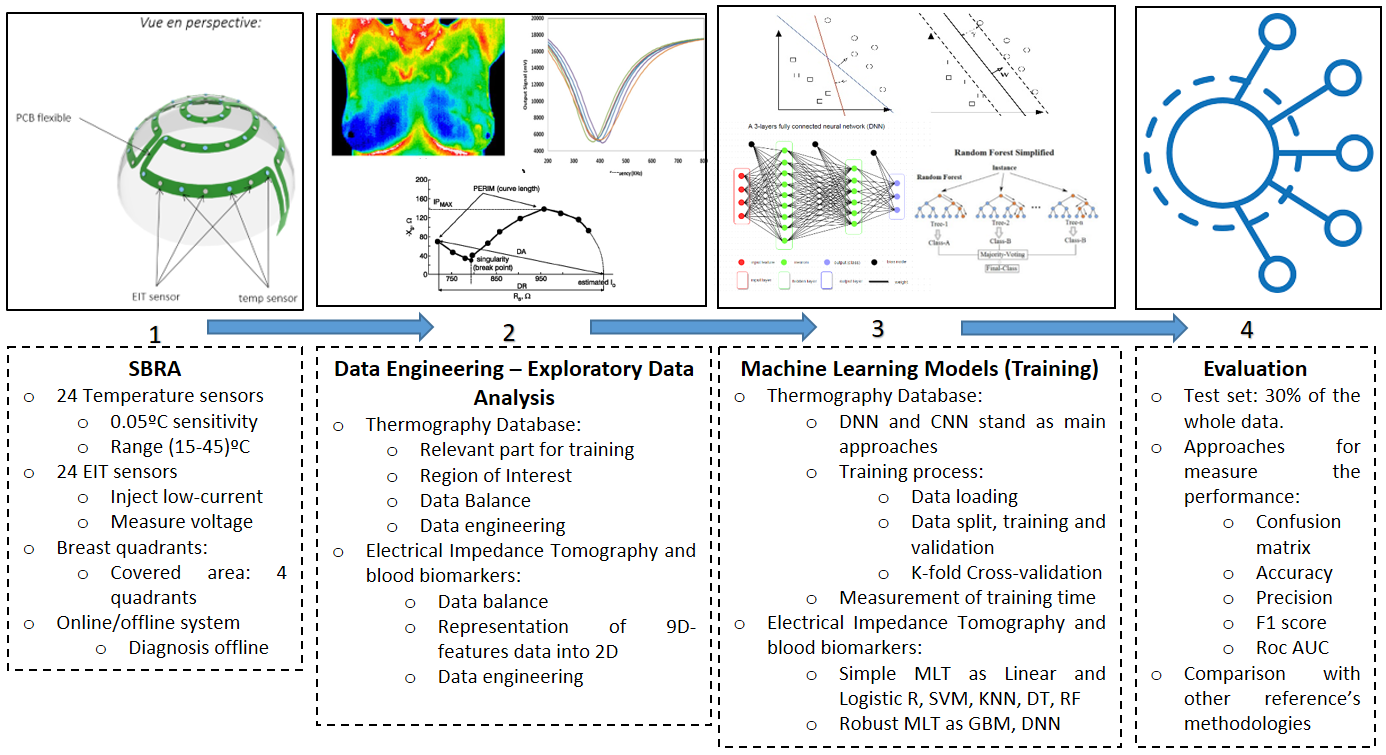} 
\caption[Global system's pipeline for the diagnosis of breast cancer]{Global system's pipeline for the diagnosis of breast cancer. This study covers from (2) to (4). (1) makes part of the global project, SBRA but won't be taken into consideration.}
\label{Met_global_system}
\end{figure}

Finally, the equations above mentioned are for 2 label classification, nonetheless with some modifications can be used for multi-class (more than 2 classes) classification. Those metrics will help to compare our results with the current ones in the bibliography. Also, this is the last step of our approach, where is obtained the performance of both, the baseline and the optimize models. In conclusion, this methodology makes part of the SBRA project, therefore, the Figure \ref{Met_global_system} illustrate the global approach and the project's key phases. The next chapter will describe the results from each type of breast cancer database.
\chapter{Results}
\label{ch:results}

This chapter comments and highlights the results achieved over the three databases. The results are based on the databases explained in Chapter \ref{ch:introduction}, likewise a further introduction in those topics are in Chapter \ref{ch:background}, then the methodology is in Chapter \ref{ch:methodology}. First, we discuss the baseline models and the performance with several types of metrics. Second, the results of each database are presented into different sections, ranging from section \ref{ch:results_blood} to \ref{ch:results_other}. The following questions will be answered in order to assess the questions in the statement of the problem. 

\begin{itemize}
\item What were the best machine learning technique in each database?
\item What are the benefits and drawbacks of database enhancement?
\item Which are the benefits in hyper-parameters optimization?
\item Which parameters were the most significant over the optimization process?
\item Is the performance increased when using optimization?
\item Which type of metrics were used to evaluate the algorithm's performance?
\item Is it possible to create a pipeline feed by the three databases simultaneously?
\end{itemize}

The coming sections are focused on the results of each database. Firstly, the maximum accuracy achieved in the blood's database during the baseline models generation we achieved an accuracy of 93\% with an XGboosting regressor and in the EIT database 94\% with XGBoosting as well. On the other hand

It is important to mention that each database possess their own structure, therefore, the MLT approach may differ.  

\section{Blood biomarkers}
\label{ch:results_blood}

The blood biomarkers database has nine features, where two are the age and body mass index, and seven are related with biomarkers found in the blood. The first step made for develop a machine learning algorithm to diagnostic breast cancer was the exploratory data analysis, here we delve into non visible components, such as correlation between features, atypical and irrelevant instances.

\subsection{Exploratory data analysis}

\begin{figure}[H]
\centering
\includegraphics[width=6.5in]{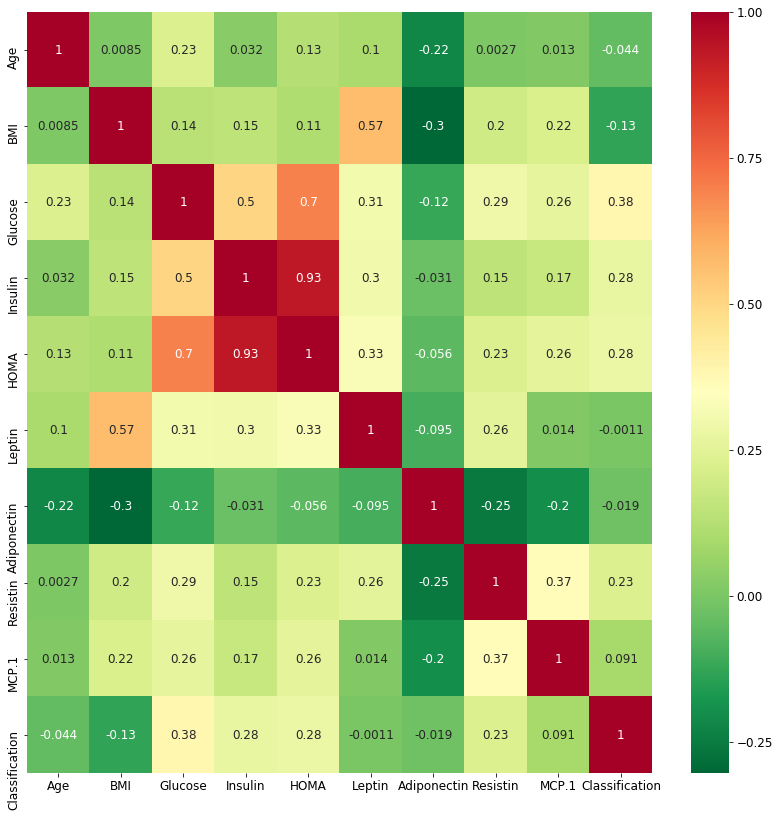} 
\caption[Pearson correlation plot of the blood biomarkers database]{Pearson correlation plot of the nine-features blood biomarkers database, the correlation rate range between -1 and 1.}
\label{res_correlation}
\end{figure}

The initial insights we obtained from the EDA were that pair of features such as insulin-HOMA, glucose-HOMA, glucose-insulin, leption-BMI and glucose-resistin are highly correlate thus, indicating that a feature augmentation can be made in this database in order to look up for quadratic correlation between features. The Figure \ref{res_correlation} (appendix \ref{appendix:fig}) displays the Pearson correlation plot between all the features from the blood database, and also the correlation between each feature and the classification labeling, specifically, glucose, insulin, HOMA and resistin with correlation index 0.38, 0.28, 0.28 and 0.23, respectively. Afterwards finding the high correlated features, we made a plot pairwise that exhibit the relationships between the features of a given dataset, however, we have plotted the age, BMI, glucose, insulin, resistin and MCP.1. The Figure \ref{res_blood_pair} in the appendix \ref{appendix:fig} display the relation between each of the most correlated features from Figure \ref{res_correlation}, this plot has a grid of Axes such that each variable in data will by shared in the y-axis across a single row and in the x-axis across a single column. The diagonal Axes are evaluated differently, drawing a plot to show the univariate distribution for both labels (cancer and healthy) of the data for the variable in that column. As a result, of the correlation and pairwise plots (Figure \ref{res_correlation} and \ref{res_blood_pair}) we concluded that was not possible to separate linearly the given features. Henceforth, several types of dimensionality reduction were implemented (see Chapter \ref{ch:background}) for find the distribution of the plots. It has been chosen diverse techniques such as, MDS, t-SNE, isomap, locally liner embedding, PCA and spectral embedding for transform the 9D database into 2D. PCA, spectral embedding and isomap, show a better-distributed map than the other ones. Indeed, clearly it can be seen on PCA and isomap scatter plots, that there is a trend of atypical values on people does have cancer, most compelling evidence is in Figure \ref{res_DR} in the appendix \ref{appendix:fig} of this report. The x-axis and y-axis represent the first and second component, respectively, in the PCA plot from the Figure \ref{res_DR} can be interpreted that the breast cancer patients have atypical values, those could due to unhealthy lifestyle or high variations on insulin, MCP.1 or HOMA (see the excel file on the appendix). Given the above points, we collected enough insights about the database, therefore we may continue to the machine learning algorithms.

\subsection{Baseline models}

\begin{figure}[H]
\centering
\includegraphics[width=6in]{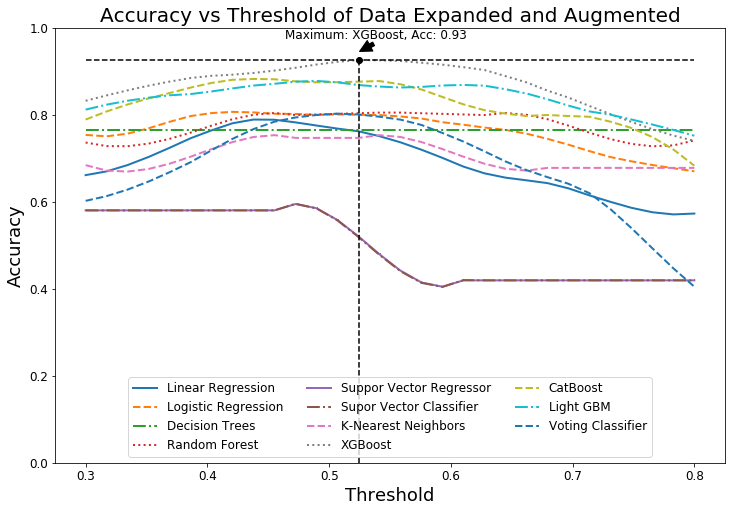} 
\caption[Performance versus threshold in the augmented and expanded database]{Performance versus threshold in the augmented and expanded database. The leading model was a XGBoosting regressor.  Model's accuracy, precision and recall of 93\%, 96\% and 92\%, respectively.}
\label{res_DoE_3}
\end{figure}

In this section, we explain the building, training and performance of the ML models. After the EDA we realized that we should apply a \textit{\textbf{"Design of Experiments"}} or DOE with database transformation and enhancement. As explained in Chapter \ref{ch:methodology} we propose four types of data enhancement: scaling, augmentation, expansion and polynomial features inclusion, further details are in the corresponding chapter (see algorithm \ref{MT_alg_1}). The DOE was composed by sixteen different test, going from "original database" until "data scaled, expanded, augmented and polynomial features database", those were created using the \textit{"itertools"} package from Python. The augmentation degree was four, that means we created three new set of instances from the initial one, similarly, the polynomial augmentation had a second degree as discussed in the Chapter \ref{ch:methodology}. A "for" loop goes from the first till the last experiment. Each experiment works as follows: firstly, a function get as input a boolean variable i. e. true or false, depending on that, the new and enhanced database is created. Secondly, it is split the database into train and test at 70/30. Thirdly, again a function build and train the models based on the default parameters that provide python, to emphasize, the MLT were linear and logistic regression, decision trees, random forest, support vector machines, XGBoost, Catboost and LGBoost machines. Fourthly, all the models gave a quantitative value between 0 and 1, therefore, we constructed a matrix of evaluation metrics like accuracy, precision, recall, F1 score and ROC AUC. Finally, we plotted the models' performance versus the threshold with the achieved metrics. The Figure \ref{res_DoE_3} display the best DOE's performance for the expanded and augmented database's test.

\begin{figure}[H]
\centering
\includegraphics[width=5.25in]{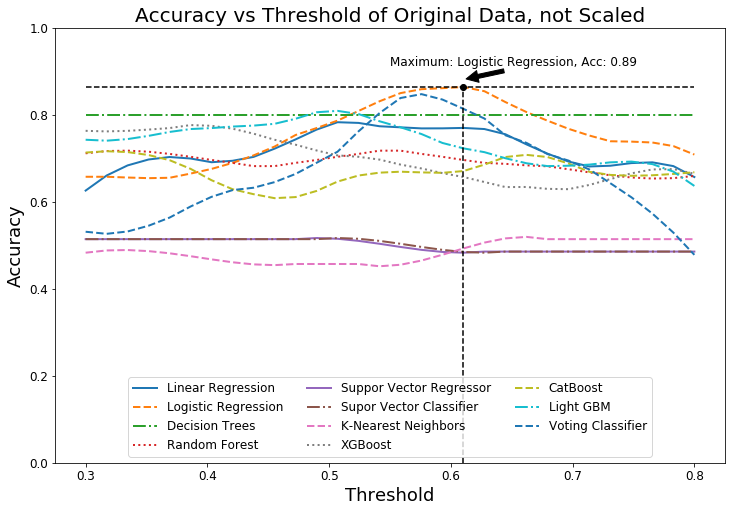} 
\caption[Performance plot in the blood biomarkers' original database]{Performance versus threshold in the blood biomarkers' original database. Model's accuracy, precision and recall of 89\%, 100\% and 78\%, respectively.}
\label{res_DoE_1}
\end{figure}

\begin{figure}[H]
\centering
\includegraphics[width=5.25in]{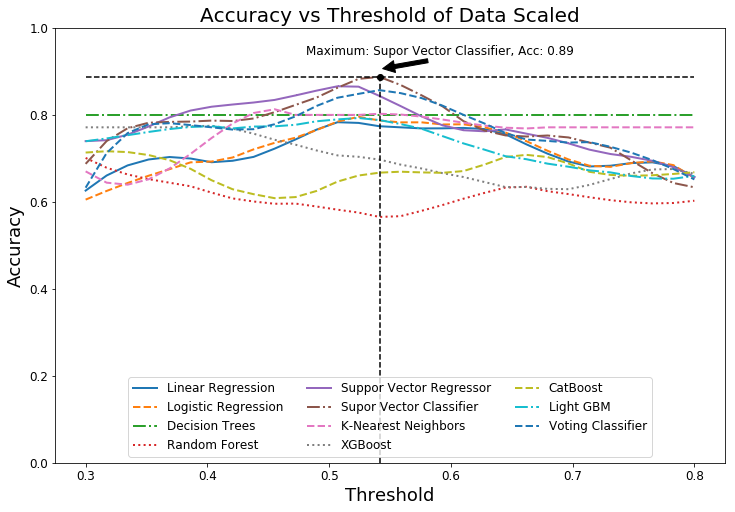} 
\caption[Performance plot in the blood biomarkers' scaled database]{Performance versus threshold in the blood biomarkers' scaled database. The leading model was a support vector classifier. Model's accuracy, precision and recall of 89\%, 89\% and 89\%, respectively.}
\label{res_DoE_2}
\end{figure}

\begin{figure}[H]
\centering
\includegraphics[width=5.25in]{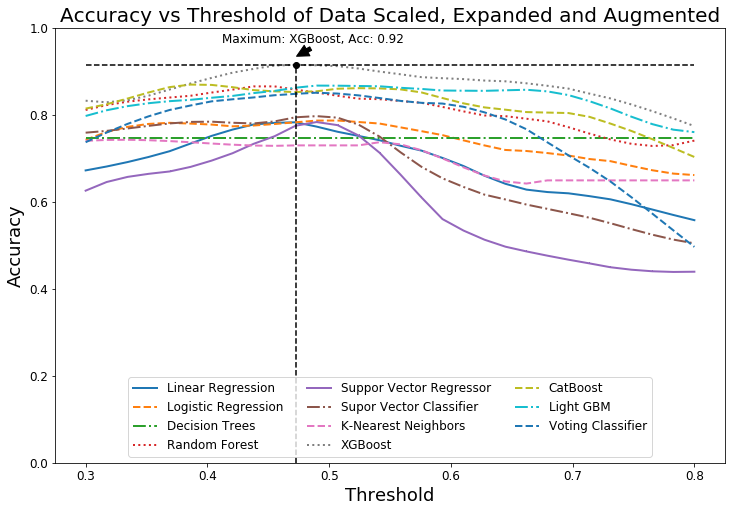} 
\caption[Performance plot in the blood biomarkers' scaled, expanded and augmented database]{Performance versus threshold in the blood biomarkers' scaled, expanded and augmented database. The leading model was a XGBoosting regressor.  Model's accuracy, precision and recall of 92\%, 89\% and 98\%, respectively.}
\label{res_DoE_4}
\end{figure}

The finest model was a XgBoosting regressor which is based on a \textbf{"scikit-learn" library}. Nonetheless, in the appendix \ref{appendix:fig} there are the plots (Figure \ref{res_DoE_1}, \ref{res_DoE_2} and \ref{res_DoE_4}) regarding the top 3 performance from the DOE's experiments. Likewise, the Table \ref{resul_Doe_blood} from the appendix \ref{appendix:tables} gives extra metrics of the evaluation process. It is important to recall, that some metrics are biased cause the number of positive and negative cases of breast cancer are not enough. On the other hand, in the topic of breast cancer diagnosis, it is much more threatening to have a high rate of false negative (FN) than false positive (FP), hence, we should vary the threshold to reduce the rate FN rising the FP rate. 

Differently, the artificial neural networks (ANN) approach, like DNN yielded unsatisfactory results. The first step was to create a parameters space containing: number of neurons and layers, batch size and normalization, learning rate, activation function and dropout rate. Secondly, a random search was implemented using K-fold cross-validation with three validation folds and 1000 iterations. Despite the number of iterations, it was not possible to obtain a successful model, to put briefly, we obtain a maximum accuracy of 75\%, recall, precision and F1 score of 56\%, 90.1\% and 69\%, respectively. The top achieved performance was with batch normalization rate and size of 0.9 and 30, respectively, also dropout and learning rate of 0.3 and 0.01, respectively. Finally, the number of hidden layers and neurons per layers were 4 and 90, respectively. Given those assumptions, it is not necessary to give further information about ANN models. 
 
\subsection{Hyper-parameters optimization}

\begin{figure}[H]
\centering
\includegraphics[width=6in]{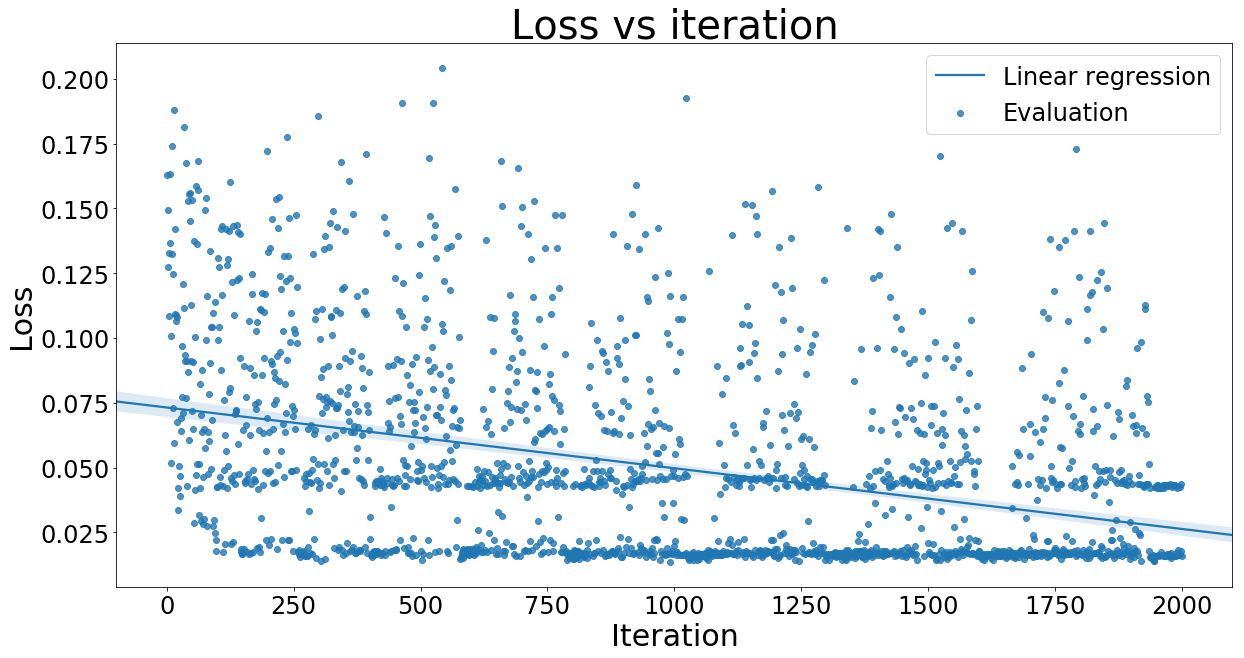} 
\caption[Scatter plot with linear regression of the PTE algorithm for two thousand iterations]{Scatter plot with linear regression of the PTE algorithm for two thousand iterations. Objective function: light gradient boosting machines.}
\label{res_blood_losses}
\end{figure}

In this section we will review the main outcomes in the hyper-parameters optimization, due to the long processing time of the "catboost" algorithm, we decided to discard it as a technique for the optimization process. Likewise, the hyper-parameters optimization for XgBoosting and light gradient boosting machines or LGBM was made with the \textit{\textbf{HyperOPT}} Python's library, that uses bayesian optimization (Tree Parzen Estimator) in a search space for evaluate a specific model. Both MLT are based in a\textbf{"scikit-learn" library}, nevertheless each one has their own tunable parameters. The blood biomarkers database is considered as "stacked database", therefore, XGBoost and LGBM are the two top ML techniques for the training process inside the objective function. The Chapter \ref{ch:background} explains the key considerations for use the HyperOpt library. First, we design an objective function for each technique, also the search space differs (hyper-parameters). Second, depending in the number of iterations, the algorithm tries to rise the performance evaluating each time the objective function with a whole new set of hyper-parameters, performing 3-fold cross-validation to decrease the overfitting and making it a robust algorithm. Third, the evaluation function receives a Pandas' Data Frame\footnote{Two-dimensional size-mutable, potentially heterogeneous tabular data structure with labeled axes (rows and columns)} as an input, containing the top number of models; then, the function create and train a new model with the desired parameters, finally it returns a data frame that has the metrics (based on new instances, the test set) of the top model's estimation. The first attempt of optimization was with an objective function, which has LGBM as main technique. We obtained an accuracy improvement using TPE - hyper-parameters optimization of 10\%, in overall we obtained an accuracy of 98.276\%. The Figure \ref{res_blood_losses} display the algorithm loss' behavior with two thousand iterations or epochs. 

Further information about the optimization process can be found in the appendix \ref{appendix:fig}, Figure \ref{res_blood_PTE_LGBM}. Hereby, it is feasible to find the performance of the optimizer in each epochs, and their behavior in each hyper-parameter, the listed parameters are learning, alpha and lambda regulation rates, number of estimators, subsample for bin, bagging fraction and frequency and minimum number of child samples per leaf. Despite the number of epochs, the algorithm always try to select the best set of hyper-parameters for the next epoch. The orange stars in the plots indicate the top 10 performances. 

\begin{table}[htb]
\caption[Metrics for the augmented and expanded database]{Metrics for the augmented and expanded database. Each phase has both techniques, XGBoosting and LGBM; it has five metrics (accuracy, precision, recall, F1 score and ROC AUC) and the execution time per each technique.}
\label{resul_blood_PTE_tbl}
\begin{tabular}{@{}lcccccc@{}}
\toprule
                    & \multicolumn{6}{l}{Phase 1: Baseline models}                                       \\ \cmidrule{2-7}
Model               & Accuracy & Precision & Recall & F1 score & ROC AUC & Time (ms) \\ \midrule
XGB Regressor       & 0.93     & 0.96      & 0.92   & 0.94     & 0.93	& 9.37      \\
LGBM                & 0.88     & 0.86      & 0.94   & 0.90     & 0.87	& 6.25      \\ \bottomrule
                    & \multicolumn{6}{l}{Phase 2: Parzen tree with bayesian optimization (2000 epochs)}                                       \\ \cmidrule{2-7}
Model               & Accuracy & Precision & Recall & F1 score & ROC AUC & Time (ms) \\ \midrule
XGB Regressor       & 0.99     & 0.98      & 1.00   & 0.99     & 0.99	& 9.37      \\
LGBM                & 0.98     & 0.98      & 0.99   & 0.98     & 0.98	& 346.8      \\ \bottomrule
\end{tabular}
\end{table}

Moreover, the bayesian optimization using XGBoosting as objective function was similar to the LGBM one. The only differences were the hyper-parameters set and search space. The Figure \ref{res_blood_PTE_XGBM} in appendix \ref{appendix:fig} exhibit the bayesian performance through two thousand epochs. The hyper-parameters space has ten variables such as, eta, gamma and learning rate, max tree deepness, lambda and alpha regulation rates, max number of leaves (max number of nodes to be added in each split) and the subsample ratio of columns when constructing each tree (\textit{colsample\_bytree}). What we achieved after the bayesian optimization so far, regarding accuracy, precision and recall were 99\%, 98\% and 100\%, respectively. To summarize, the Table \ref{resul_blood_PTE_tbl} has the main metrics about each algorithm during the two phases, before and after the bayesian optimization. It is important to recall that firstly, XGBoosting is the best MLT for this database, secondly, some types of data enhancement don't worked as we expected, maybe due to the redundancy or combination of features e. g. polynomial features and expansion. Thirdly, the number of estimators in LGBM, eta and gamma in XGBM and learning rate in both techniques, were the most significant parameters. Finally, we achieved a huge increase in the performance when bayesian optimization was used, reducing the FN and FP rates.

\section{Electrical impedance tomography}
\label{ch:results_EIT}

The electrical impedance tomography database has also nine features, all measured from breast tissue biopsies. It is mandatory  to remember that the aforementioned dataset is extremely similar to the blood dataset; therefore, the majority of results are quite similar. Indeed, the algorithms attached to this report in the annex are alike, with a minor differences. The first step towards a ML algorithm consists in the exploratory data analysis; here we examine the main dataset's characteristics, like correlation between features, atypical and irrelevant instances, likewise the opportunity to reduce this multi-class (6 classes) into a bi-class dataset. 

\subsection{Exploratory data analysis}

\begin{figure}[H]
\centering
\includegraphics[width=6.5in]{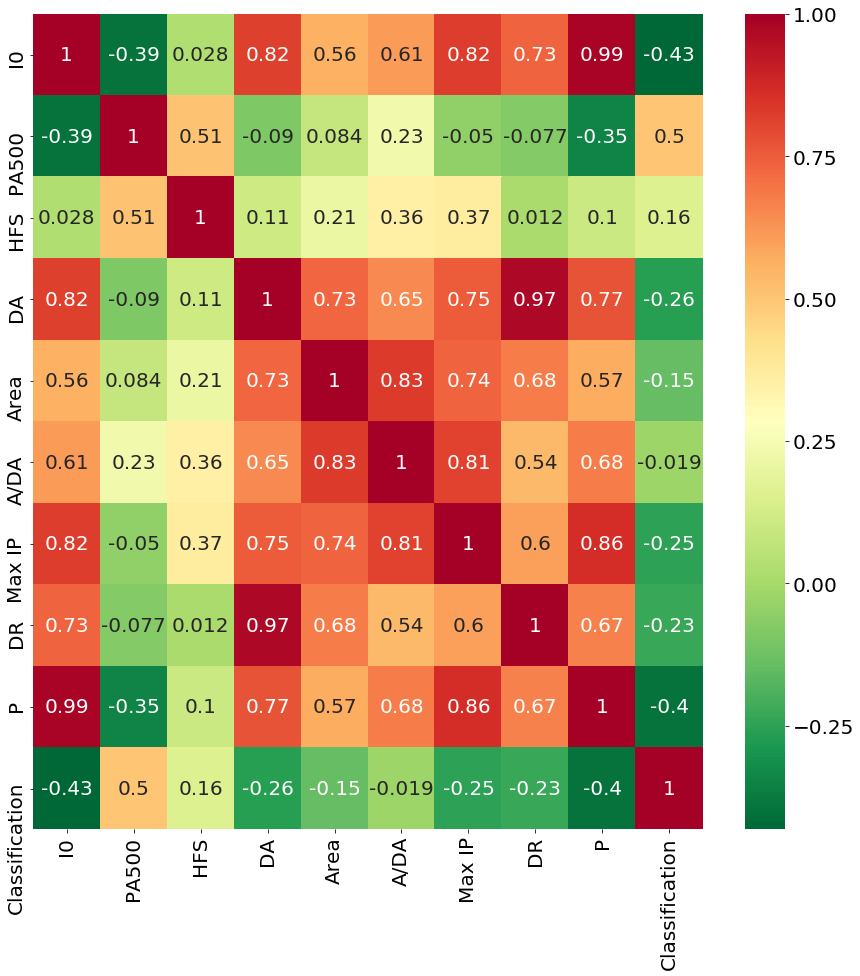} 
\caption[Pearson correlation plot for the electrical impedance tomography's database]{Pearson correlation plot of the nine-features electrical impedance tomography's database, the correlation rate range between -1 and 1.}
\label{res_correlation_eit}
\end{figure}

The EIT's EDA has similar structure in comparison with the blood biomarkers one. The nine-features database possess seven highly correlated features, we consider those as by-products, because they were calculated using either, physical or electrical formulas, therefore the correlation could be linear or non-linear. The Figure \ref{res_correlation_eit} (appendix \ref{appendix:fig}) displays the Pearson correlation plot between all the features from the EIT's database, and also the correlation between each feature and the classification label, specifically, \textit{PA500} and \textit{HFS} are the higher positive correlated features (0.5 and 0.16, respectively), contrary \textit{I0} and \textit{P} are the higher negative correlated (-0.43 and -0.4, respectively). Further information, regarding this database could be found in Chapter  \ref{ch:background} and \ref{ch:methodology}. It is important to see the relation and distribution between all the key features, therefore the Figure \ref{res_eit_pair} in the appendix \ref{appendix:fig} display 2D plots between each of the most correlated features from Figure \ref{res_correlation_eit}. The whole figure has a grid of Axes such that each variable in data will by shared in the y-axis across a single row and in the x-axis across a single column. The diagonal Axes are evaluated differently, drawing a plot to show the univariate distribution for both labels (cancerous tissue or other type of tissue) of the data for the variable in that column. As a result of the correlation plots (Figure \ref{res_correlation_eit}) we concluded that was not possible to separate linearly the given features with six-class labeling. Nonetheless, as was suggested by the following studies \cite{eit_1, eit_2}, we transformed the six-dimensional database into two-dimensional, the process has further information in Chapter \ref{ch:methodology}. 

Another key part of our EDA after the transformation into two-classes, was the dimensionality reduction for defining whether it is or isn't chance to linearly separate the database. Thus, it has been chosen diverse techniques such as, MDS, t-SNE, isomap, locally liner embedding, PCA and spectral embedding for transform the 9D database into 2D with just two-classes labeling. t-SNE and PCA show a clearly separable map than the other techniques, the Figure \ref{res_EIT} in the appendix \ref{appendix:fig}, summarize the scatter plots.

\subsection{Baseline models}

\begin{figure}[H]
\centering
\includegraphics[width=5.5in]{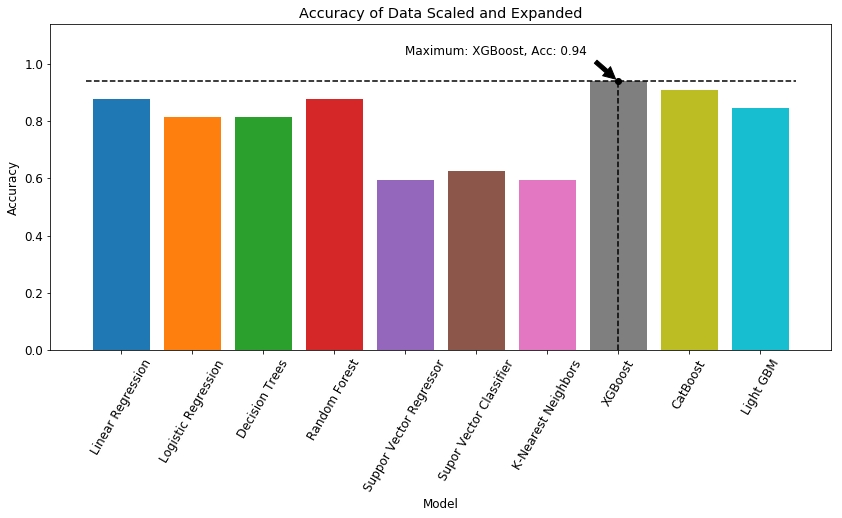} 
\caption[Performance versus threshold in the EIT's scaled, augmented and expanded database]{Performance versus threshold in the EIT's scaled, augmented and expanded database. The leading model was a XGBoosting regressor.  Model's accuracy, precision and recall of 94\%, 100\% and 83\%, respectively.}
\label{res_DoE_EIT_3}
\end{figure}

The DOE implemented here shares the same structure and data enhancement that blood biomarkers does. Similarly, in this section, we explain the building, training and performance of the ML models; we applied four types of data enhancement: scaling, augmentation, expansion and polynomial features inclusion. Likewise, the DOE has sixteen test, going from "original database" until "data scaled, expanded, augmented and polynomial features database", those were created using the \textit{"itertools"} package from Python. The EIT's algorithm for baseline models building, training and evaluation follows the same rules as was explained in the blood biomarkers baseline models did. The Figure \ref{res_DoE_EIT_3} display the best DOE's performance for the scaled, expanded and augmented database's test set.

\begin{figure}[H]
\centering
\includegraphics[width=5.5in]{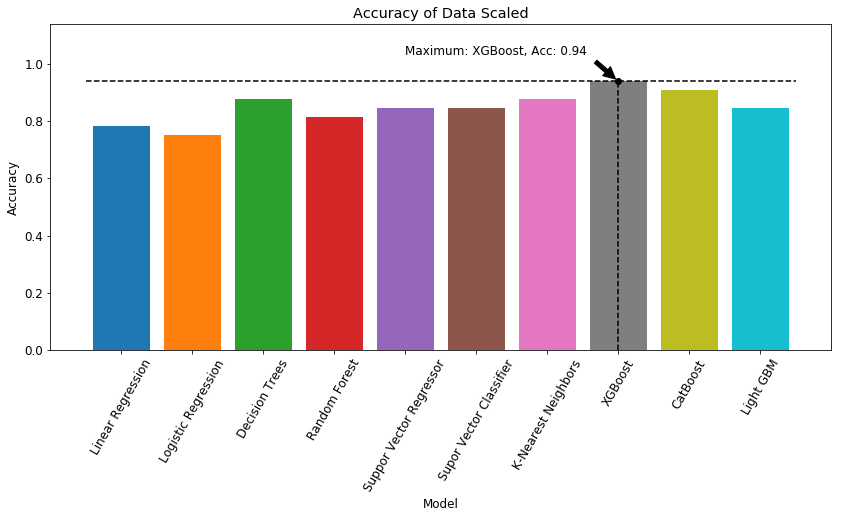} 
\caption[Performance versus threshold in the EIT's scaled database]{Performance versus threshold in the EIT's scaled database. The leading model was XGBoosting regressor. Model's accuracy, precision and recall of 94\%, 92\% and 92\%, respectively.}
\label{res_DoE_EIT_2}
\end{figure}

\begin{figure}[H]
\centering
\includegraphics[width=5.5in]{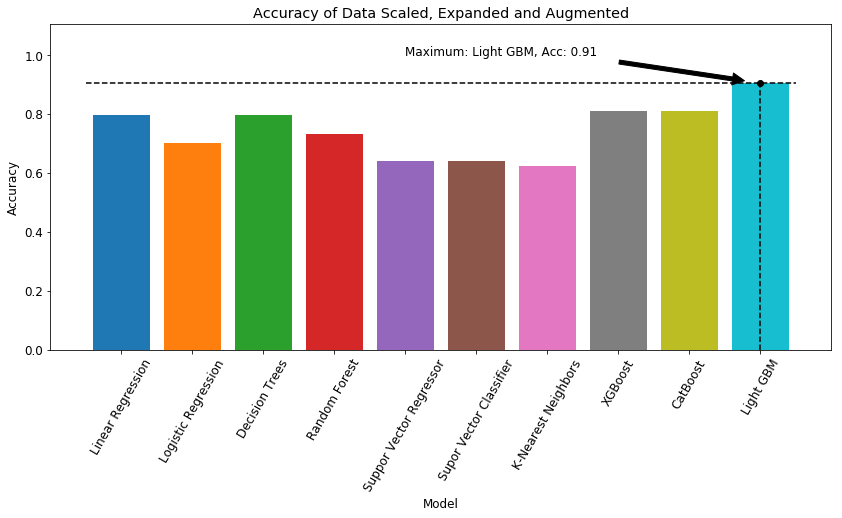} 
\caption[Performance versus threshold in the EIT's scaled, expanded and augmented database]{Performance versus threshold in the EIT's scaled, expanded and augmented database. The leading model was a LGBM. Model's accuracy, precision and recall of 91\%, 90\% and 82\%, respectively.}
\label{res_DoE_EIT_4}
\end{figure}

The finest model was a XgBoosting regressor which is based on a \textbf{"scikit-learn" library}. Nonetheless, in the appendix \ref{appendix:fig} there are plots (Figure \ref{res_DoE_EIT_1}, \ref{res_DoE_EIT_2} and \ref{res_DoE_EIT_4}) regarding the top 3 performance from the DOE's experiments. Likewise, the Table \ref{resul_Doe_EIT} from the appendix \ref{appendix:tables} gives extra metrics of the evaluation process. Alike blood biomarkers' database, the ANN approach was not strong enough (accuracy: <70\%), even applying random hyper-parameters optimization.  

\subsection{Hyper-parameters optimization}

\begin{figure}[H]
\centering
\includegraphics[width=5.5in]{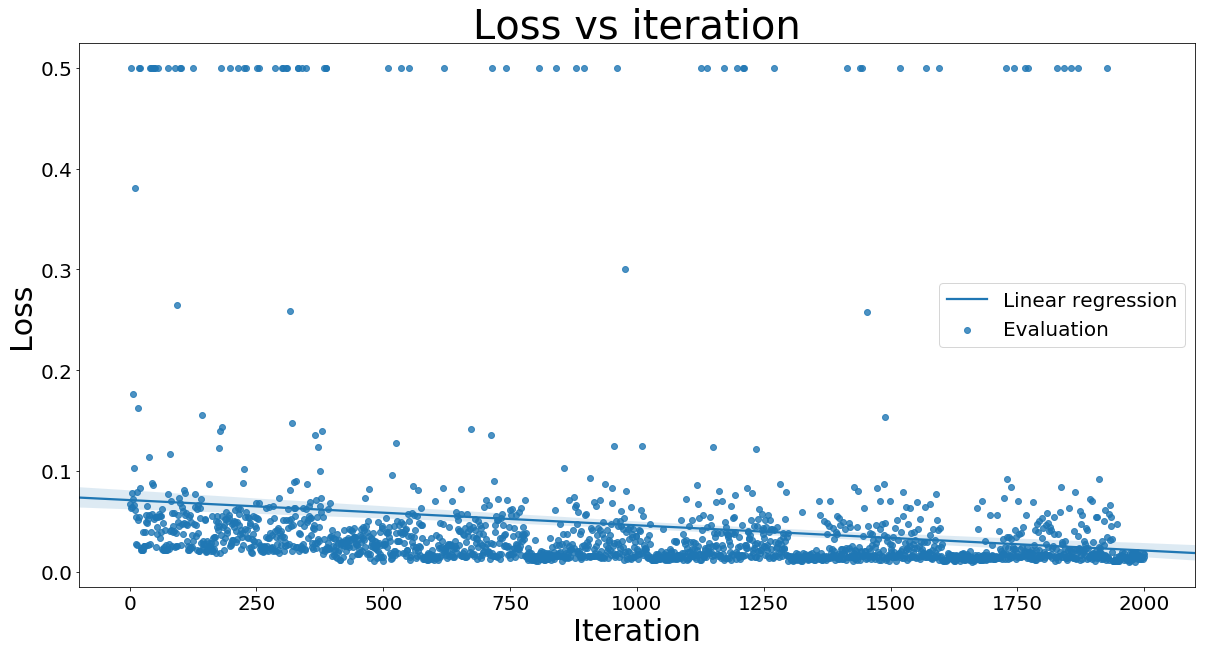} 
\caption[Scatter plot with linear regression of the PTE algorithm for two thousand iterations]{Scatter plot with linear regression of the PTE algorithm for two thousand iterations. Objective function: light gradient boosting machines for EIT's database.}
\label{res_eit_losses}
\end{figure}

In this section we will review the main outcomes in the hyper-parameters optimization, due to the long processing time of the "catboost" algorithm, we decided to discard it as a technique for the optimization process. This process is similar to the hyper-parameters optimization in the blood biomarkers database, in fact, it is composed of four phases, and it uses the parzen tree optimizer again. We obtained an accuracy improvement using TPE - hyper-parameters optimization for LGBM of 6.79\%, in overall we obtained an accuracy of 9.79\%. The Figure \ref{res_eit_losses} display the algorithm loss' behavior with two thousand iterations/epochs, also the linear regression represents the performance's increase from the beginning till the last evaluation. The loss values corresponds to the "validation set" fold, obtained from each iteration of the cross-validation algorithm (explained before).

Further information about the optimization process can be found in the appendix \ref{appendix:fig}, Figure \ref{res_EIT_PTE_LGBM}. Hereby, it is feasible to find the performance of the optimizer through each epochs, and the "intelligent" behavior for each hyper-parameter's set. The listed parameters for the light gradient boosting machine are the same as those of the previous database (blood biomarkers). Despite the number of epochs, the algorithm always try to select the best candidates of hyper-parameters for the next following iteration. The orange stars in the plots indicate the top 10 performances. Moreover, the bayesian optimization using XGBoosting as objective function was similar to the LGBM one. The only differences were the hyper-parameters set and search space. The Figure \ref{res_blood_PTE_XGBM} in appendix \ref{appendix:fig} exhibit the bayesian performance through two thousand epochs. The hyper-parameters space has the same ten variables as those from the previous database. What we achieved after the bayesian optimization so far, regarding accuracy, precision and recall were 99\%, 97\% and 100\%, respectively. 

\begin{table}[htb]
\caption[Top metrics in the EIT's database (before and after optimization)]{Top metrics in the EIT's database (before and after optimization). Each phase has both techniques, XGBoosting and LGBM; it has five metrics (accuracy, precision, recall, F1 score and ROC AUC) and the execution time per each technique.}
\label{resul_EIT_PTE_tbl}
\begin{tabular}{@{}lcccccc@{}}
\toprule
                    & \multicolumn{6}{l}{Phase 1: Top baseline models }                                       \\ \cmidrule{2-7}
Model               & Accuracy & Precision & Recall & F1 score & ROC AUC & Time (ms) \\ \midrule
XGB regression & 0.94 & 1.00 & 0.83 & 0.91 & 0.92 & 1.56\\
LGBM & 0.91 & 0.90 & 0.82 & 0.86 & 0.89 & 1.56\\
\bottomrule
                    & \multicolumn{6}{l}{Phase 2: Parzen tree with bayesian optimization (2000 epochs)}                                       \\ \cmidrule{2-7}
Model               & Accuracy & Precision & Recall & F1 score & ROC AUC & Time (ms) \\ \midrule
XGB Regressor & 0.99 & 0.97 & 1.00 & 0.98 & 0.99 & 161.15 \\
LGBM & 0.98 & 1.00 & 0.93 & 0.96 & 0.97	& 84.3 \\ 
\bottomrule
\end{tabular}
\end{table}

To summarize, the Table \ref{resul_EIT_PTE_tbl} has the main metrics about each algorithm during the two phases, before and after the bayesian optimization. It is important to recall that firstly, XGBoosting is the best MLT for this database, secondly, some types of data enhancement don't worked as we expected, maybe due to the redundancy or combination of features e. g. polynomial features and scaling. Thirdly, the number of estimators, lambda and alpha regularization in LGBM, eta and gamma in XGBM and learning rate in both techniques, were the hyper-parameters which brought a better chance of performance increasing. Finally, we achieved a more than 7\% increasing in the performance when bayesian optimization was used, reducing the FN and FP rates.


\section{Thermography}
\label{ch:results_thermal}

\begin{figure}[H]
\centering
\includegraphics[width=6in]{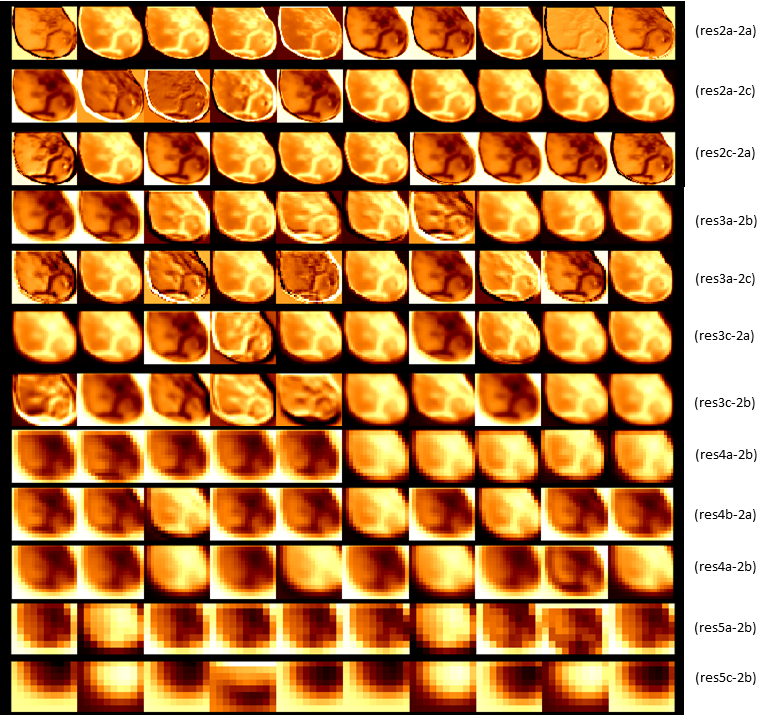} 
\caption[Layers' outputs for the CNN ResNet50 architecture going from layer res2a-2a to res5c-2b]{Layers' outputs for the CNN ResNet50 \cite{resnet} architecture going from layer res2a-2a to res5c-2b (The number of layers are many more, but for graphical purposes we display a few ones per block). In this image, it is represented the first ten outputs from the whole set of outputs of those layers. Each layer possess an specific number of filters, and each filter produce one image (i.e. this set of images belong to the first CNN benchmark experiment, ResNet50). The first layer (res2a-2a) normally tries to find global patterns, like breast's shape and size, nevertheless, deeper layers like \textit{res5a-2b} or \textit{res5c-2b} try to look for specific patterns like malformations and/or clustering of high temperature. This figure represent the left breast of a woman diagnosed with breast cancer.}
\label{res_thermography_CNN}
\end{figure}

This section shows the main results covering the thermal database. The main two approaches taken for image classification where deep neural networks, convolutional neural networks and a Keras applications module, that contains several Keras pre-trained deep learning modules, such as VGG16, ResNet50, Xception, MobileNet, and more. Here, we used the ResNet50\footnote{More information about the CNN architecture and the approaches could be found in \cite{resnet}} and NASNetMobile\footnote{More information about the NASNetMobile and NASNetLarge architectures are in \cite{nasnet}} architectures to compare the advantages and drawbacks of state-of-the-art techniques for image classification. It is important to recall, that the majority of those architectures are designed for multi-class classification of large datasets, nevertheless, the thermal dataset is just composed of two possibles classes, healthy or cancerous breast. The Python code regarding the thermal database, works as follows, firstly, we use the function \textit{"os.listdir"} to create a list of all the folders' path, for both, healthy and breast cancer cases. Then, with a exclusive function for data uploading, it is read all the 20 images per patient and we stored then in a list full of numpy arrays. The images are in grey scale, therefore there is just one channel per image (RGB images normally have three channels). The pre-processing algorithm, take an image as an input, then using \textit{OpenCV} it is found the image's contours. We create a bounding box from the vertices, afterwards a function processed and cropped the input image. The cropping is necessary, cause each image has a black background that will increase the processing time in the case of not using processing (as shown in Figure \ref{res_thermography_CNN}). Finally, each image had a different size, ranging from 200x180 to 300x400 pixels, thus, the final step is resize all images into an standard size, 250x300 pixels.

\begin{figure}
\centering
\includegraphics[width=5.5in]{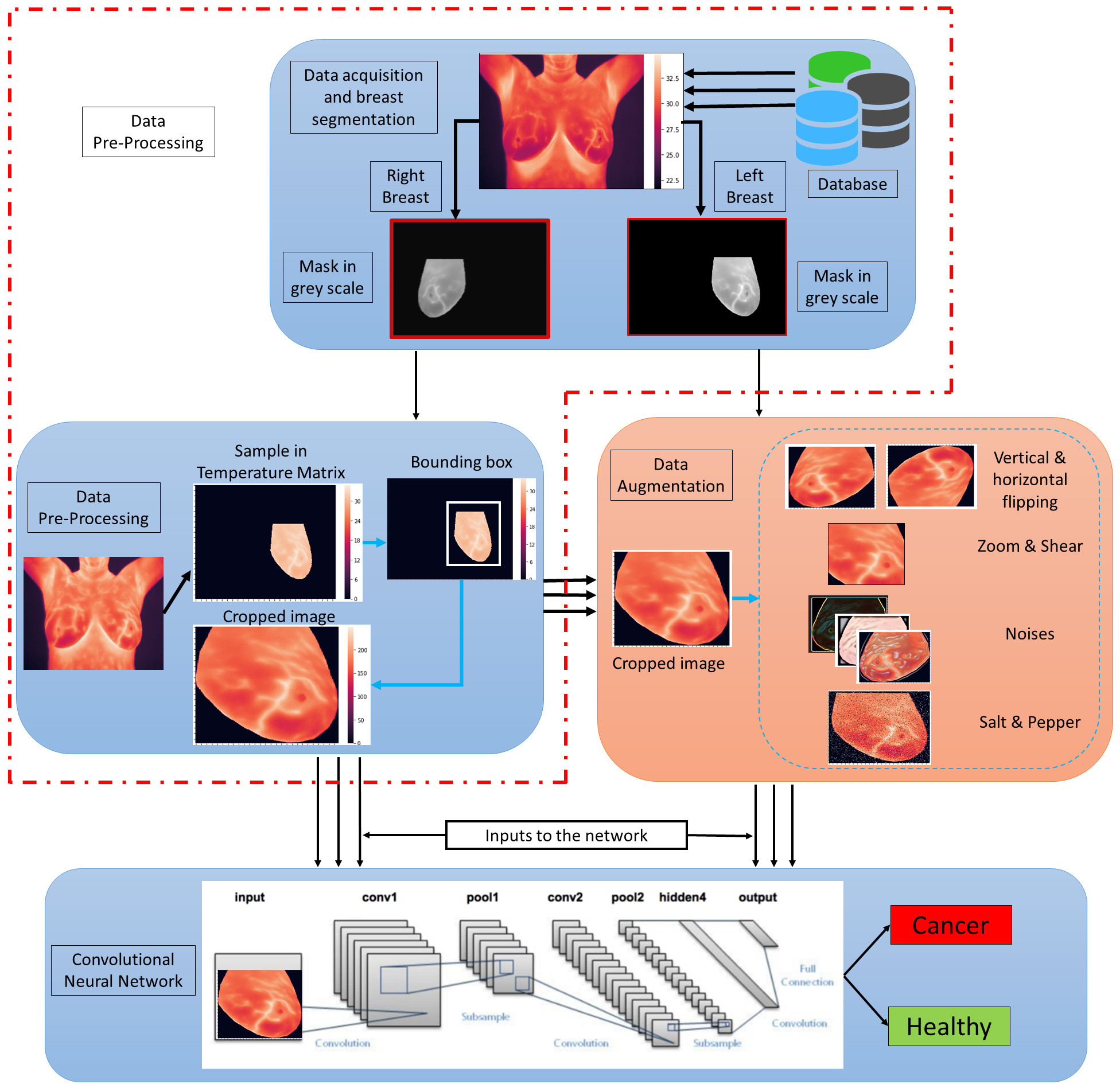} 
\caption{Global approach for data pre-processing, augmentation, and training based on CNN models for the thermography database.}
\label{res_thermography_CNN_model}
\end{figure}

The data augmentation phase involves vertical and horizontal flipping, horizontal rotation (90 degrees), addition of gaussian and salt \& pepper noise. This process was fundamental to reduce the model's overfitting and increase the robust against new testing samples. Therefore, Figure \ref{res_thermography_CNN_model} show the global approach for data pre-processing, augmentation and training phase. It is important to mention, that the thermal images are either from healthy or sick patients, nonetheless, after doing an EDA, it was found that there are not provided the tumor's location, size or depth. Hence, it is possible to suggest that the machine learning model needs to learn towards spatial information, the CNN architecture stand as top learner for predicting breast cancer from breast thermograms. We carried out several experiments with different machine learning techniques, nevertheless the CNN models yielded the best results, such as 97\%, 100\% and 95\% in accuracy, precision and recall, respectively. On the other hand, we created baseline models with less complex machine learning techniques such as, linear, logistic regression, decision trees, random forest and XGBoosting. The data pre-processing for those baseline models had one more step called "dataset unrolling" where we converted a 250x300 pixels image into a 75000 values array, because it is a mandatory input shape for train and test these types of Python models. The Table \ref{resul_Doe_thermal} gives further information about their performances, nevertheless, through this section, we will be focused in convolutional neural networks rather than other type of models. 

\begin{figure}
\centering
\includegraphics[width=5.4in]{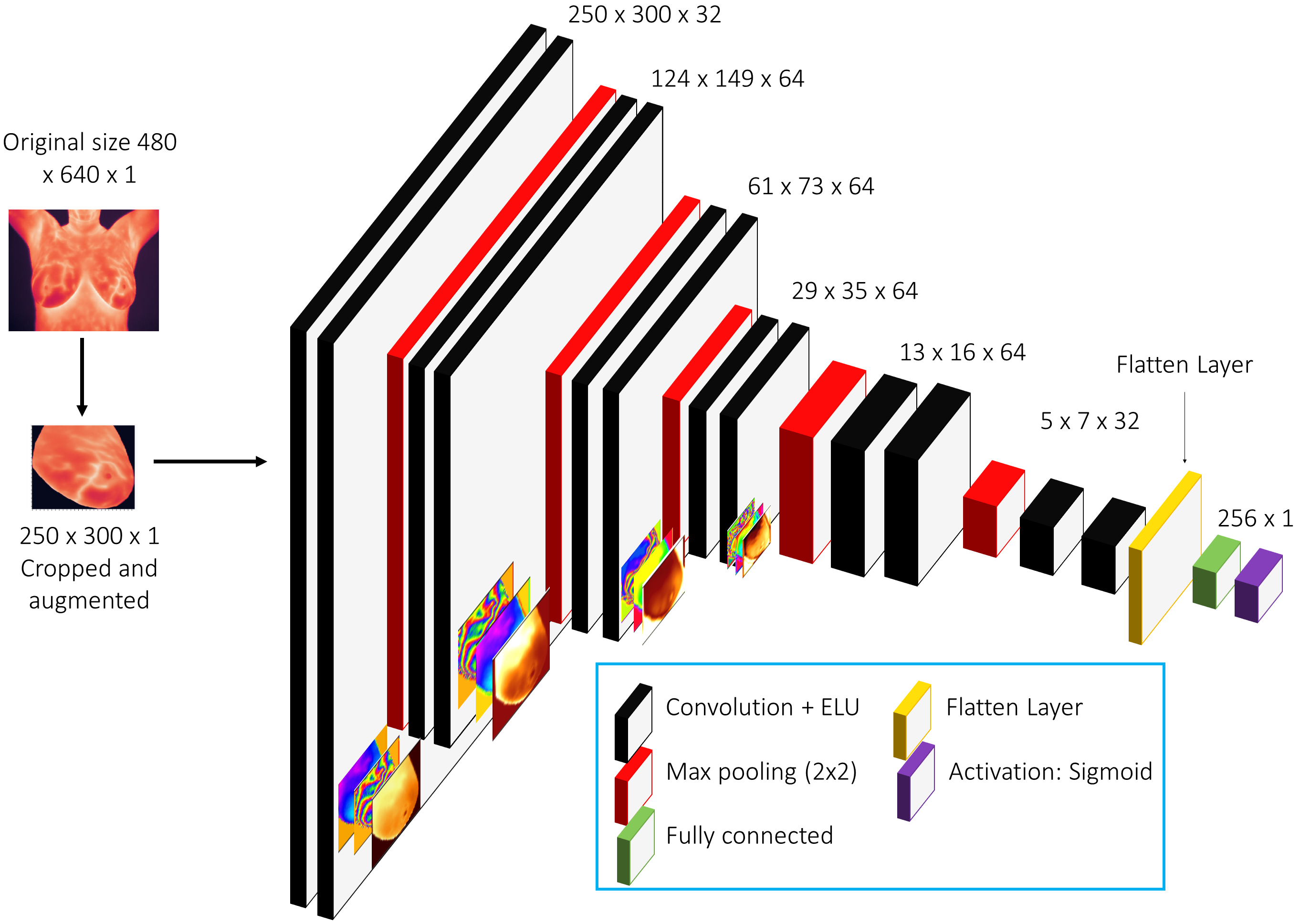} 
\caption{Convolutional neural network architecture for the experiment number two.}
\label{res_thermography_CNN_arch}
\end{figure}

\begin{table}[htb]
\caption[Top metrics for the thermography database baseline models]{Top metrics for the thermography database baseline models (six MLT). Likewise, each MLT has five metrics (accuracy, precision, recall, F1 score and ROC AUC), the threshold (0.2-0.8) which yielded the best performance, and the execution time.}
\label{resul_Doe_thermal}
\begin{tabular}{@{}lcccccc@{}}
\toprule
                    & \multicolumn{6}{l}{Original Dataset for baseline models}                                       \\ \cmidrule{2-7}
Model               & Accuracy & Precision & Recall & F1 score & ROC AUC  & Time (ms) \\ \midrule
Linear Regression   & 0.33     & 0.50      & 0.25   & 0.33     & 0.38 & 1875      \\
Logistic Regression & 0.33     & 0.50      & 0.17   & 0.33     & 0.41 & 8793      \\
Decision Trees      & 0.33     & 0.50      & 0.29   & 0.33     & 0.35 & 5946      \\
Random Forest       & 0.33     & 0.50      & 0.20   & 0.33     & 0.40 & 28000      \\
KNN                 & 0.33     & 0.50      & 0.33   & 0.39     & 0.34 & 6051      \\
XGB Reressor        & 0.35     & 0.50      & 0.37   & 0.43     & 0.32 & 57525      \\
\bottomrule
\end{tabular}
\end{table}

The CNN is a powerful machine learning technique composed by different type of layers, such as convolutional, max pooling, flatten, average pooling and full connected layers. Normally, CNNs are suited for image classification and segmentation, therefore this technique has been rising in popularity in the medical field.

Figure \ref{res_thermography_CNN_arch} shows a CNN architecture (experiment 2 of Table \ref{resul_thermal_CNN_DNN_tbl} and \ref{resul_thermal_CNN_DNN_tbl_2}) for image classification of breast thermograms. An advantage of CNN over a normal DNN is the processing time. Currently, the GPU and TPU\footnote{A tensor processing unit (TPU) is an AI accelerator application-specific integrated circuit (ASIC) developed by Google specifically for neural network machine learning} are key units for drastically reduce the training times (> 20 times) of those models. The requirements to use the Colab platform are (1) posses an Gmail account, (2) internet connection and (3) knowledge in Python-based Jupyter notebooks. During the training phase, we used the free Google's Colab \footnote{Google Colab is a free to use Jupyter notebook, that allows you to use a free Tesla K80 GPU it also gives a total of 12 GB of ram, and it can be use up to 12 hours in row} GPU unit. The GPU reduced more than twenty times the processing time. Further information and metrics regarding the CNN model presented in the Figure \ref{res_thermography_CNN_arch} could be found in the Table \ref{resul_thermal_CNN_DNN_tbl}. Here, we try to compare the performance of several architectures, and how the number of fully connected layer, their units, and dropout rate influenced the performance. Similar to Table \ref{resul_thermal_CNN_DNN_tbl}, the Table \ref{resul_thermal_CNN_DNN_tbl_2} show the performance during the testing phase. It is important to mention that all the models were trained with validation sets in order to reduce the bias and over-fitting. After the training process the, we wanted to provide a better insights on how works the convolutional layers into a CNN model, therefore the Figure \ref{res_thermography_CNN} from appendix \ref{appendix:fig:thermo} shows the output of 10 convolutional layers from the experiment 5. 

\begin{table}[htb]
\caption[Convolutional neural networks architectures for the thermography database]{Convolutional neural networks architectures for the thermography database. Each experiment has a different structure and characteristics like number of layers, dropout or batch normalization rate.}
\label{resul_thermal_CNN_DNN_tbl}
\begin{tabular}{@{}lcccccccc@{}}
\toprule
                    & \multicolumn{8}{l}{Convolutional Neural Network top models}\\ \cmidrule{2-9}
                    & \multicolumn{3}{l}{Layers} & \multicolumn{2}{l}{Net characteristics} & \multicolumn{3}{l}{Hyper-parameters}\\ \cmidrule{2-9}
                                        
Model               & Fully C. & Conv.	& Pooling & Filters & Kernel & Batch & Dropout & Activation \\ 
	                & (Units) & 2D	 	& (Size)  & (Size)  & (Size) & Normal.  & (Rate) & function \\ \midrule
Exp 1. 				& 1 (128) & 6 	& 2x2 & 2x32, 4x64 & 3x3 & No & 0.25 & ReLU\\
Exp 2. 				& 1 (256) & 12 	& 2x2 & 4x32, 8x64 & 3x3 & Yes & 0.3  & ELU\\
Exp 3. 				& 2 (512) & 14 	& 2x2 & 8x32, 6x64 & 3x3 & Yes & 0.1  & ELU\\
Exp 4. 				& 1 (1024)& 8   & 2x2 & 6x128, 2x256 & 3x3 & Yes & 0.01  & ELU\\
\bottomrule
\end{tabular}
\end{table}

Each layer from Figure \ref{res_thermography_CNN_arch} has a number of filter, and from each filter we can get one image, nevertheless, it is shown the first 10 filters from each layer. On the other hand, we made a benchmark of state-of-the-art CNN architectures like, NasNet, Inception, Resnet and so forth. Table \ref{resul_thermal_CNN_DNN_tbl_3} exhibits the metrics for those models, it is employed a "\textbf{fit-generator}" method, where are created mini data-augmented training batches (32 augmented training samples). Firstly, the training process had 50 steps per epoch (50 evaluation of 32 instances), for 50 epochs. Additionally, in each epoch it is 30 validation steps. The metrics for each CNN architecture has been done with blind test samples i.e. samples that has not been seen by the models. In the benchmark process, it is keep the architectures, but the top layer and weights are deleted. A flatten and fully connected layer of 1024 units are created after the CNNs architectures. The output dimension is 1 unit with a sigmoid activation function.   

\begin{table}[htb]
\centering
\caption[Top metrics in the experiments on the thermography's database]{Top metrics in the thermography's database. Each neural network architecture has five metrics (accuracy, precision, recall, F1 score and ROC AUC), execution time per each experiment and the optimizer type.}
\label{resul_thermal_CNN_DNN_tbl_2}
\begin{tabular}{@{}lccccccc@{}}
\toprule
                    & \multicolumn{7}{l}{Convolutional Neural Network top models}                                       \\ \cmidrule{2-8}
Model               & Accuracy & Precision & Recall & F1 score & ROC AUC & Time (ms) & Optimizer \\ \midrule
Exp 1. & 0.65 & 1.00 & 0.78 & 0.65 & 0.5  & 700 	& Adam\\
Exp 2. & 0.83 & 1.00 & 0.72 & 0.81 & 0.86 & 1150	& Adam\\ 
Exp 3. & 0.95 & 0.97 & 0.95 & 0.95 & 0.94 & 1300	& Adam\\
Exp 4. & 0.97 & 1.00 & 0.95 & 0.97 & 0.97 & 690		& Adam\\
\bottomrule
\end{tabular}
\end{table}

\begin{table}[htb]
\centering
\caption[Top metrics in the benchmark of CNN architectures on the thermography's database]{Top metrics in the thermography's database. Each CNN architecture has five metrics (accuracy, precision, recall, F1 score and ROC AUC), execution time per each epoch and optimizer type. This table display the benchmark results of the predefined models from the Keras application module.}
\label{resul_thermal_CNN_DNN_tbl_3}
\begin{tabular}{@{}lcccccc@{}}
\toprule
                    & \multicolumn{6}{l}{Convolutional Neural Network top models}                                       \\ \cmidrule{2-7}
Model               & Accuracy & Precision & Recall & F1 score & ROC AUC & Time (s)\\ 
\midrule
ResNet50 \cite{resnet}						& 0.90 & 0.88 & 0.98 & 0.93 & 0.85 & 20	\\
NASNetMobile \cite{nasnet}					& 0.67 & 0.67 & 1.00 & 0.79 & 0.5  & 25	\\
InceptionResNetV2 \cite{inception_resnet} 	& 0.83 & 0.85 & 0.90 & 0.87 & 0.80 & 45	\\ 
MobileNetV2 \cite{mobile_net}				& 0.67 & 0.67 & 1.00 & 0.80 & 0.5  & 18	\\
Xception \cite{Xception}					& 0.91 & 0.91 & 0.96 & 0.94 & 0.90 & 50	\\
\bottomrule
\end{tabular}
\end{table}

\begin{figure}[H]
\centering
\includegraphics[width=5.5in]{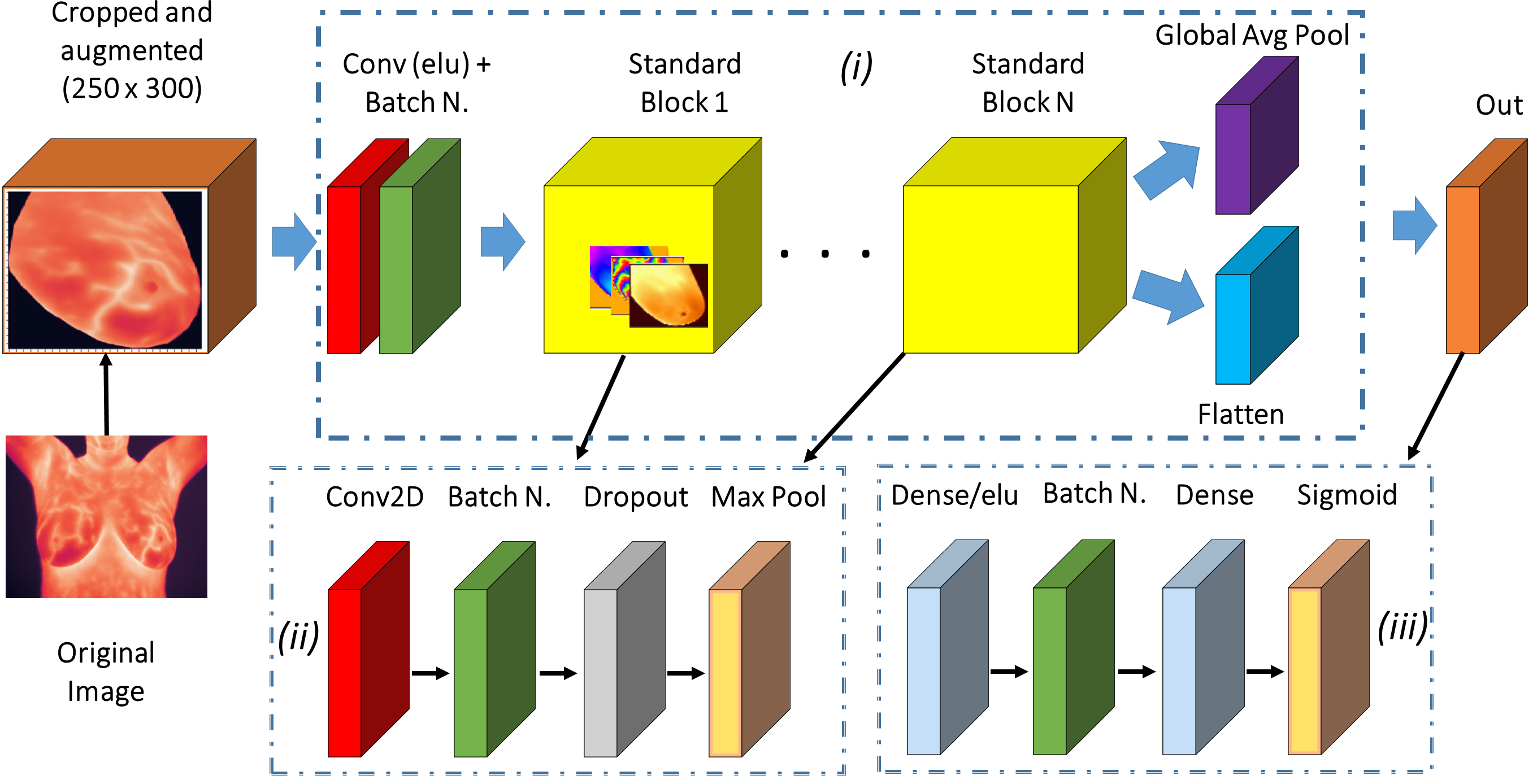} 
\caption{Convolutional neural network architecture for the unbiased experiments.}
\label{res_thermography_CNN_arch_new}
\end{figure}

The training and testing sets for the previous models have been extracted from the whole shuffled database, therefore, there could be the chance that one image from the same patient is in both sets, train and test. That has been done in order to compare the results with similar studies. Nevertheless, the next models and tables will show the results obtained in what we called "unbiased experiments", here we ensure that an image from the same patient is just going to be in either, the train or test set. During those experiments, we have used the CNN architecture from Figure \ref{res_thermography_CNN_arch_new}. This configuration allows us to create (i) "standard blocks" with (ii) 2D convolutional layers then batch normalization and dropouts; for down-sampling the output of each block, we implemented 2D max pooling. Finally, we use flatten and 2D global average pooling to connect the model with the (iii) output block, made of two dense layers of 1024 units each one, separated by a batch normalization function. 

Further information and results about experiments done with our proposed architecture from Figure \ref{res_thermography_CNN_arch_new} are in the appendix \ref{appendix:tables}, Table \ref{resul_thermal_CNN_DNN_tbl_3_unbiased} for benchmark models. It is important to recall, that the second experiment with the Inception module (version 3) yielded an accuracy of 90\% (benchamark top model). On other other hand, Table \ref{resul_thermal_CNN_new_blocks} present the results for our architecture using unbiased train and testing sets. Here, in the experiment number four, we achieved a model accuracy, precision and recall of 92\%, 94\% and 91\%, respectively. To summarize, we conclude that despite the number of layers, hyper-parameters such as, dropout rate, activation function and optimizer influenced in a bigger proportion the model's performance. Besides, we achieved a 97\% accuracy in diagnosis breast cancer from the thermal database, indeed, much higher than other conventional techniques mentioned in Tabble \ref{summarize_table}, such as KNN or SVM \cite{KNN_SVM_1, KNN_SVM_2}. On the other hand, the predefined models from the \textit{Keras applications module} were not as successful as the ones presented on Table \ref{resul_thermal_CNN_DNN_tbl} and \ref{resul_thermal_CNN_DNN_tbl_2}, we assume that our models are better for small datasets and not for image multi-class classification as such presented in Table \ref{resul_thermal_CNN_DNN_tbl_3}. Finally, it is important to mention that the thermography database is composed of 56 patients where each one has 20 thermal images, therefore, we split the sets in a way that all the images per patients went to the training or testing set (second set of experiments, Table \ref{res_thermography_CNN_arch_new} and \ref{resul_thermal_CNN_DNN_tbl_3_unbiased}), contrary as other methodologies where both sets contain images from the same patient \cite{thermo_CNN}.

\section{Deliverables}
\label{ch:results_other}

The following deliverables will be handed in with this thesis:

\begin{itemize}
\item This work produced three main algorithms. The first one and second one are similar cause the databases were similar. One algorithm for the blood biomarkers database and other for electrical impedance tomography database. The last algorithm is for the thermography database. Each algorithm provide comments about reading, pre-processing and augmentation in the database, then it is trained several machine learning techniques. 
\item The state of the art from this thesis helped to develop a review article submitted to a top journal (one copy of the article is handed annex).
\item The Python scripts regarding the three databases are attached to the annex of this report. Each database has a main and special functions scripts.
\item For future research is planned an article describing the whole machine learning process for breast cancer diagnosis with three approaches.
\end{itemize}

\chapter{Conclusion}
\label{ch:conclusion}

The early-detection of BC plays a key role in reducing the mortality rate; nevertheless, many authors explain the limitations when only humans take part in the breast cancer judgement, occasionally with non-permissible false positive and false negative cases, consequently, the integration of CAD systems into the current BC screening methods, surely will boost up the performance and popularity of these new techniques. Indeed, thermography, EIT and biomarkers found in blood test, are not contemporary techniques for breast cancer screening, nevertheless, until the last years, the price of implementation, complexity and accurateness compared with similar techniques, diminished the chance of proliferation as a BC diagnosis technique. In addition, higher the confident rate higher the technique cost (as showed in Table \ref{x_comparison_between_technique}), therefore, many researchers have been working in develop new easy-to-use, intelligent and inexpensive systems for detecting BC. 

Nowadays, these techniques normally are coupled with a CAD system made of a measurement phases, typically thermography, EIT, blood biomarkers, mammography, etc., afterwards, a CAD system will pre-process, find hidden patterns (such as the ones presented in Figure \ref{res_thermography_CNN}), post-process and diagnosing. Certainly, these objectives allowed the inclusion of MLT in the pipeline for BC detection; Yassin et al. \cite{review_thermography_3} reviews the MLTs in thermal imaging for BC, explaining that the biomedical information like image, signals or stacked data, are a complex set of information that describe a person health. In conclusion, it was possible to find that the best machine learning techniques for the above databases may vary depending the type of data. On one side, the stacked databases such as, EIT's and blood biomarkers are considered stacked data, and we confirm that gradient boosting machines were the top MLTs (further information can be seen in Table \ref{resul_Doe_blood} and \ref{resul_Doe_EIT}). On the other hand, the thermography database needed further pre-processing and augmentation in order to obtain an acceptable result. It is also proved that the exploratory data analysis gives us better insights regarding our data, with correlation matrices, pair plots and dimensionality reduction. During the training and testing phase, it is concluded that the hyper-parameters optimization let the MLT went further and achieve a better performance, for example, with XGBoosting regressor was possible to reduce to 0\% the false negative rate and almost 0\% the false positive rate (for EIT and blood biomarkers databases). Regarding the thermal images dataset, it was more important the quantity of filters in the convolutional layers rather than the number of layers and complexity. The benchmark models from the Table \ref{resul_thermal_CNN_DNN_tbl_3} did not performed as well as the ones we presented in Table \ref{resul_thermal_CNN_DNN_tbl_2}, hence, we assume that the predefined models maybe are too complex for thermal data where the temperature variation could be just a few degrees. Although the results showed in this report, many questions still pending, e.g. is it abundantly and well balance the available databases? Is it necessary to follow a protocol prior to an EIT, thermal test or blood analysis? Is it enough the false positive and false negative rate or the current models still lack performance? Those questions may bring new insights in the future work that researchers will take. We propose as future research, find the optimal thermal points required for detect a tumor with 1 cm size and 2.5 cm depth, likewise, the evaluation of 3D and 2D breast models in thermal and electrical specialized CAD software.

\begin{singlespace}
\bibliographystyle{vancouver}
\bibliography{main}

\begin{thebibliography}{100}

\bibitem{cancer_1}
{What is Cancer? National Cancer Institute}; 2015.
\newblock [Online; accessed 03-March-2019].
\newblock
  \url{https://www.cancer.gov/about-cancer/understanding/what-is-cancer}.

\bibitem{cancer_4}
Mambou S, Maresova P, Krejcar O, Selamat A, Kuca K.
\newblock Breast Cancer Detection Using Infrared Thermal Imaging and a Deep
  Learning Model.
\newblock Sensors. 2018;18(9):2799.

\bibitem{art_1}
Li X, Bond EJ, Van~Veen BD, Hagness SC.
\newblock An overview of ultra-wideband microwave imaging via space-time
  beamforming for early-stage breast-cancer detection.
\newblock IEEE Antennas and Propagation Magazine. 2005;47(1):19-34.

\bibitem{cancer_2}
{What is Cancer? Cancer Research UK}; 2019.
\newblock [Online; accessed 03-March-2019].
\newblock \url{https://www.cancerresearchuk.org/about-cancer/what-is-cancer}.

\bibitem{cancer_3}
{All Cancer Globocan 2018 - International Agency for Research on Cancer WHO};
  2019.
\newblock [Online; accessed 03-March-2019].
\newblock
  \url{http://gco.iarc.fr/today/data/factsheets/cancers/39-All-cancers-fact-sheet.pdf}.

\bibitem{bray2018global}
Bray F, Ferlay J, Soerjomataram I, Siegel RL, Torre LA, Jemal A.
\newblock Global cancer statistics 2018: GLOBOCAN estimates of incidence and
  mortality worldwide for 36 cancers in 185 countries.
\newblock CA: a cancer journal for clinicians. 2018;68(6):394-424.

\bibitem{bray2012}
Bray F, Jemal A, Grey N, Ferlay J, Forman D.
\newblock Global cancer transitions according to the Human Development Index
  (2008--2030): a population-based study.
\newblock The lancet oncology. 2012;13(8):790-801.

\bibitem{gersten2002}
Gersten O, Wilmoth JR.
\newblock The cancer transition in Japan since 1951.
\newblock Demographic Research. 2002;7:271-306.

\bibitem{omran49}
Omran A.
\newblock The epidemiologic transition: a theory of the epidemiology of
  population change-Milbank Mem.
\newblock Fund Quart-49-1971. 2005:509-38.

\bibitem{maule2012}
Maule M, Merletti F.
\newblock Cancer transition and priorities for cancer control.
\newblock The lancet oncology. 2012;13(8):745-6.

\bibitem{art_2}
Remennick L.
\newblock The challenge of early breast cancer detection among immigrant and
  minority women in multicultural societies.
\newblock The breast journal. 2006;12:S103-10.

\bibitem{kaiser}
{Women’s Health Insurance Coverage - Kaiser Family Foundation}; 2018.
\newblock [Online; accessed 11-March-2019].
\newblock
  \url{https://www.kff.org/womens-health-policy/fact-sheet/womens-health-insurance-coverage-fact-sheet/}.

\bibitem{review_thermography_2}
Kandlikar SG, Perez-Raya I, Raghupathi PA, Gonzalez-Hernandez JL, Dabydeen D,
  Medeiros L, et~al.
\newblock Infrared imaging technology for breast cancer detection--Current
  status, protocols and new directions.
\newblock International Journal of Heat and Mass Transfer. 2017;108:2303-20.

\bibitem{art_4}
Mandelblatt JS, Cronin KA, Bailey S, Berry DA, De~Koning HJ, Draisma G, et~al.
\newblock Effects of mammography screening under different screening schedules:
  model estimates of potential benefits and harms.
\newblock Annals of internal medicine. 2009;151(10):738-47.

\bibitem{art_5}
G{\o}tzsche PC, J{\o}rgensen KJ.
\newblock Screening for breast cancer with mammography.
\newblock Cochrane database of systematic reviews. 2013;1(6).

\bibitem{art_3}
L{\o}berg M, Lousdal ML, Bretthauer M, Kalager M.
\newblock Benefits and harms of mammography screening.
\newblock Breast Cancer Research. 2015;17(1):63.

\bibitem{art_6}
Von~Eschenbach AC.
\newblock NCI remains committed to current mammography guidelines.
\newblock The oncologist. 2002;7(3):170-1.

\bibitem{art_7}
Schousboe JT, Kerlikowske K, Loh A, Cummings SR.
\newblock Personalizing mammography by breast density and other risk factors
  for breast cancer: analysis of health benefits and cost-effectiveness.
\newblock Annals of internal medicine. 2011;155(1):10-20.

\bibitem{art_8}
Corredor S, Valbuena M, Zuluaga J, Barrios M.
\newblock Design and Construction a measurer of total body water, fat mass and
  fat free mass using LabVIEW.
\newblock In: 2014 III International Congress of Engineering Mechatronics and
  Automation (CIIMA). IEEE; 2014. p. 1-4.

\bibitem{art_9}
Zheng B, Lederman D, Sumkin JH, Zuley ML, Gruss MZ, Lovy LS, et~al.
\newblock A preliminary evaluation of multi-probe resonance-frequency
  electrical impedance based measurements of the breast.
\newblock Academic radiology. 2011;18(2):220-9.

\bibitem{marques2012}
Marques RdS.
\newblock Segmenta{\c{c}}{\~a}o autom{\'a}tica das mamas em imagens
  t{\'e}rmicas. Disserta{\c{c}}{\~a}o de mestrado.
\newblock In: Instituto de Computa{\c{c}}{\~a}o Universidade Federal
  Fluminense; 2012. .

\bibitem{silva2015}
Da~Silva TE.
\newblock Uma Metodologia de Auxilio ao Diagnostico de Doencas de Mama a Partir
  de Termografias Dinamicas (in Portuguese).
\newblock Niterói, RJ, Brasil; 2015. .

\bibitem{silva2014new}
Silva L, Saade D, Sequeiros G, Silva A, Paiva A, Bravo R, et~al.
\newblock A new database for breast research with infrared image.
\newblock Journal of Medical Imaging and Health Informatics. 2014;4(1):92-100.

\bibitem{eit_1}
Jossinet J.
\newblock Variability of impedivity in normal and pathological breast tissue.
\newblock Medical and biological engineering and computing. 1996;34(5):346-50.

\bibitem{eit_2}
Da~Silva JE, De~S{\'a} JM, Jossinet J.
\newblock Classification of breast tissue by electrical impedance spectroscopy.
\newblock Medical and Biological Engineering and Computing. 2000;38(1):26-30.

\bibitem{patricio_1}
Patr{\'\i}cio M, Pereira J, Cris{\'o}stomo J, Matafome P, Gomes M, Sei{\c{c}}a
  R, et~al.
\newblock Using Resistin, glucose, age and BMI to predict the presence of
  breast cancer.
\newblock BMC cancer. 2018;18(1):29.

\bibitem{krawczyk2013}
Krawczyk B, Schaefer G, Zhu SY.
\newblock Breast cancer identification based on thermal analysis and a
  clustering and selection classification ensemble.
\newblock In: International Conference on Brain and Health Informatics.
  Springer; 2013. p. 256-65.

\bibitem{art_10}
Jones BF.
\newblock A reappraisal of the use of infrared thermal image analysis in
  medicine.
\newblock IEEE transactions on medical imaging. 1998;17(6):1019-27.

\bibitem{art_1_1}
Gautherie M.
\newblock Thermobiological assessment of benign and malignant breast diseases.
\newblock American journal of obstetrics and gynecology. 1983;147(8):861-9.

\bibitem{mammography_2}
McCormack VA, dos Santos~Silva I.
\newblock Breast density and parenchymal patterns as markers of breast cancer
  risk: a meta-analysis.
\newblock Cancer Epidemiology and Prevention Biomarkers. 2006;15(6):1159-69.

\bibitem{mammography_1}
Ursin G, Hovanessian-Larsen L, Parisky YR, Pike MC, Wu AH.
\newblock Greatly increased occurrence of breast cancers in areas of
  mammographically dense tissue.
\newblock Breast Cancer Research. 2005;7(5):R605.

\bibitem{mammography_3}
Li T, Sun L, Miller N, Nicklee T, Woo J, Hulse-Smith L, et~al.
\newblock The association of measured breast tissue characteristics with
  mammographic density and other risk factors for breast cancer.
\newblock Cancer Epidemiology and Prevention Biomarkers. 2005;14(2):343-9.

\bibitem{kontos2011}
Kontos M, Wilson R, Fentiman I.
\newblock Digital infrared thermal imaging (DITI) of breast lesions:
  sensitivity and specificity of detection of primary breast cancers.
\newblock Clinical radiology. 2011;66(6):536-9.

\bibitem{low_recall_1}
Moskowitz M, Milbrath J, Gartside P, Zermeno A, Mandel D.
\newblock Lack of efficacy of thermography as a screening tool for minimal and
  stage I breast cancer.
\newblock New England Journal of Medicine. 1976;295(5):249-52.

\bibitem{low_recall_2}
Threatt B, Norbeck JM, Ullman NS, Kummer R, Roselle PF.
\newblock Thermography and Breast Cancer. An Analysis of a Blind Reading.
\newblock Annals of the New York Academy of Sciences. 1980;335(1):501-19.

\bibitem{thermography_1}
Lawson R.
\newblock Implications of surface temperatures in the diagnosis of breast
  cancer.
\newblock Canadian Medical Association Journal. 1956;75(4):309.

\bibitem{thermography_2}
Maaopust L, Gardner W.
\newblock The infrared phlebogram in the diagnosis of breast complaints. An
  analysis of 1,200 cases.
\newblock Surg Gynecol Obstet. 1953;97:619-29.

\bibitem{thermography_3}
Lawson R.
\newblock A new infrared imaging device.
\newblock Canadian Medical Association Journal. 1958;79(5):402.

\bibitem{patent_1}
Alt LL, Lawson RN. Diagnostic thermography method and means. Google Patents;
  1967.
\newblock US Patent 3,335,716.

\bibitem{thermography_4}
Vogler WR, Powell RW.
\newblock A clinical evaluation of thermography and heptyl aldehyde in breast
  cancer detection.
\newblock Cancer research. 1959;19(2):207-9.

\bibitem{thermography_5}
Williams KL, Williams FL, Handley R.
\newblock Infra-red radiation thermometry in clinical practice.
\newblock The Lancet. 1960;276(7157):958-9.

\bibitem{patent_2}
Howell WL. Apparatus for use in differential clinical thermometry. Google
  Patents; 1967.
\newblock US Patent 3,339,542.

\bibitem{thermography_7}
Swearingen AG.
\newblock Thermography: Report of the radiographic and thermographic
  examinations of the breasts of 100 patients.
\newblock Radiology. 1965;85(5):818-24.

\bibitem{patent_3}
Barnes RB, Engborg NE. Thermographic scanner and recorder. Google Patents;
  1970.
\newblock US Patent 3,531,642.

\bibitem{patent_4}
Bowling BR. Process of diagnosis by infrared thermography. Google Patents;
  1966.
\newblock US Patent 3,245,402.

\bibitem{thermography_8}
ISARD HJ, BECKER W, SHILO R, OSTRUM BJ.
\newblock Breast thermography after four years and 10,000 studies.
\newblock American Journal of Roentgenology. 1972;115(4):811-21.

\bibitem{thermography_9}
Anbar M.
\newblock Clinical thermal imaging today.
\newblock IEEE Engineering in Medicine and Biology Magazine. 1998;17(4):25-33.

\bibitem{ng2009review}
Ng EK.
\newblock A review of thermography as promising non-invasive detection modality
  for breast tumor.
\newblock International Journal of Thermal Sciences. 2009;48(5):849-59.

\bibitem{tem_new_1}
Ng Y, Ung L, Ng F, LSJ~Sim E.
\newblock Statistical analysis of healthy and malignant breast thermography.
\newblock Journal of medical engineering \& technology. 2001;25(6):253-63.

\bibitem{tem_new_2}
Chen Y, Ng E, Ung L, Ng F.
\newblock Patterns and cyclic variations of thermography in female breasts.
\newblock In: Proc ICMMB-11: Int Conf Mech Med Biol; 2000. p. 2-5.

\bibitem{new_thermography_1}
Arena F, Barone C, DiCicco T.
\newblock Use of digital infrared imaging in enhanced breast cancer detection
  and monitoring of the clinical response to treatment.
\newblock In: Proceedings of the 25th Annual International Conference of the
  IEEE Engineering in Medicine and Biology Society (IEEE Cat. No. 03CH37439).
  vol.~2. IEEE; 2003. p. 1129-32.

\bibitem{new_thermography_2}
Partridge P, Wrobel L.
\newblock An inverse geometry problem for the localisation of skin tumours by
  thermal analysis.
\newblock Engineering Analysis with Boundary Elements. 2007;31(10):803-11.

\bibitem{new_thermography_2_1}
Das K, Mishra SC.
\newblock Estimation of tumor characteristics in a breast tissue with known
  skin surface temperature.
\newblock Journal of Thermal Biology. 2013;38(6):311-7.

\bibitem{new_thermography_3}
Kennedy DA, Lee T, Seely D.
\newblock A comparative review of thermography as a breast cancer screening
  technique.
\newblock Integrative cancer therapies. 2009;8(1):9-16.

\bibitem{new_thermography_4}
Acharya UR, Ng EYK, Tan JH, Sree SV.
\newblock Thermography based breast cancer detection using texture features and
  support vector machine.
\newblock Journal of medical systems. 2012;36(3):1503-10.

\bibitem{tem_new_3}
Ng EK, Fok S, Peh Y, Ng F, Sim L.
\newblock Computerized detection of breast cancer with artificial intelligence
  and thermograms.
\newblock Journal of medical engineering \& technology. 2002;26(4):152-7.

\bibitem{tem_new_4}
Ng E, Kee E.
\newblock Advanced integrated technique in breast cancer thermography.
\newblock Journal of Medical Engineering \& Technology. 2008;32(2):103-14.

\bibitem{new_thermography_5}
Schaefer G, Z{\'a}vi{\v{s}}ek M, Nakashima T.
\newblock Thermography based breast cancer analysis using statistical features
  and fuzzy classification.
\newblock Pattern Recognition. 2009;42(6):1133-7.

\bibitem{new_thermography_6}
Ara{\'u}jo MC, Lima RC, De~Souza RM.
\newblock Interval symbolic feature extraction for thermography breast cancer
  detection.
\newblock Expert Systems with Applications. 2014;41(15):6728-37.

\bibitem{new_thermography_8}
Sathish D, Kamath S, Prasad K, Kadavigere R.
\newblock Role of normalization of breast thermogram images and automatic
  classification of breast cancer.
\newblock The Visual Computer. 2017:1-14.

\bibitem{new_thermography_10}
Mambou S, Maresova P, Krejcar O, Selamat A, Kuca K.
\newblock Breast Cancer Detection Using Infrared Thermal Imaging and a Deep
  Learning Model.
\newblock Sensors. 2018;18(9):2799.

\bibitem{review_thermography_3}
Yassin NI, Omran S, El~Houby EM, Allam H.
\newblock Machine learning techniques for breast cancer computer aided
  diagnosis using different image modalities: A systematic review.
\newblock Computer methods and programs in biomedicine. 2018;156:25-45.

\bibitem{review_thermography_1}
Borchartt TB, Conci A, Lima RC, Resmini R, Sanchez A.
\newblock Breast thermography from an image processing viewpoint: A survey.
\newblock Signal Processing. 2013;93(10):2785-803.

\bibitem{new_thermography_7}
Mahmoudzadeh E, Montazeri M, Zekri M, Sadri S.
\newblock Extended hidden Markov model for optimized segmentation of breast
  thermography images.
\newblock Infrared Physics \& Technology. 2015;72:19-28.

\bibitem{new_thermography_9}
Silva LF, Santos AAS, Bravo RS, Silva AC, Muchaluat-Saade DC, Conci A.
\newblock Hybrid analysis for indicating patients with breast cancer using
  temperature time series.
\newblock Computer methods and programs in biomedicine. 2016;130:142-53.

\bibitem{new_thermography_9_1}
Gonzalez-Hernandez JL, Recinella AN, Kandlikar SG, Dabydeen D, Medeiros L,
  Phatak P.
\newblock Technology, application and potential of dynamic breast thermography
  for the detection of breast cancer.
\newblock International Journal of Heat and Mass Transfer. 2019;131:558-73.

\bibitem{new_thermography_11}
Lugoda P, Hughes-Riley T, Morris R, Dias T.
\newblock A Wearable Textile Thermograph.
\newblock Sensors. 2018;18(7):2369.

\bibitem{new_thermography_13}
Gogoi UR, Bhowmik MK, Bhattacharjee D, Ghosh AK.
\newblock Singular value based characterization and analysis of thermal patches
  for early breast abnormality detection.
\newblock Australasian physical \& engineering sciences in medicine.
  2018;41(4):861-79.

\bibitem{tem_new_5}
Joro R, L{\"a}{\"a}peri AL, Soimakallio S, J{\"a}rvenp{\"a}{\"a} R,
  Kuukasj{\"a}rvi T, Toivonen T, et~al.
\newblock Dynamic infrared imaging in identification of breast cancer tissue
  with combined image processing and frequency analysis.
\newblock Journal of medical engineering \& technology. 2008;32(4):325-35.

\bibitem{new_thermography_12}
Zhou Y, Herman C.
\newblock Optimization of skin cooling by computational modeling for early
  thermographic detection of breast cancer.
\newblock International Journal of Heat and Mass Transfer. 2018;126:864-76.

\bibitem{new_thermography_12_1}
Figueiredo AAA, do~Nascimento JG, Malheiros FC, da~Silva~Ignacio LH, Fernandes
  HC, Guimaraes G.
\newblock Breast tumor localization using skin surface temperatures from a 2D
  anatomic model without knowledge of the thermophysical properties.
\newblock Computer methods and programs in biomedicine. 2019;172:65-77.

\bibitem{new_thermography_14}
Gogoi UR, Majumdar G, Bhowmik MK, Ghosh AK.
\newblock Evaluating the Efficiency of Infrared Breast Thermography for Early
  Breast Cancer Risk Prediction in Asymptomatic Population.
\newblock Infrared Physics \& Technology. 2019.

\bibitem{pennes1948analysis}
Pennes HH.
\newblock Analysis of tissue and arterial blood temperatures in the resting
  human forearm.
\newblock Journal of applied physiology. 1948;1(2):93-122.

\bibitem{tem_new_6}
Lin Q, Yang H, Xie S, Wang Y, Ye Z, Chen S.
\newblock Detecting early breast tumour by finite element thermal analysis.
\newblock Journal of medical engineering \& technology. 2009;33(4):274-80.

\bibitem{eit_3}
Kubicek W, Patterson R, Witsoe D.
\newblock Impedance cardiography as a noninvasive method of monitoring cardiac
  function and other parameters of the cardiovascular system.
\newblock Annals of the New York Academy of Sciences. 1970;170(2):724-32.

\bibitem{eit_new_new_1}
Brown BH.
\newblock Electrical impedance tomography (EIT): a review.
\newblock Journal of medical engineering \& technology. 2003;27(3):97-108.

\bibitem{eit_6}
Stasiak M, Sikora J, Filipowicz SF, Nita K.
\newblock Principal component analysis and artificial neural network approach
  to electrical impedance tomography problems approximated by multi-region
  boundary element method.
\newblock Engineering Analysis with Boundary Elements. 2007;31(8):713-20.

\bibitem{eit_5}
Zou Y, Guo Z.
\newblock A review of electrical impedance techniques for breast cancer
  detection.
\newblock Medical engineering \& physics. 2003;25(2):79-90.

\bibitem{eit_first}
Fricke H, Morse S.
\newblock The electric capacity of tumors of the breast.
\newblock The Journal of Cancer Research. 1926;10(3):340-76.

\bibitem{eit_4_1}
Cheney M, Isaacson D, Newell JC, Simske S, Goble J.
\newblock NOSER: An algorithm for solving the inverse conductivity problem.
\newblock International Journal of Imaging systems and technology.
  1990;2(2):66-75.

\bibitem{eit_4}
Cheney M, Isaacson D, Newell JC.
\newblock Electrical impedance tomography.
\newblock SIAM review. 1999;41(1):85-101.

\bibitem{eit_7}
Zheng B, Lederman D, Sumkin JH, Zuley ML, Gruss MZ, Lovy LS, et~al.
\newblock A preliminary evaluation of multi-probe resonance-frequency
  electrical impedance based measurements of the breast.
\newblock Academic radiology. 2011;18(2):220-9.

\bibitem{eit_8}
Shetiye PC, Ghatol AA, Ghate VN.
\newblock Neural network based breast cancer classifier using electrical
  impedance.
\newblock In: 2012 Nirma University International Conference on Engineering
  (NUiCONE). IEEE; 2012. p. 1-4.

\bibitem{eit_9}
Calle-Alonso F, P{\'e}rez C, Arias-Nicol{\'a}s J, Mart{\'\i}n J.
\newblock Computer-aided diagnosis system: A Bayesian hybrid classification
  method.
\newblock Computer methods and programs in biomedicine. 2013;112(1):104-13.

\bibitem{eit_10}
Stojadinovic A, Nissan A, Gallimidi Z, Lenington S, Logan W, Zuley M, et~al.
\newblock Electrical impedance scanning for the early detection of breast
  cancer in young women: preliminary results of a multicenter prospective
  clinical trial.
\newblock Journal of Clinical Oncology. 2005;23(12):2703-15.

\bibitem{eit_11}
Haeri Z, Shokoufi M, Jenab M, Janzen R, Golnaraghi F.
\newblock Electrical impedance spectroscopy for breast cancer diagnosis:
  Clinical study.
\newblock Journal Integrative Cancer Science and Therapeutics. 2016;3(6):1-6.

\bibitem{eit_12}
Zarafshani A, Bach T, Chatwin CR, Tang S, Xiang L, Zheng B.
\newblock Conditioning electrical impedance mammography system.
\newblock Measurement. 2018;116:38-48.

\bibitem{device_10}
Carlak HF, Gencer NG, Besikci C.
\newblock Theoretical assessment of electro-thermal imaging: A new technique
  for medical diagnosis.
\newblock Infrared Physics \& Technology. 2016;76:227-34.

\bibitem{device_11}
Menegaz GL, Guimar{\~a}es G.
\newblock Development of a New Technique for Breast Tumor Detection Based on
  Thermal Impedance and a Damage Metric.
\newblock Infrared Physics \& Technology. 2019.

\bibitem{blood_1}
Moore LE, Pfeiffer RM, Poscablo C, Real FX, Kogevinas M, Silverman D, et~al.
\newblock Genomic DNA hypomethylation as a biomarker for bladder cancer
  susceptibility in the Spanish Bladder Cancer Study: a case--control study.
\newblock The lancet oncology. 2008;9(4):359-66.

\bibitem{blood_2}
Thakur BK, Zhang H, Becker A, Matei I, Huang Y, Costa-Silva B, et~al.
\newblock Double-stranded DNA in exosomes: a novel biomarker in cancer
  detection.
\newblock Cell research. 2014;24(6):766.

\bibitem{blood_3}
Lofton-Day C, Model F, DeVos T, Tetzner R, Distler J, Schuster M, et~al.
\newblock DNA methylation biomarkers for blood-based colorectal cancer
  screening.
\newblock Clinical chemistry. 2008;54(2):414-23.

\bibitem{blood_4}
Kobayashi Y, Absher DM, Gulzar ZG, Young SR, McKenney JK, Peehl DM, et~al.
\newblock DNA methylation profiling reveals novel biomarkers and important
  roles for DNA methyltransferases in prostate cancer.
\newblock Genome research. 2011;21(7):1017-27.

\bibitem{blood_5}
Brennan K, Garcia-Closas M, Orr N, Fletcher O, Jones M, Ashworth A, et~al.
\newblock Intragenic ATM methylation in peripheral blood DNA as a biomarker of
  breast cancer risk.
\newblock Cancer research. 2012;72(9):2304-13.

\bibitem{blood_6}
Hammond MEH, Hayes DF, Dowsett M, Allred DC, Hagerty KL, Badve S, et~al.
\newblock American Society of Clinical Oncology/College of American
  Pathologists guideline recommendations for immunohistochemical testing of
  estrogen and progesterone receptors in breast cancer (unabridged version).
\newblock Archives of pathology \& laboratory medicine. 2010;134(7):e48-72.

\bibitem{blood_7}
Dowsett M, Houghton J, Iden C, Salter J, Farndon J, A'hern R, et~al.
\newblock Benefit from adjuvant tamoxifen therapy in primary breast cancer
  patients according oestrogen receptor, progesterone receptor, EGF receptor
  and HER2 status.
\newblock Annals of Oncology. 2006;17(5):818-26.

\bibitem{blood_8}
Fu J, Zhong C, Wu L, Li D, Xu T, Jiang T, et~al.
\newblock Young Patients with Hormone Receptor-Positive Breast Cancer Have a
  Higher Long-Term Risk of Breast Cancer Specific Death.
\newblock Journal of Breast Cancer. 2019;22(1):96-108.

\bibitem{blood_9}
Zemouri R, Omri N, Morello B, Devalland C, Arnould L, Zerhouni N, et~al.
\newblock Constructive Deep Neural Network for Breast Cancer Diagnosis.
\newblock IFAC-PapersOnLine. 2018;51(27):98-103.

\bibitem{blood_10}
Howlader N, Noone A, Krapcho M, Miller D, Bishop K, Kosary C, et~al.
\newblock SEER Cancer Statistics Review, 1975-2014, National Cancer Institute.
  Bethesda, MD.
\newblock SEER Cancer Stat Facts: Kidney and Renal Pelvis Cancer. 2017;2017.

\bibitem{blood_11}
Kuchenbaecker KB, Hopper JL, Barnes DR, Phillips KA, Mooij TM, Roos-Blom MJ,
  et~al.
\newblock Risks of breast, ovarian, and contralateral breast cancer for BRCA1
  and BRCA2 mutation carriers.
\newblock Jama. 2017;317(23):2402-16.

\bibitem{blood_12}
Semmler L, Reiter-Brennan C, Klein A.
\newblock BRCA1 and Breast Cancer: a Review of the Underlying Mechanisms
  Resulting in the Tissue-Specific Tumorigenesis in Mutation Carriers.
\newblock Journal of Breast Cancer. 2019;22(1):1-14.

\bibitem{blood_13}
Weigel MT, Dowsett M.
\newblock Current and emerging biomarkers in breast cancer: prognosis and
  prediction.
\newblock Endocrine-related cancer. 2010;17(4):R245-62.

\bibitem{blood_14}
Brennan K, Garcia-Closas M, Orr N, Fletcher O, Jones M, Ashworth A, et~al.
\newblock Intragenic ATM methylation in peripheral blood DNA as a biomarker of
  breast cancer risk.
\newblock Cancer research. 2012;72(9):2304-13.

\bibitem{blood_15}
Wrensch M, Petrakis NL, King EB, Lee MM, Miike R.
\newblock Breast cancer risk associated with abnormal cytology in nipple
  aspirates of breast fluid and prior history of breast biopsy.
\newblock American journal of epidemiology. 1993;137(8):829-33.

\bibitem{blood_16}
Liu Y, Wang JL, Chang H, Barsky SH, Nguyen M.
\newblock Breast-cancer diagnosis with nipple fluid bFGF.
\newblock The Lancet. 2000;356(9229):567.

\bibitem{blood_17}
Paweletz CP, Trock B, Pennanen M, Tsangaris T, Magnant C, Liotta LA, et~al.
\newblock Proteomic patterns of nipple aspirate fluids obtained by SELDI-TOF:
  potential for new biomarkers to aid in the diagnosis of breast cancer.
\newblock Disease markers. 2001;17(4):301-7.

\bibitem{blood_18}
Li J, Zhang Z, Rosenzweig J, Wang YY, Chan DW.
\newblock Proteomics and bioinformatics approaches for identification of serum
  biomarkers to detect breast cancer.
\newblock Clinical chemistry. 2002;48(8):1296-304.

\bibitem{blood_19}
Mertins P, Mani D, Ruggles KV, Gillette MA, Clauser KR, Wang P, et~al.
\newblock Proteogenomics connects somatic mutations to signalling in breast
  cancer.
\newblock Nature. 2016;534(7605):55.

\bibitem{blood_20}
Dalamaga M, Sotiropoulos G, Karmaniolas K, Pelekanos N, Papadavid E, Lekka A.
\newblock Serum resistin: a biomarker of breast cancer in postmenopausal women?
  Association with clinicopathological characteristics, tumor markers,
  inflammatory and metabolic parameters.
\newblock Clinical biochemistry. 2013;46(7-8):584-90.

\bibitem{blood_21}
Santill{\'a}n-Ben{\'\i}tez JG, Mendieta-Zer{\'o}n H, G{\'o}mez-Oliv{\'a}n LM,
  Torres-Ju{\'a}rez JJ, Gonz{\'a}lez-Ba{\~n}ales JM, Hern{\'a}ndez-Pe{\~n}a LV,
  et~al.
\newblock The Tetrad BMI, Leptin, Leptin/Adiponectin (L/A) Ratio and CA 15-3
  are Reliable Biomarkers of Breast Cancer.
\newblock Journal of clinical laboratory analysis. 2013;27(1):12-20.

\bibitem{blood_22}
Hwa HL, Kuo WH, Chang LY, Wang MY, Tung TH, Chang KJ, et~al.
\newblock Prediction of breast cancer and lymph node metastatic status with
  tumour markers using logistic regression models.
\newblock Journal of evaluation in clinical practice. 2008;14(2):275-80.

\bibitem{blood_23}
Vandenberghe ME, Scott ML, Scorer PW, S{\"o}derberg M, Balcerzak D, Barker C.
\newblock Relevance of deep learning to facilitate the diagnosis of HER2 status
  in breast cancer.
\newblock Scientific reports. 2017;7:45938.

\bibitem{blood_24}
Robertson S, Azizpour H, Smith K, Hartman J.
\newblock Digital image analysis in breast pathology—from image processing
  techniques to artificial intelligence.
\newblock Translational Research. 2018;194:19-35.

\bibitem{blood_25}
Saha M, Chakraborty C.
\newblock Her2net: A deep framework for semantic segmentation and
  classification of cell membranes and nuclei in breast cancer evaluation.
\newblock IEEE Transactions on Image Processing. 2018;27(5):2189-200.

\bibitem{blood_26}
Mukundan R.
\newblock Analysis of Image Feature Characteristics for Automated Scoring of
  HER2 in Histology Slides.
\newblock Journal of Imaging. 2019;5(3):35.

\bibitem{geron2017hands}
G{\'e}ron A.
\newblock Hands-on machine learning with Scikit-Learn and TensorFlow: concepts,
  tools, and techniques to build intelligent systems.
\newblock " O'Reilly Media, Inc."; 2017.

\bibitem{pedregosa2011scikit}
Pedregosa F, Varoquaux G, Gramfort A, Michel V, Thirion B, Grisel O, et~al.
\newblock Scikit-learn: Machine learning in Python.
\newblock Journal of machine learning research. 2011;12(Oct):2825-30.

\bibitem{abadi2016tensorflow}
Abadi M, Barham P, Chen J, Chen Z, Davis A, Dean J, et~al.
\newblock Tensorflow: A system for large-scale machine learning.
\newblock In: 12th Symposium on Operating Systems Design and Implementation;
  2016. p. 265-83.

\bibitem{resnet}
Yu X, Yu Z, Ramalingam S.
\newblock Learning Strict Identity Mappings in Deep Residual Networks.
\newblock 2018 IEEE/CVF Conference on Computer Vision and Pattern Recognition.
  2018.

\bibitem{nasnet}
Zoph B, Vasudevan V, Shlens J, Le QV.
\newblock Learning Transferable Architectures for Scalable Image Recognition.
\newblock 2018 IEEE/CVF Conference on Computer Vision and Pattern Recognition.
  2018.

\bibitem{inception_resnet}
Szegedy C, Ioffe S, Vanhoucke V, Alemi AA.
\newblock Inception-v4, inception-resnet and the impact of residual connections
  on learning.
\newblock In: Thirty-First AAAI Conference on Artificial Intelligence; 2017. .

\bibitem{mobile_net}
Sandler M, Howard A, Zhu M, Zhmoginov A, Chen LC.
\newblock Mobilenetv2: Inverted residuals and linear bottlenecks.
\newblock In: Proceedings of the IEEE Conference on Computer Vision and Pattern
  Recognition; 2018. p. 4510-20.

\bibitem{Xception}
Chollet F.
\newblock Xception: Deep learning with depthwise separable convolutions.
\newblock In: Proceedings of the IEEE conference on computer vision and pattern
  recognition; 2017. p. 1251-8.

\bibitem{KNN_SVM_1}
Ali MA, Sayed GI, Gaber T, Hassanien AE, Snasel V, Silva LF.
\newblock Detection of breast abnormalities of thermograms based on a new
  segmentation method.
\newblock In: 2015 Federated Conference on Computer Science and Information
  Systems (FedCSIS). IEEE; 2015. p. 255-61.

\bibitem{KNN_SVM_2}
Sathish D, Kamath S, Prasad K, Kadavigere R, Martis RJ.
\newblock Asymmetry analysis of breast thermograms using automated segmentation
  and texture features.
\newblock Signal, Image and Video Processing. 2017;11(4):745-52.

\bibitem{thermo_CNN}
Fern{\'a}ndez-Ovies FJ, Alf{\'e}rez-Baquero ES, de~Andr{\'e}s-Galiana EJ,
  Cernea A, Fern{\'a}ndez-Mu{\~n}iz Z, Fern{\'a}ndez-Mart{\'\i}nez JL.
\newblock Detection of Breast Cancer Using Infrared Thermography and Deep
  Neural Networks.
\newblock In: International Work-Conference on Bioinformatics and Biomedical
  Engineering. Springer; 2019. p. 514-23.

\bibitem{new_thermography_15}
Sarigoz T, Ertan T, Topuz O, Sevim Y, Cihan Y.
\newblock Role of digital infrared thermal imaging in the diagnosis of breast
  mass: A pilot study: Diagnosis of breast mass by thermography.
\newblock Infrared Physics \& Technology. 2018;91:214-9.

\bibitem{device_5}
Ye G, Lim KH, George~Jr RT, Ybarra GA, Joines WT, Liu QH.
\newblock 3D EIT for breast cancer imaging: System, measurements, and
  reconstruction.
\newblock Microwave and Optical Technology Letters. 2008;50(12):3261-71.

\bibitem{device_6}
Pak D, Rozhkova N, Kireeva M, Ermoshchenkova M, Nazarov A, Fomin D, et~al.
\newblock Diagnosis of breast cancer using electrical impedance tomography.
\newblock Biomedical Engineering. 2012;46(4):154-7.

\bibitem{device_7}
Assenheimer M, Laver-Moskovitz O, Malonek D, Manor D, Nahaliel U, Nitzan R,
  et~al.
\newblock The T-SCANTM technology: electrical impedance as a diagnostic tool
  for breast cancer detection.
\newblock Physiological measurement. 2001;22(1):1.

\bibitem{device_8}
Hong S, Lee K, Ha U, Kim H, Lee Y, Kim Y, et~al.
\newblock A 4.9 m$\Omega$-sensitivity mobile electrical impedance tomography IC
  for early breast-cancer detection system.
\newblock IEEE Journal of Solid-State Circuits. 2015;50(1):245-57.

\bibitem{device_9}
Halter RJ, Hartov A, Paulsen KD.
\newblock A broadband high-frequency electrical impedance tomography system for
  breast imaging.
\newblock IEEE Transactions on biomedical engineering. 2008;55(2):650-9.

\end{thebibliography}
\end{singlespace}

\chapter{Figures}
\label{appendix:fig}

This appendix contains the Figures from the Chapter \ref{ch:introduction}. Also encompass the main graphical results from the Chapter \ref{ch:results} following the methodology explained in Chapter \ref{ch:methodology}. The appendix has three section, each one consist of the results from each database.

\begin{figure}[H]
\centering
\includegraphics[width=6.5in]{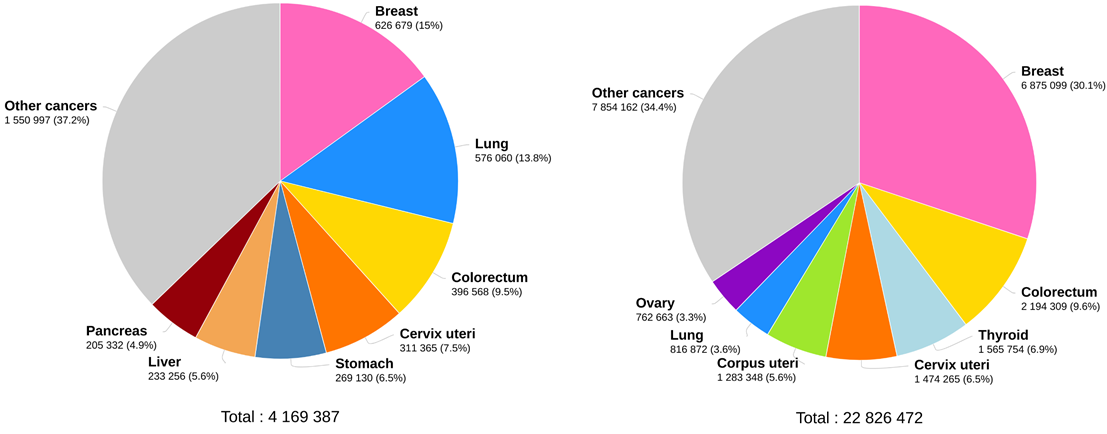} 
\caption[Pies charts of the number of cancer’s estimate number of deaths and prevalence]{Pies charts of the number of cancer’s estimate number of deaths (left) and prevalence cases in a 5-year window (right) in 2018, worldwide in women \cite{cancer_3}.}
\label{cancer_actual}
\end{figure}

\pagebreak 

\section{Blood biomarkers results}
\label{appendix:fig:BB}

The appendix \ref{appendix:fig:BB} presents additional information regarding the blood biomarkers database, going from the exploratory data analysis phase, until the hyper-parameters optimization results.  

\begin{figure}[H]
\centering
\includegraphics[width=6in]{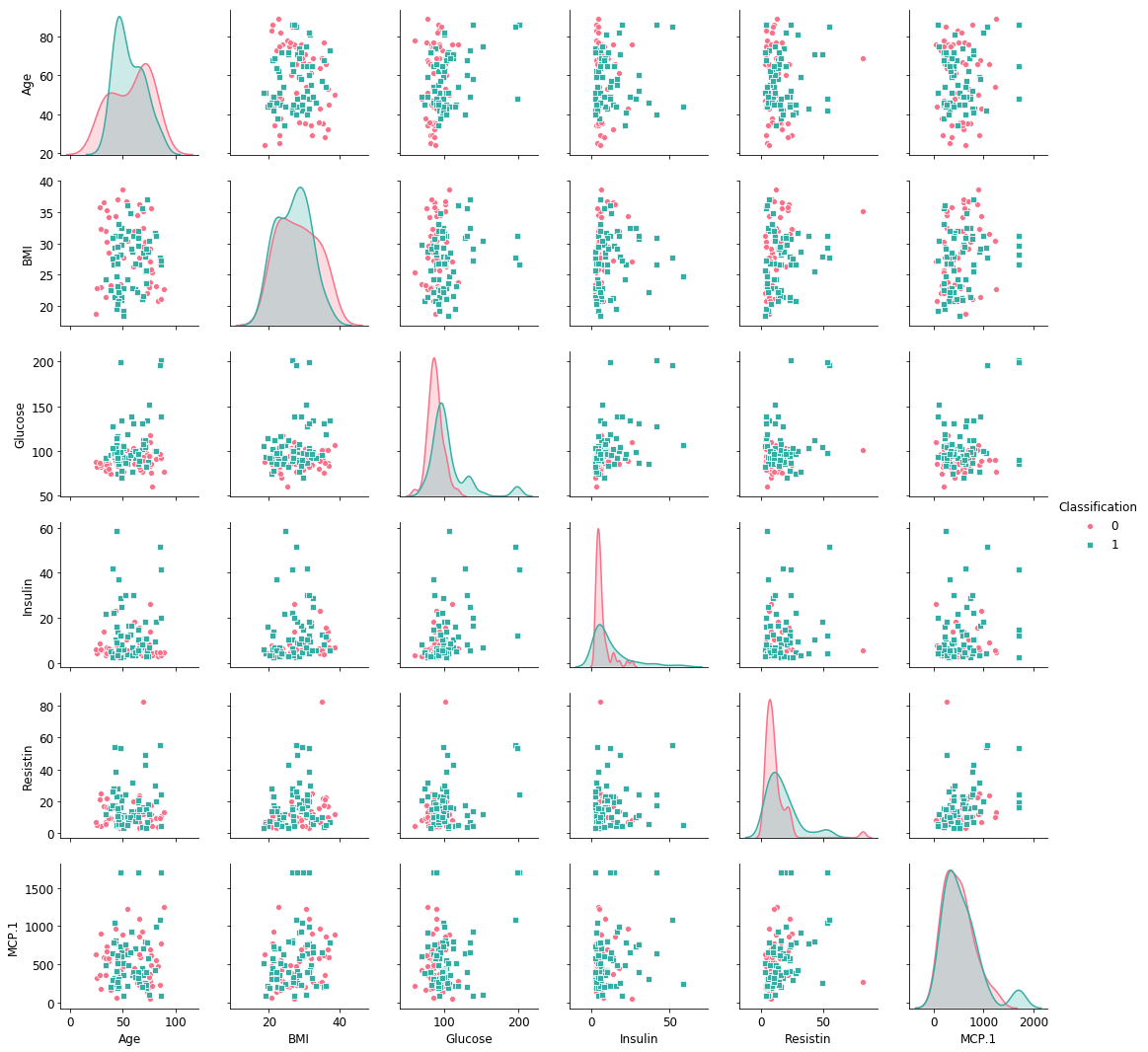} 
\caption[Pair plot of blood biomarkers]{Pair plot of blood biomarkers. Each cell show the relation between two specific features. The diagonal plots represent the distribution depending the classification, "one" stands as cancerous case and "zero" stands as a healthy patient.}
\label{res_blood_pair}
\end{figure}

\begin{figure}[H]
\centering
\includegraphics[width=6in]{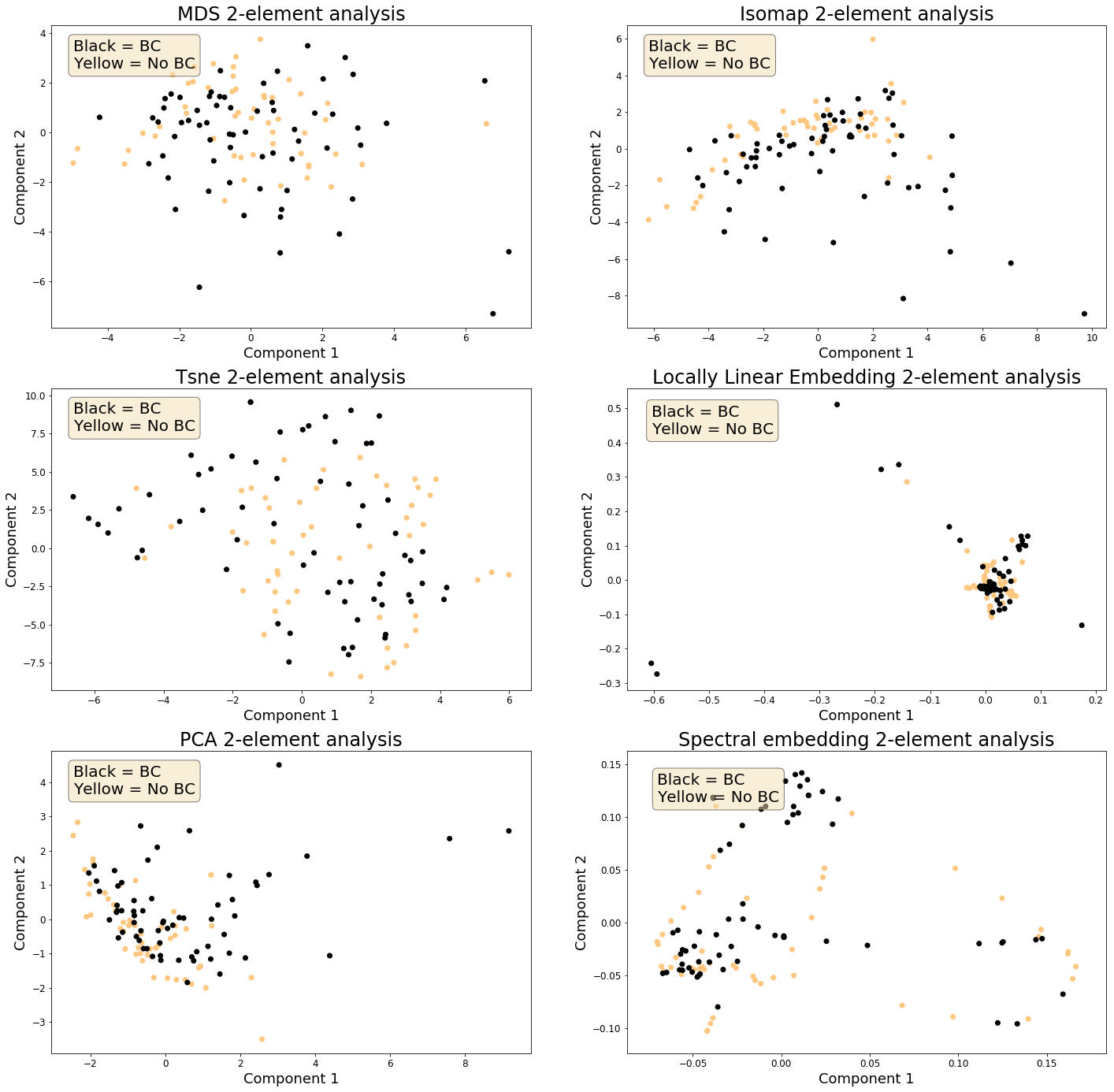} 
\caption[Six types of dimensionality reduction for the blood biomarkers database]{Six types of dimensionality reduction for the blood biomarkers database. MDS, t-SNE, PCA, isomap, locally liner embedding and spectral embedding plots from top to bottom and left to right. All the plots depicts the reduction from a 9D space into 2D.}
\label{res_DR}
\end{figure}

\begin{figure}[H]
\centering
\includegraphics[width=6in]{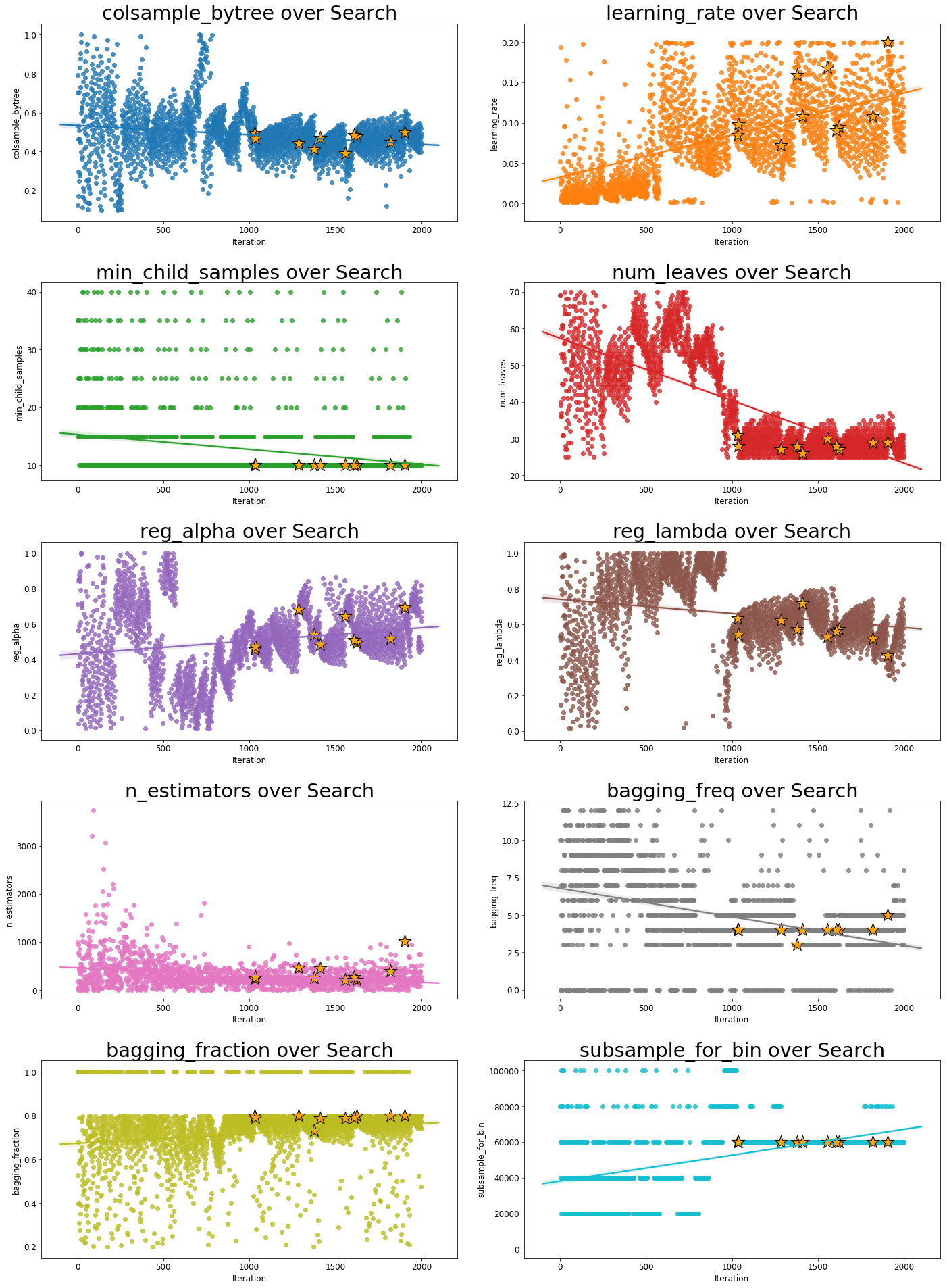} 
\caption[LGBM optimized hyper-parameters plot for blood biomarkers' database]{Optimized hyper-parameters plots for the blood biomarkers' database using a parzen tree estimator and the historical data on a LGBM model.}
\label{res_blood_PTE_LGBM}
\end{figure}

\begin{figure}[H]
\centering
\includegraphics[width=6in]{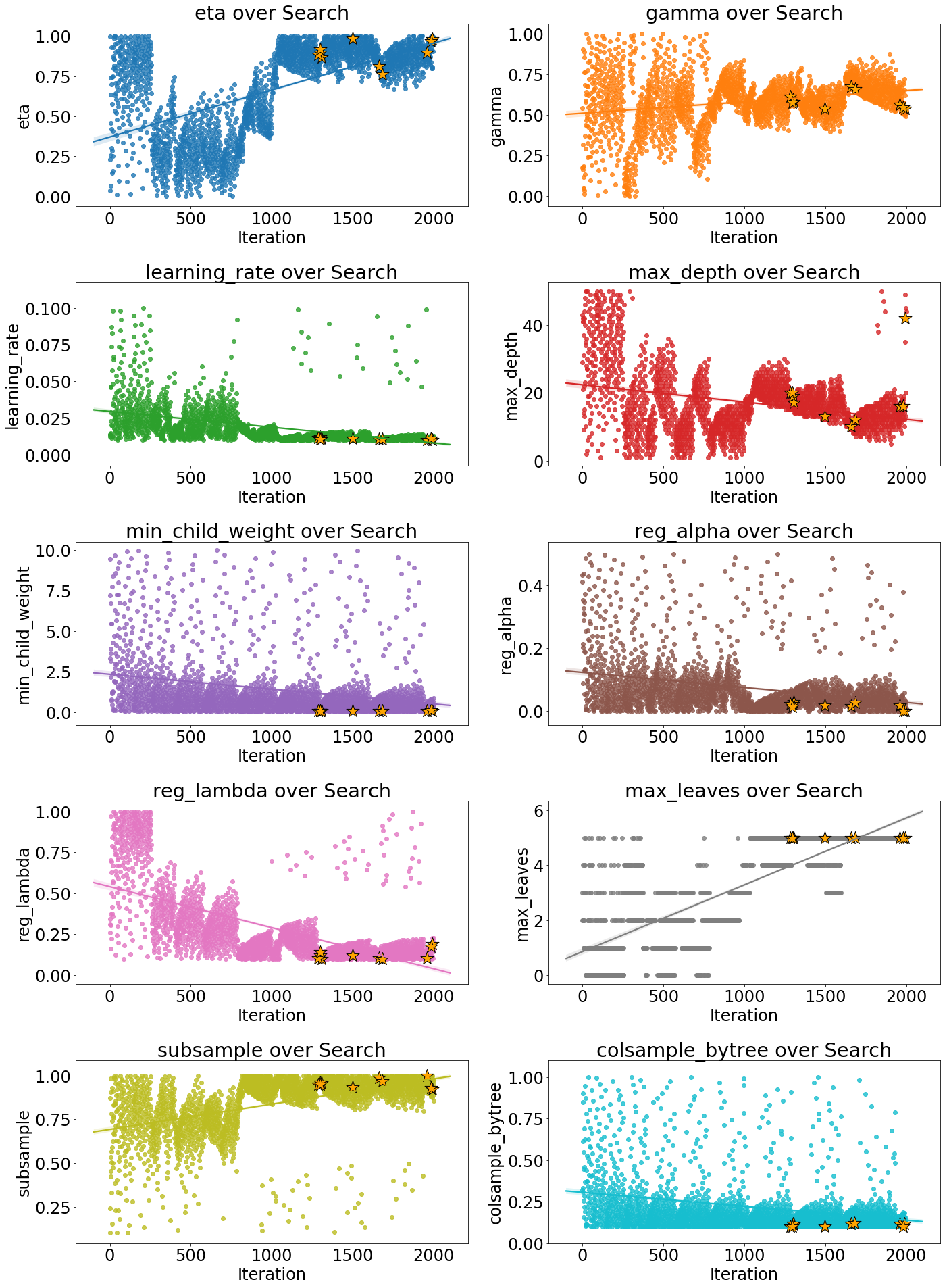} 
\caption[XGBoosting optimized hyper-parameters plot for blood biomarkers' database]{Optimized hyper-parameters plots for the blood biomarkers' database using a parzen tree estimator and the historical data on a XGBoosting model.}
\label{res_blood_PTE_XGBM}
\end{figure}

\section{Electrical impedance tomography results}
\label{appendix:fig:EIT}

The appendix \ref{appendix:fig:EIT} presents further information regarding the EIT database, going from the exploratory data analysis phase until the hyper-parameters optimization results.  

\begin{figure}[H]
\centering
\includegraphics[width=6in]{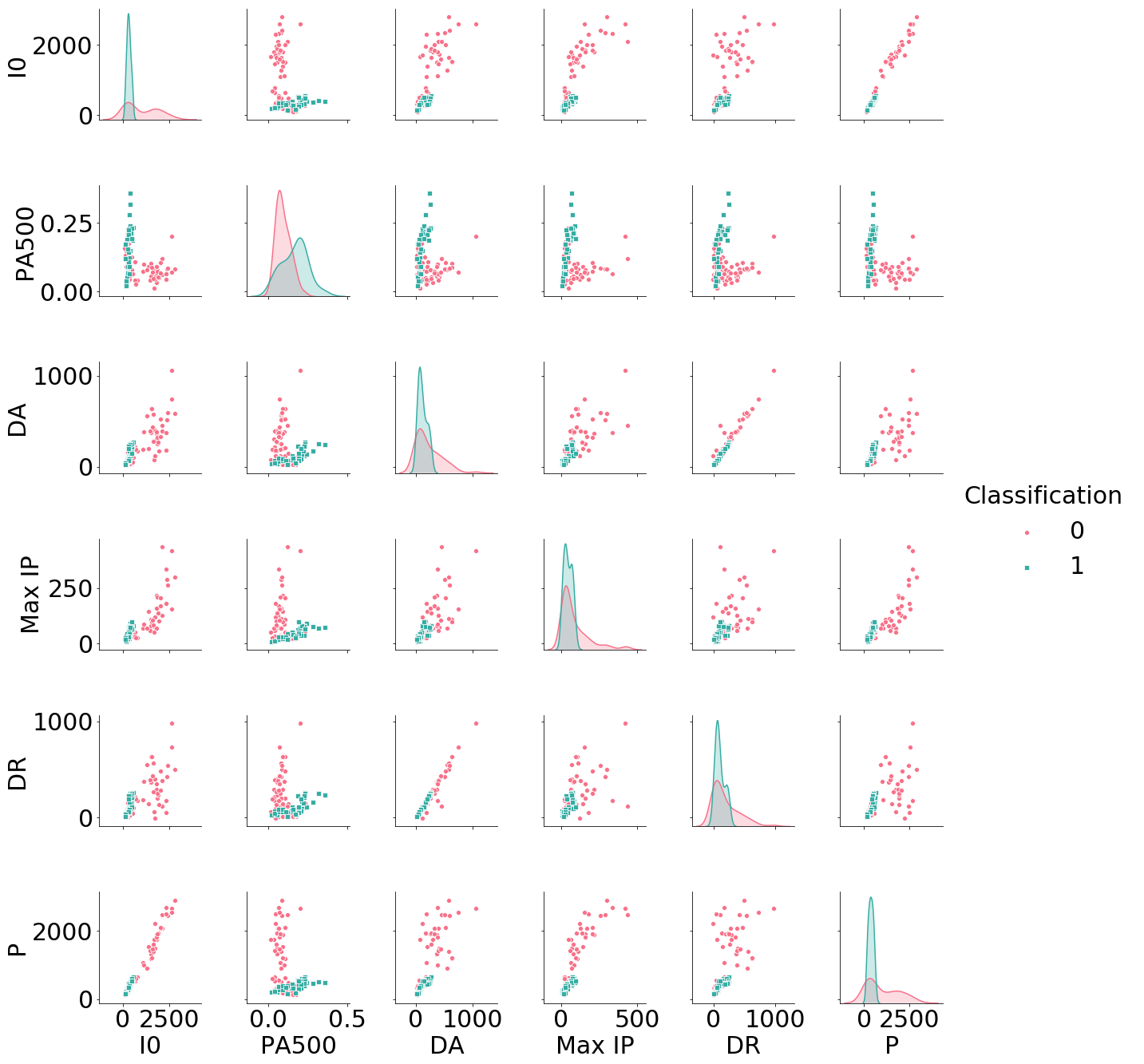} 
\caption[Pair plot of higher correlated EIT's features]{Pair plot of higher correlated EIT's features. Each cell show the relation between two specific features. The diagonal plots represent the distribution depending the classification, "one" stands as cancerous tissue and "zero" stands as a non-cancerous tissue.}
\label{res_eit_pair}
\end{figure}

\begin{figure}[H]
\centering
\includegraphics[width=6in]{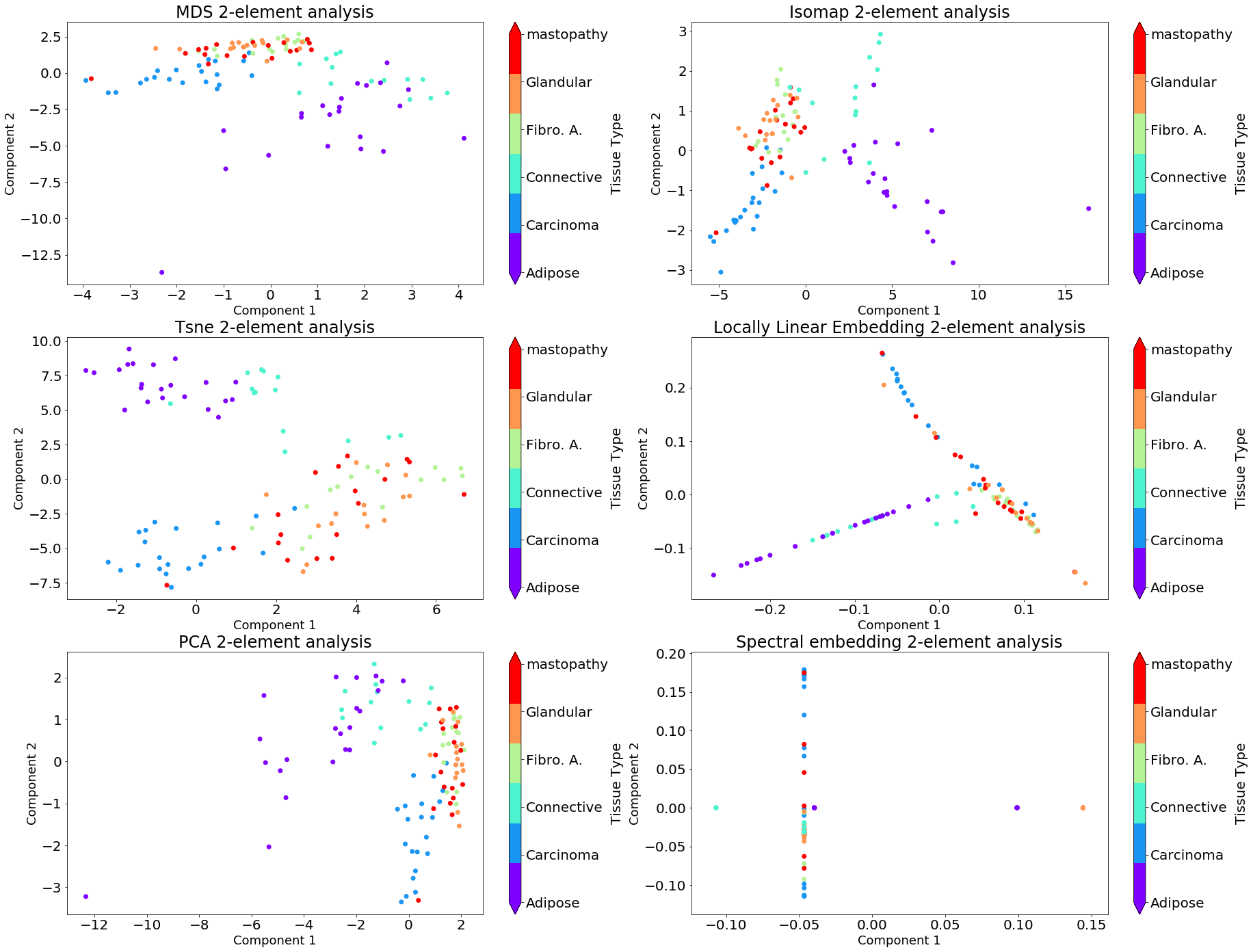} 
\caption[Six types of dimensionality reduction for electrical impedance tomography's database]{Six types of dimensionality reduction for electrical impedance tomography's database. MDS, t-SNE, PCA, isomap, locally liner embedding and spectral embedding plots from top to bottom and left to right. All the plots depicts the reduction from a 9D space into 2D.}
\label{res_EIT}
\end{figure}

\begin{figure}[H]
\centering
\includegraphics[width=5.5in]{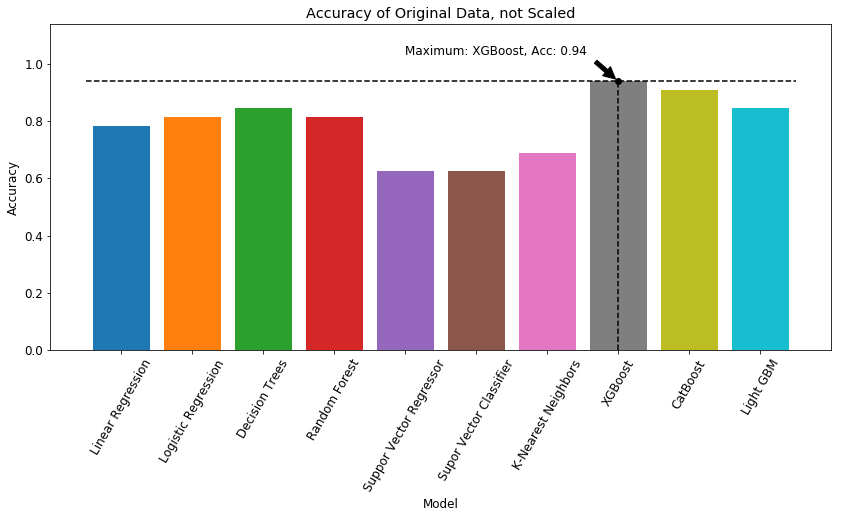} 
\caption[Performance versus threshold in the EIT's original database]{Performance versus threshold in the EIT's original database. The leading model was XGBoosting regressor. Model's accuracy, precision and recall of 94\%, 92\% and 92\%, respectively.}
\label{res_DoE_EIT_1}
\end{figure}

\begin{figure}[H]
\centering
\includegraphics[width=6in]{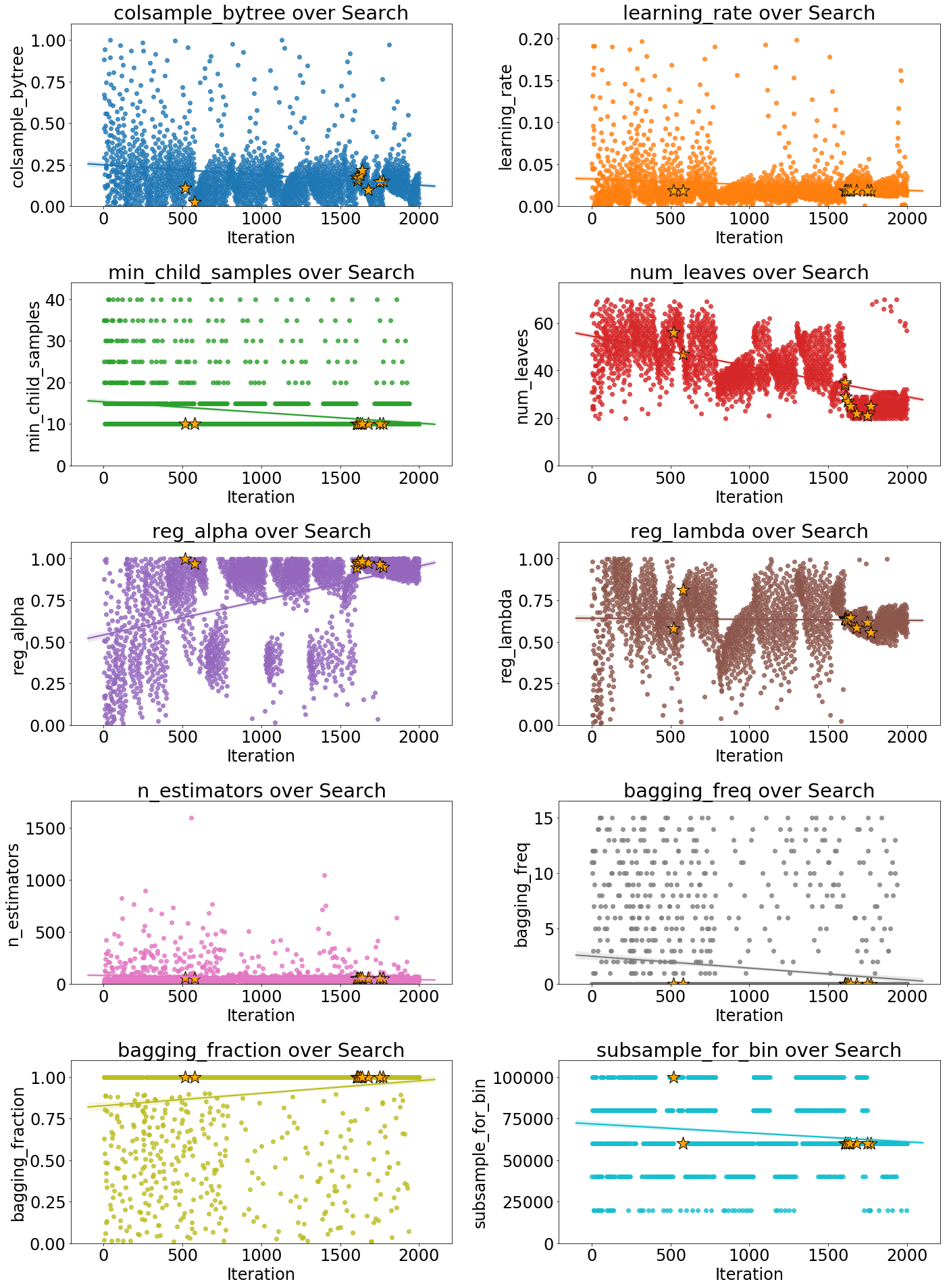} 
\caption{Optimized hyper-parameters plots for the EIT's database using a parzen tree estimator and the historical data on a LGBM model.}
\label{res_EIT_PTE_LGBM}
\end{figure}

\begin{figure}[H]
\centering
\includegraphics[width=6in]{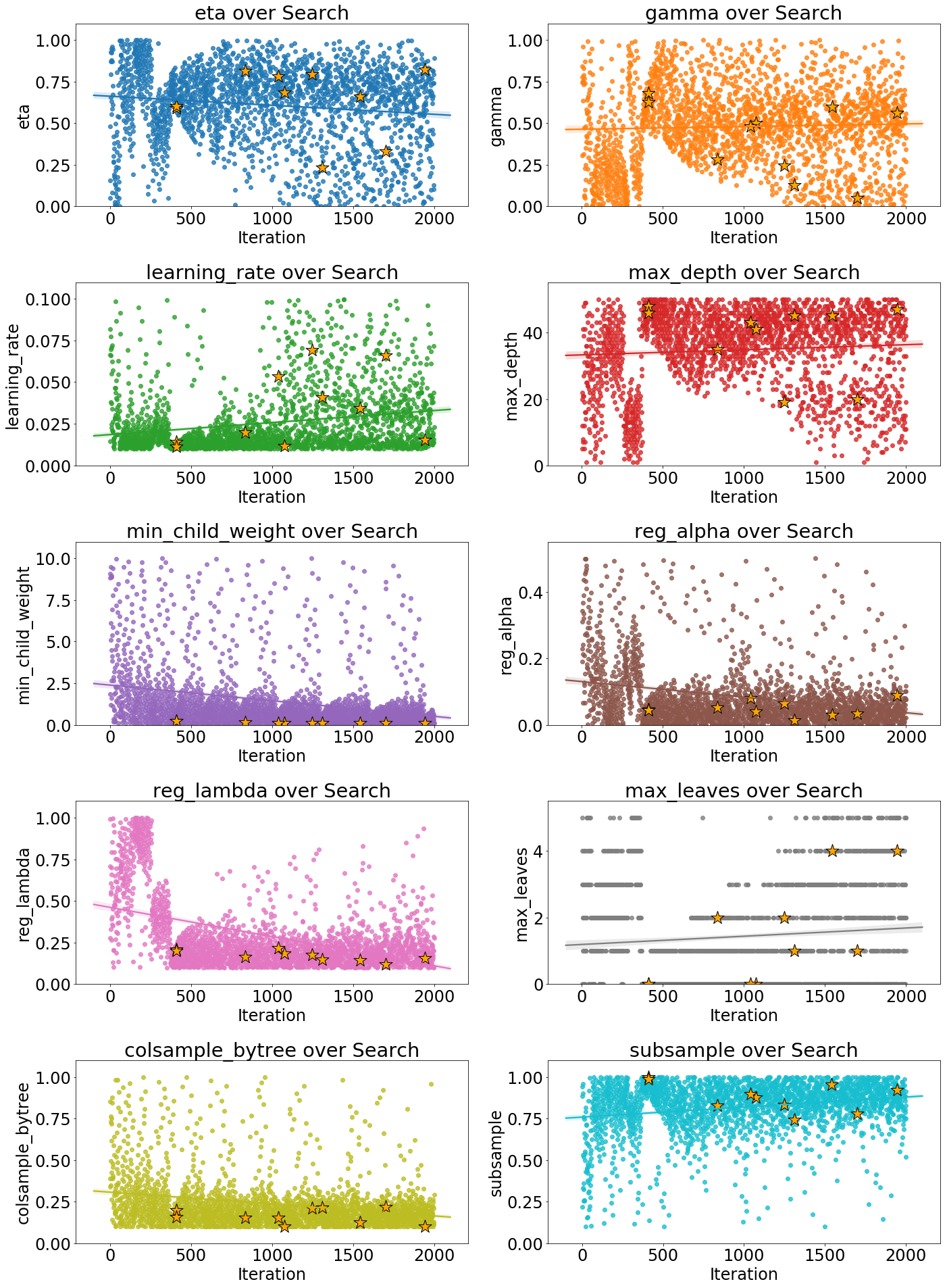} 
\caption{Optimized hyper-parameters plots for the EIT's database using a parzen tree estimator and the historical data on a XGBoosting model.}
\label{res_EIT_PTE_XGBM}
\end{figure}

\section{Thermography results}
\label{appendix:fig:thermo}

\clearpage
\newpage

\chapter{Tables}

The appendix B firstly presents three tables regarding state-of-the-art studies in thermography and electrical impedance tomography. Additionally,  further results about baseline's metrics like, accuracy, precision and recall, will allow the reader to contrast the main advantages and drawbacks from each model, e.g. processing time. 

\label{appendix:tables}

\begin{longtable}{p{7.7cm}p{3.5cm}p{2.3cm}c}
\caption[Summarized thermography methods]{Summarized thermography methods. The main parameters of evaluation are: Accuracy (Acc), sensitivity (Sen), specificity (Sp), AUC (area under the curve), ROC (receiver operating characteristic curve) and PPV (positive predictive value).}\label{summarize_table}\\
\toprule
    Scope of the project &  Machine Learning \newline Technique (MLT) & Evaluation \newline Result &  Ref. \\
\midrule

	Clustering and selection using several MLT in one ensemble unit, 5 folds cross-validation & SVM-RBF, DT \newline RF & Acc: 90\% \newline Sen: 82.6\% \newline Sp: 91.9\% &\cite{krawczyk2013}\\ 
	
	First statistical approach for breast thermograms in a ninety patients study, using a 256x200 IR camera & Bayes Rules & Acc: 59\% \newline Sen: 54\% \newline Sp: 67\% \newline PPV: 74\% &\cite{tem_new_1}\\	
	
    Classifying normal/at risk in confirmed cases of cancer, normal and post operation & No MLT  \newline Weighted Algorithm & Acc: 99\% \newline Sen: 99\% & \cite{new_thermography_1}\\

    Localization of skin tumors, based on temperature & Genetic algorithm & Acc: 100\%\footnotemark &\cite{new_thermography_2} \\
    
	(continued on next page) & & & \\
	
	Table \ref{summarize_table} (continued) &  &  &\\
	
\toprule
    Scope of the project &  Machine Learning \newline Technique (MLT) & Evaluation \newline Result &  Ref. \\
\midrule
     
    Presence, location, size and properties of the tumor & Genetic algorithm & Max Error \newline E:2.6\% &\cite{new_thermography_2_1} \\    
    
    Analysis and comparison of thermography vs breast cancer screening techniques & No MLT & Acc: 83\% \newline Sen: 83\% & \cite{new_thermography_3}\\
        
    Thermography classifier of breast cancer - Graphical user Interface & SVM & Acc: 88.1\% \newline Sen: 85.7\% \newline Sp: 90.48\% &\cite{new_thermography_4}\\   
     
    AI approach for breast cancer diagnosis with thermograms, taking in account the menstrual cycle & ANN, Bayes Rules & Acc: 61.5\% \newline Sen: 69\% \newline Sp: 40\% \newline PPV: 90.91\% &\cite{tem_new_3}\\
     		
	Integrated technique for breast cancer diagnosis using a bio-statistical method as pre-processing technique & ANN, linear regression (RBFN) & Acc: 80.9\% \newline Sen: 100\% \newline Sp: 71\% &\cite{tem_new_4}\\
     
     Statistical features from infrared signals and asymmetry from both breasts & Fuzzy Logic classifier & Acc: 79.5\% \newline Sen: 79.9\% \newline Sp: 79.5\% &\cite{new_thermography_5}\\
  		         
	Symbolic data analysis for classification as malignant, benign and cyst breasts thermograms from \cite{marques2012} and \cite{silva2015} database & Linear Discriminant \newline Parzen-window & Acc: 84\% \newline Sen: 85.7\% \newline Sp: 86.5\% &\cite{new_thermography_6}\\ 

	Normalization of breast cancer infrared images improve the global accuracy (min-max approach)& SVM \newline Kernel: Gaussian & Acc: 91\% \newline Sen: 87.23\% \newline Sp: 94.34\% &\cite{new_thermography_8}\\ 				
	Current status of breast cancer diagnosis with thermography \cite{cancer_4, review_thermography_1, review_thermography_2} & Many MLT & -- & \cite{review_thermography_3}\\
		
	Breast thermal images segmentation as a pre-processing method & EHMM & Execution time reduced &\cite{new_thermography_7}\\
	 
	(continued on next page) & & & \\
	
	Table \ref{summarize_table} (continued) &  &  &\\
	
\toprule
    Scope of the project &  Machine Learning \newline Technique (MLT) & Evaluation \newline Result &  Ref. \\
\midrule

	Analysis of thermal breast images as time series. First, region of interest (ROI) is segmented and then \textit{k-means} is implemented & Bayes Net \newline Decision Table \newline RF & Acc: 95.4\% \newline Sen: 95.37\% \newline Sp: 95.4\% \newline ROC: 0.97 &\cite{new_thermography_9}\\ 
	
	Extensive review in last advances regarding dynamic breast thermography's & Many MLT&  --  &\cite{new_thermography_9_1}\\
		
	Analysis of thermal patches as key features for determine the presence of cancerous tissue in the breast & SVM, DT, AB, RF, KNN, ANN, LDA& Acc: 98\% \newline Sen: 98\% \newline Sp: 98\% &\cite{new_thermography_13}\\

	Thermal breast model in COMSOL\textregistered , with tissue properties per layers & -- & -- &\cite{new_thermography_12}\\ 	

	Tumor localization from skin temperatures in COMSOL\textregistered  & -- & -- &\cite{new_thermography_12_1}\\ \\	

	Comparison between Temperature based analysis, intensity based analysis and tumor location matching & SVM & Acc: 83.2\% \newline Sen: 85.6\% \newline Sp: 73.2\% &\cite{new_thermography_14}\\		

	Analysis of thermography (DITI) in the diagnosis of breast mass, side diagnostic with Ultrasound and/or MRI as validation techniques& No NLT & Acc: 79.6\% \newline Sen: 95.2\% \newline Sp: 72.8\% &\cite{new_thermography_15}\\	

	\bottomrule
\end{longtable}

\footnotetext{Tiny tumors cannot be detected with this method \cite{new_thermography_2}}

\pagebreak


\begin{longtable}{p{7.6cm}p{3.5cm}p{2.5cm}c}
\caption[Summarized Electrical Impedance Tomography methodologies]{Summarized Electrical Impedance Tomography methodologies. The main parameters of evaluation are: Accuracy (Acc), sensitivity (Sen), specificity (Sp), AUC (area under the curve), ROC (receiver operating characteristic curve), correlation coefficient (CC) and mean square error (MSE).}
\label{summarize_table_eit}\\
\toprule
    Scope of the project &  Machine Learning \newline Technique (MLT) & Evaluation \newline Result &  Ref. \\
\midrule

	Impedance variability on six breast tissue, 9 different features, 1 target & No MLT & Mean ($\mu$) \newline Std. Dev. ($\sigma$)&\cite{eit_1}\\ 

	Classification of breast sample tissues using STATISTICA\textregistered , 9 features, 1 target & Statistical analysis & Acc: 92\% &\cite{eit_2}\\ 

	Modeling of EIT distribution in a breast with a tumor, using ANN, PCA amd BEM & ANN & Abs. E: 0.23 & \cite{eit_6}\\ 

	Review of the main advances in EIT for breast cancer diagnosis & No MLT & -- &\cite{eit_5}\\ 

	First test ever of EIT in breast cancerous tissue (1926) & No MLT & -- &\cite{eit_first}\\
				 
	REIS system with 7 electrodes, 1 in the center, and 6 concentrically separated.REIS showed high false positive rate & ANN & Acc: 67\% \newline Sen: 54\% \newline Sp: 90\% & \cite{eit_7}\\
		
	Multi-Layer perceptron algorithm for the EIT data set from \cite{eit_1, eit_2} & ANN - MLP & Acc: 96\% \newline MSE: 0.1 \newline CC: 0.99 & \cite{eit_8}\\ 	

	CAD system for breast cancer classification in the EIT data set from \cite{eit_1, eit_2} & LR + NBN & Acc-1: 97.5\% \newline  Acc-2: 89.7\% \newline  Acc-3: 77.35\% & \cite{eit_9}\\
		
	EIT system for early detection of breast cancer in 1103 women & Multi LR & -- & \cite{eit_10}\\

	10 women clinical study, using 2 different setups with a EIT-Probe & LDA, LSE & -- & \cite{eit_11}\\	
		
	System for make a 3D breast's map, it uses 85 electrodes and frequencies from 10kHz - 3MHz & -- & -- & \cite{eit_12}\\				
	\bottomrule
\end{longtable}

\begin{landscape}
\addtolength{\voffset}{40pt}
\begin{table}
\large
	\begin{center}
	\captionsetup{font=Large}
	\caption{\label{table_eit_devices}Electrical Impedance Tomography devices and properties}
	\begin{tabular}{p{3cm}  p{4cm} p{4cm}  p{4cm}  p{4cm}  p{4cm}}
	   \hline
	   Reference & USA \cite{device_5} & Russia \cite{device_6} & Germany \cite{device_7} & Korea \cite{device_8} & USA \cite{device_9} \\
	   \hline
	   Device &
	   \includegraphics[width=1.6in]{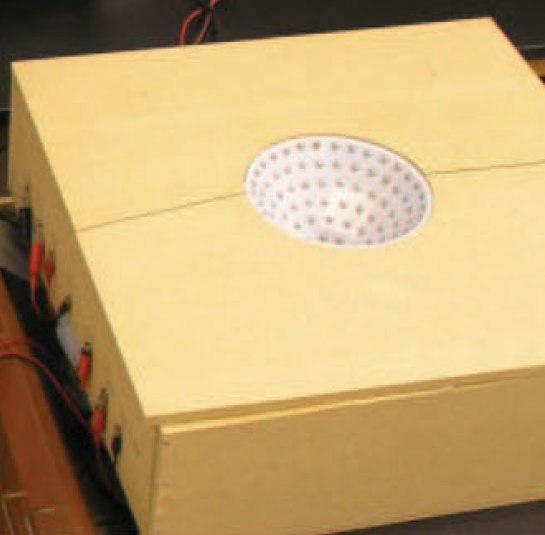} &
	   \includegraphics[width=1.6in]{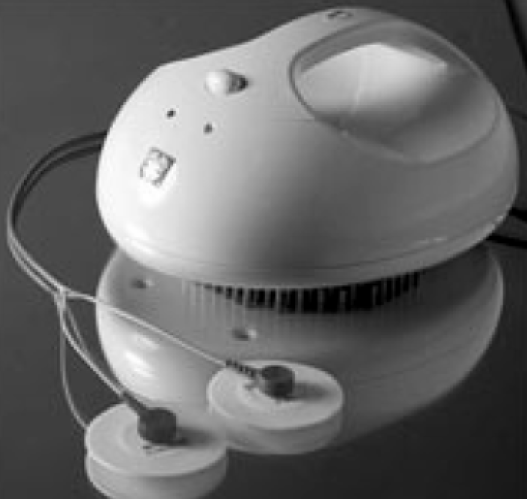} &
	   \includegraphics[width=1.6in]{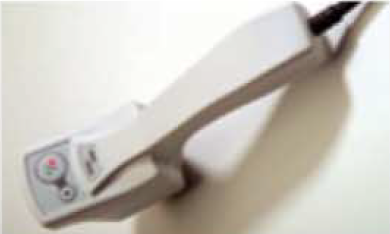} &
	   \includegraphics[width=1.6in]{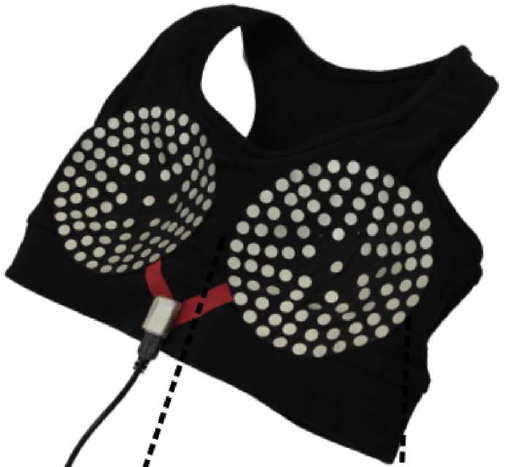} &
	   \includegraphics[width=1.6in]{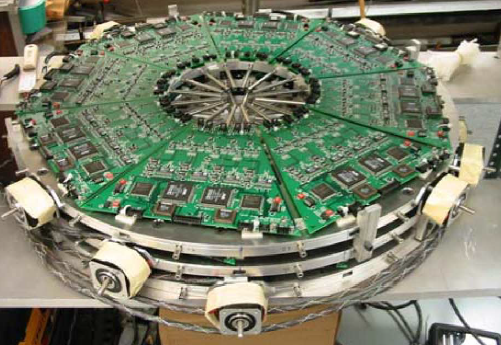} \\
	   \hline \\[-1.5ex]
	   Year & 2008 & 2012 & 2001 & 2014 & 2007 \\
	   Location & Bed & Hand-held device \newline +ref. electrode & Probe \newline + ref. Probe & BRA (wearable) & Bed \\
	   Dimension & 11.7 cm height \newline 19.1 cm diameter & 16x18x10 cm & 7.2x7.2 cm & 30x25x5 cm & 60 cm diameter \\
	   Weight & N/A & 2 Kg & N/A & 72g & N/A \\
	   Img. device & Computer & Computer & Computer & Mobile device & Computer \\
	   Dimension & 2D slices & 2D slices & 2D  & 3D & 3D \\
	   \# electrodes & 128 (7 layers) & 256 (planar) & 256 (planar)  &  92 (flexible) & 64 (4 layers) \\
	   Frequency & 10kHz & 10 - 50 kHz & 58 Hz - 5kHz  &  100 Hz - 100kHz & 10kHz - 10MHz \\
	   Amplitude & 1mA & 0.5mA & 1V - 2.5V & 10$\mu$A - 400$\mu$A &  N/A \\
	   SNR (dB) & 77dB & N/A & N/A  &  90dB & 94dB \\
	   Minimum \newline detectable size  & 12mm & N/A & N/A  &  5mm & N/A \\
	   \hline	   	   	   
	\end{tabular}
	\end{center}
\end{table}
\end{landscape}


\begin{table}[htb]
\caption[Design of experiments top metrics for the blood biomarkers database]{Design of experiments top metrics for the blood biomarkers database. Each experiment has nine MLT, likewise, each MLT has five metrics (accuracy, precision, recall, F1 score and ROC AUC), the threshold (between 0.2 and 0.8) which yielded the best performance, and the execution time.}
\label{resul_Doe_blood}
\begin{tabular}{@{}lccccccc@{}}
\toprule
                    & \multicolumn{7}{l}{Original Dataset}                                       \\ \cmidrule{2-8}
Model               & Accuracy & Precision & Recall & F1 score & ROC AUC & Threshold & Time (ms) \\ \midrule
Linear Reg.   & 0.80     & 0.92      & 0.67   & 0.77     & 0.80    & 0.54      & 0.00      \\
Logistic Reg. & 0.89     & 1.00      & 0.78   & 0.88     & 0.89    & 0.59      & 0.00      \\
Decision Tree       & 0.80     & 0.79      & 0.83   & 0.81     & 0.80    & 0.30      & 0.00      \\
Random Forest       & 0.71     & 0.67      & 0.89   & 0.76     & 0.71    & 0.30      & 1.56      \\
SVC                 & 0.51     & 0.51      & 1.00   & 0.68     & 0.50    & 0.30      & 0.00      \\
KNN                 & 0.51     & 0.55      & 0.33   & 0.41     & 0.52    & 0.80      & 1.56      \\
XGB Regressor       & 0.77     & 0.71      & 0.94   & 0.81     & 0.77    & 0.30      & 3.12      \\
CatBoost            & 0.71     & 0.67      & 0.89   & 0.76     & 0.71    & 0.30      & 397.63    \\
LGBM                & 0.83     & 0.83      & 0.83   & 0.83     & 0.83    & 0.52      & 1.57      \\
\bottomrule
                    & \multicolumn{7}{l}{Scaled Dataset}                                       \\ \cmidrule{2-8}
Linear Reg.   & 0.80     & 0.92      & 0.67   & 0.77     & 0.80    & 0.54      & 0.10      \\
Logistic Reg. & 0.80     & 0.92      & 0.67   & 0.77     & 0.80    & 0.56      & 0.10      \\
Decision Tree       & 0.80     & 0.79      & 0.83   & 0.81     & 0.80    & 0.30      & 0.00      \\
Random Forest       & 0.71     & 0.64      & 1.00   & 0.78     & 0.71    & 0.30      & 0.70      \\
SVC                 & 0.89     & 0.89      & 0.89   & 0.89     & 0.89    & 0.56      & 0.30      \\
KNN                 & 0.80     & 0.76      & 0.89   & 0.82     & 0.80    & 0.56      & 0.10      \\
XGB Reg.       & 0.77     & 0.71      & 0.94   & 0.81     & 0.77    & 0.30      & 3.59      \\
CatBoost            & 0.71     & 0.67      & 0.89   & 0.76     & 0.71    & 0.30      & 395.88    \\
LGBM                & 0.80     & 0.79      & 0.83   & 0.81     & 0.80    & 0.49      & 1.00      \\ \bottomrule
                    & \multicolumn{7}{l}{Expanded and Augmented Dataset}			\\ \cmidrule{2-8}
Linear Reg.   & 0.79     & 0.80      & 0.86   & 0.83     & 0.78    & 0.44      & 0.00      \\
Logistic Reg. & 0.81     & 0.83      & 0.85   & 0.84     & 0.80    & 0.42      & 1.56      \\
Decision Tree       & 0.76     & 0.81      & 0.77   & 0.79     & 0.76    & 0.30      & 0.00      \\
Random Forest       & 0.80     & 0.86      & 0.79   & 0.82     & 0.81    & 0.56      & 1.56      \\
SVC                 & 0.58     & 0.58      & 1.00   & 0.73     & 0.50    & 0.30      & 7.81      \\
KNN                 & 0.75     & 0.77      & 0.80   & 0.79     & 0.74    & 0.56      & 0.00      \\
XGB Regressor       & 0.93     & 0.96      & 0.92   & 0.94     & 0.93    & 0.52      & 9.37      \\
CatBoost            & 0.89     & 0.89      & 0.93   & 0.91     & 0.88    & 0.44      & 675.69    \\
LGBM                & 0.88     & 0.86      & 0.94   & 0.90     & 0.87    & 0.46      & 6.25      \\ \bottomrule
                    & \multicolumn{7}{l}{Scaled, Expanded and Augmented Dataset}			\\ \cmidrule{2-8}
Linear Reg.   & 0.79     & 0.78      & 0.89   & 0.83     & 0.77    & 0.44      & 0.00      \\
Logistic Reg. & 0.79     & 0.83      & 0.81   & 0.82     & 0.79    & 0.49      & 0.00      \\
Decision Tree       & 0.75     & 0.78      & 0.79   & 0.78     & 0.74    & 0.30      & 0.00      \\
Random Forest       & 0.86     & 0.86      & 0.91   & 0.88     & 0.85    & 0.44      & 3.12      \\
SVC                 & 0.81     & 0.81      & 0.88   & 0.84     & 0.80    & 0.51      & 7.81      \\
KNN                 & 0.74     & 0.72      & 0.91   & 0.80     & 0.71    & 0.30      & 0.00      \\
XGB Regressor       & 0.92     & 0.89      & 0.98   & 0.93     & 0.91    & 0.47      & 9.37      \\
CatBoost            & 0.87     & 0.83      & 0.99   & 0.90     & 0.85    & 0.37      & 674.87    \\
LGBM                & 0.87     & 0.89      & 0.89   & 0.89     & 0.87    & 0.52      & 6.25      \\ \bottomrule
\end{tabular}
\end{table}


\begin{table}[htb]
\caption[Design of experiments top metrics for the electrical impedance tomography's database]{Design of experiments top metrics for the electrical impedance tomography's database. Each experiment has nine MLT, likewise, each MLT has five metrics (accuracy, precision, recall, F1 score and ROC AUC), the threshold (between 0.2 and 0.8) which yielded the best performance, and the execution time.}
\label{resul_Doe_EIT}
\begin{tabular}{@{}lcccccc@{}}
\toprule
                    & \multicolumn{6}{l}{Original Dataset}                                       \\ \cmidrule{2-7}
Model               & Accuracy & Precision & Recall & F1 score & ROC AUC & Time (ms) \\ \midrule
Linear regression & 0.78 & 0.78 & 0.58 & 0.67 & 0.74 & 0.00 \\
Logistic regression & 0.81 & 0.88 & 0.58 & 0.70 & 0.77 & 1.56\\
Decision Tree & 0.84 & 0.77 & 0.83 & 0.80 & 0.84 & 0.00\\
Random Forest & 0.81 & 0.75 & 0.75 & 0.75 & 0.80 & 0.00\\
SVC & 0.63 & 0.00 & 0.00 & 0.00 & 0.50 & 0.00\\
KNN & 0.69 & 0.58 & 0.58 & 0.58 & 0.67 & 1.56\\
XGB regression & 0.94 & 0.92 & 0.92 & 0.92 & 0.93 & 1.56\\
CatBoost & 0.91 & 1.00 & 0.75 & 0.86 & 0.88 & 430.35\\
LGBM & 0.84 & 0.89 & 0.67 & 0.76 & 0.81 & 1.56\\
\bottomrule
                    & \multicolumn{6}{l}{Scaled Dataset}                                       \\ \cmidrule{2-7}
Linear regression & 0.78 & 0.78 & 0.58 & 0.67 & 0.74 & 0.00 \\
Logistic regression & 0.75 & 0.70 & 0.58 & 0.64 & 0.72 & 0.0\\
Decision Tree & 0.88 & 0.79 & 0.92 & 0.85 & 0.88 & 0.00 \\
Random Forest & 0.81 & 0.75 & 0.75 & 0.75 & 0.80 & 1.56\\
SVC & 0.84 & 1.00 & 0.58 & 0.74 & 0.79 & 0.00\\
KNN & 0.88 & 1.00 & 0.67 & 0.80 & 0.83 &  0.00\\
XGB regression & 0.94 & 0.92 & 0.92 & 0.92 & 0.93 & 1.56\\
CatBoost & 0.91 & 1.00 & 0.75 & 0.86 & 0.88 & 451.04\\
LGBM & 0.84 & 0.89 & 0.67 & 0.76 & 0.81 & 0.00\\
\bottomrule
                    & \multicolumn{6}{l}{Scaled and Expanded Dataset}			\\ \cmidrule{2-7}
Linear regression & 0.88 & 0.83 & 0.83 & 0.83 & 0.87 & 0.00\\
Logistic regression & 0.81 & 0.71 & 0.83 & 0.77 & 0.82 & 0.00 \\
Decision Tree & 0.81 & 0.71 & 0.83 & 0.77 & 0.82 & 0.00\\
Random Forest & 0.88 & 0.90 & 0.75 & 0.82 & 0.85 & 0.00\\
SVC & 0.63 & 0.00 & 0.00 & 0.00 & 0.50 & 0.00\\
KNN & 0.59 & 0.45 & 0.42 & 0.43 & 0.56 & 0.00\\
XGB regression & 0.94 & 1.00 & 0.83 & 0.91 & 0.92 & 1.56\\
CatBoost & 0.91 & 0.91 & 0.83 & 0.87 & 0.89 & 441.13\\
LGBM & 0.84 & 0.77 & 0.83 & 0.80 & 0.84 & 0.00\\
\bottomrule
                    & \multicolumn{6}{l}{Scaled, Expanded and Augmented Dataset}			\\ \cmidrule{2-7}
Linear regression & 0.80 & 0.71 & 0.68 & 0.70 & 0.77 & 0.00 \\
Logistic regression & 0.70 & 0.56 & 0.64 & 0.60 & 0.69 &0.00 \\
Decision Tree & 0.80 & 0.68 & 0.77 & 0.72 & 0.79 &0.00 \\
Random Forest & 0.73 & 0.60 & 0.68 & 0.64 & 0.72 & 0.00 \\
SVC & 0.64 & 0.00 & 0.00 & 0.00 & 0.49 & 1.56 \\
KNN & 0.63 & 0.47 & 0.64 & 0.54 & 0.63 &  0.00 \\
XGB regression & 0.81 & 0.73 & 0.73 & 0.73 & 0.79 &  4.82 \\
CatBoost & 0.81 & 0.73 & 0.73 & 0.73 & 0.79 & 549.18 \\
LGBM & 0.91 & 0.90 & 0.82 & 0.86 & 0.89 & 1.56\\
\bottomrule
\end{tabular}
\end{table}


\begin{table}[htb]
\centering
\caption[Top metrics in the benchmark of CNN architectures on the thermography's database (unbiased)]{Top metrics in the thermography's database. Each CNN architecture has five metrics (precision, recall, F1 score, accuracy and ROC AUC) and execution time per each epoch. This table display the benchmark results of the predefined models from the Keras application module for the unbiased experiments.}
\label{resul_thermal_CNN_DNN_tbl_3_unbiased}
\begin{tabular}{@{}lcccccc@{}}
\toprule
                    & \multicolumn{6}{l}{Convolutional Neural Network top models}                                       \\ \cmidrule{2-7}
Model               & Precision & Recall & F1 score & Accuracy & ROC AUC (s) & Time (s)\\ 
\midrule
ResNet50 & 0.90 & 0.68 & 0.77 & 0.79 & 0.80 & 30.00 \\
VGG16 & 1.00 & 0.52 & 0.68 & 0.74 & 0.76 & 29.00 \\
InceptionV3 & 1.00 & 0.44 & 0.62 & 0.70 & 0.72 & 18.00 \\
Inception V3 & 1.00 & 0.83 & 0.90 & 0.90 & 0.91 & 21.00 \\
InceptionResNetV2 & 0.54 & 0.50 & 0.52 & 0.50 & 0.50 & 45.00 \\
InceptionResNetV2 & 0.57 & 0.79 & 0.66 & 0.58 & 0.56 & 53.00 \\
Xception & 0.63 & 0.83 & 0.71 & 0.64 & 0.63 & 29.00 \\
Xception & 0.72 & 0.80 & 0.76 & 0.73 & 0.72 & 47.00\\ 
\bottomrule
\end{tabular}
\end{table}

\begin{table}[htb]
\centering
\caption[Top metrics and architectures of the created CNN models with predefined blocks for the thermography's database (unbiased)]{Top metric sand architectures of the created CNN models in the thermography's database using the predefined blocks from image \ref{res_thermography_CNN_arch_new}. Each CNN architecture has five metrics (precision, recall, F1 score, accuracy and ROC AUC) and execution time per each epoch. This table display the results and the architectures for each experiment for the unbiased experiments.}
\label{resul_thermal_CNN_new_blocks}
\begin{tabular}{@{}lcccccc@{}}
\toprule
                    & \multicolumn{6}{l}{Convolutional Neural Network top models}                                       \\ \cmidrule{2-7}
Model               & Precision & Recall & F1 score & Accuracy & ROC AUC (s) & Time (s)\\ 
\midrule
CNN 1 & 0.83 & 0.89 & 0.86 & 0.85 & 0.84 & 10 \\
CNN 2 & 0.88 & 0.91 & 0.89 & 0.88 & 0.88 & 30 \\
CNN 3 & 0.92 & 0.75 & 0.82 & 0.83 & 0.84 & 41 \\
CNN 4 & 0.94 & 0.91 & 0.92 & 0.92 & 0.92 & 40 \\
CNN 5 & 0.61 & 0.96 & 0.75 & 0.66 & 0.63 & 50 \\
CNN 6 & 0.80 & 0.82 & 0.81 & 0.79 & 0.79 & 33 \\
CNN 7 & 0.84 & 0.90 & 0.87 & 0.86 & 0.85 & 46 \\
CNN 8 & 0.99 & 0.83 & 0.90 & 0.90 & 0.91 & 64\\ 
\toprule
                    & \multicolumn{6}{l}{Convolutional Neural Network top models}                                       \\ \cmidrule{2-7}
Model               & Num. blocks & Layer/block & Optimizer & Kernel & Top Layer & Dropout\\ 
\midrule
CNN 1 & 2.00 & 3.00 & Adam & 3.00 & GAP & 0.01 \\
CNN 2 & 4.00 & 2.00 & Adam & 3.00 & GAP & 0.10 \\
CNN 3 & 6.00 & 2.00 & Adam & 4.00 & GAP & 0.00 \\
CNN 4 & 6.00 & 3.00 & RMSPROP & 3.00 & Flatten & 0.00 \\
CNN 5 & 7.00 & 2.00 & RMSPROP & 3.00 & Flatten & 0.00 \\
CNN 6 & 7.00 & 2.00 & RMSPROP & 3.00 & GAP & 0.10 \\
CNN 7 & 5.00 & 3.00 & RMSPROP & 3.00 & GAP & 0.50 \\
CNN 8 & 6.00 & 4.00 & RMSPROP & 3.00 & GAP & 0.00\\ 
\bottomrule
\end{tabular}
\end{table}

\clearpage
\newpage

\end{document}